\documentclass{article}
\usepackage{arxiv}

\usepackage{mathtools}
\usepackage[utf8]{inputenc} 
\usepackage[T1]{fontenc}    
\usepackage{hyperref}       
\usepackage{url}            
\usepackage{booktabs}       
\usepackage{amsfonts}       
\usepackage{nicefrac}       
\usepackage{microtype}      
\usepackage{xcolor}         
\usepackage{cite}    

\usepackage{bm}
\usepackage{bbm}
\usepackage{comment}         
\usepackage{multicol}
\usepackage{multirow}
\usepackage{caption}
\usepackage{braket}
\usepackage{enumitem}

\mathchardef\mhyphen="2D

\usepackage{amsmath,amssymb,amsthm}

\theoremstyle{plain}

\newtheorem{theorem}{Theorem}[section]
\newtheorem{lemma}[theorem]{Lemma}

\newtheorem{corollary}[theorem]{Corollary}
\newtheorem{proposition}[theorem]{Proposition}
\newtheorem{remark}{Remark}

\newtheorem{definition}[theorem]{Definition}

\usepackage{color}

\usepackage{graphicx}

\DeclareMathOperator*{\argmax}{arg\,max}
\DeclareMathOperator*{\argmin}{arg\,min}

\usepackage{times}

\newcommand{\pa}{\mathrm{\pa}}

\newcommand{\RN}[1]{%
  \textup{\uppercase\expandafter{\romannumeral#1}}%
}

\usepackage{xcolor}
\newcount\Comments  
\newcommand{\kibitz}[2]{\ifnum\Comments=1\textcolor{#1}{#2}\fi}

\usepackage{algorithm}
\usepackage{algorithmic}

\newcommand*{\email}[1]{\href{mailto:#1}{#1}}

\usepackage{authblk}
\allowdisplaybreaks

\title{Federated Best Arm Identification \\with Heterogeneous Clients}

\renewcommand{\shorttitle}{\small Federated Best Arm Identification with Heterogeneous Clients}

\author{Zhirui Chen}
\author{P. N. Karthik}
\author{Vincent Y. F. Tan}
\author{Yeow Meng Chee}

\affil{National University of Singapore
\thanks{Emails: \email{zhiruichen@u.nus.edu}, \email{\{karthik,vtan,ymchee\}@nus.edu.sg}.}}

\begin{document}

\maketitle

\begin{abstract}
We study best arm identification in a federated multi-armed bandit setting with a central server and multiple clients, when each client has access to a {\em subset}  of arms and each arm yields independent Gaussian observations. 
The  goal is to identify the best arm of each client subject to an upper bound on the error probability; here, the best arm is one that has the largest {\em average} value of the means averaged across all clients having access to the arm. Our interest is in the asymptotics as the error probability vanishes. 
We provide an asymptotic lower bound on the growth rate of the expected stopping time of any algorithm. Furthermore, we show that for any algorithm whose upper bound on the expected stopping time matches with the lower bound up to a multiplicative constant ({\em almost-optimal} algorithm), the ratio of any two consecutive communication time instants must be {\em bounded}, a result that is of independent interest. We thereby infer that an algorithm can communicate no more sparsely than at exponential time instants in order to be almost-optimal.
For the class of almost-optimal algorithms, we present the first-of-its-kind asymptotic lower bound on the expected number of {\em communication rounds} until stoppage. We propose a novel algorithm 
that communicates at exponential time instants, and demonstrate that it is asymptotically almost-optimal.
\end{abstract}

\section{Introduction}
\label{sec:introduction}
The problem of best arm identification 
\cite{even2006action, lattimore_szepesvari_2020} 
deals with finding the best arm in a multi-armed bandit as quickly as possible, and falls under the class of optimal stopping problems in decision theory. This problem has been studied under two complementary regimes: (a) the {\em fixed-confidence} regime in which the goal is to minimise the expected time (number of samples) required to find the best arm subject to an upper bound on the error probability \cite{even2006action,jamieson2014lil}, and (b) the {\em fixed-budget} regime in which the goal is to minimise the error probability subject to an upper bound on the number of samples \cite{audibert2010best,bubeck2011pure}. In this paper, we study best arm identification in the fixed-confidence regime.
\subsection{Problem Setup and Objective}
\label{subsec:problem-formulation}
\begin{figure}[!t]
    \centering
    \includegraphics[width=0.75\textwidth]{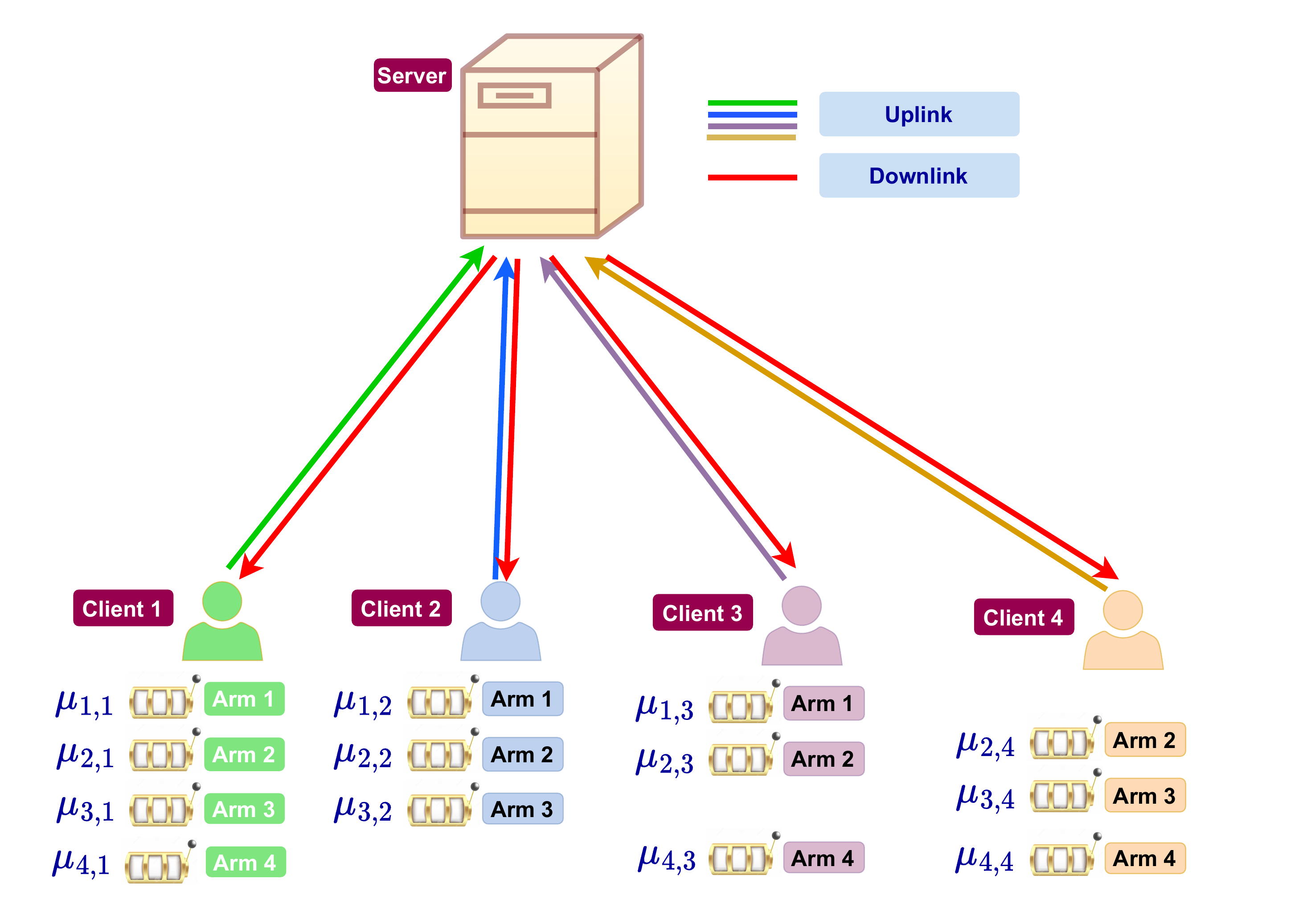}
    \caption{A depiction of the federated learning setup with a central server and $M=4$ clients, each having access to a subset of $K=4$ arms. Client $1$ has access to the subset $S_1=\{1,2,3,4\}$, client $2$ to the subset $S_2=\{1,2,3\}$, client $3$ to the subset $S_3=\{1,2,4\}$, and client $4$ to the subset $S_4=\{2,3,4\}$. We assume that arm $i$ of client $m$ generates independent Gaussian rewards with mean $\mu_{i,m}$ and variance $1$.}
    \label{fig:setup}
\end{figure}
We consider a federated learning  setup \cite{mcmahan2017communication,kairouz2021advances} with a central {\em server} and $M$ {\em clients} in which each client has access to a {\em subset} of arms from a $K$-armed bandit ({\em heterogeneous clients}). For $i \in [K] \coloneqq \{1, \ldots, K\}$ and $m \in [M] \coloneqq \{1, \ldots, M\}$, arm $i$ of client $m$ generates independent Gaussian {\em observations} with mean $\mu_{i,m}$ and unit variance. We assume that the clients do not communicate directly with each other, but instead communicate via the server. 
We let $S_m$ denote the subset of arms accessible by client $m$. For each $i \in S_m$, we let $\mu_i$ denote the average of the values in $\{\mu_{i,m}: i \in S_m\}$. {\color{black} We define the {\em mean reward} of arm $i$ at client $m$ to be $\mu_{i}$, 
which is consistent with other similar works in federated multi-armed bandits \cite{zhu2021federated,shi2021federated,yan2022federated} in which $S_m=[K]$ for every $m$; in our work, we allow for the case when $S_m \subseteq [K]$ for every $m$.}
Defining the {\em best arm} of client $m$ as $\arg\max_{i \in S_m} \mu_i$, the goal is to find the best arms of the clients with minimal expected stopping time, subject to an upper bound on the error probability. Figure \ref{fig:setup} depicts the problem setup pictorially.

{\color{black} Because the mean reward of an arm is the average of the means from all clients having access to the arm, it is necessary for the clients to {\em communicate} with the server in order to determine their individual best arms}.  Intuitively, more frequent communication between the clients and the server implies smaller expected stopping time. Thus, there is a close interplay between (a) frequency of communication, and (b) expected stopping time. {\color{black} Also, intuitively, the smaller the error probability, the larger the expected stopping time}. Our objectives in this paper are two-fold: (i) to provide a rigorous theoretical characterisation of the trade-off between (a) and (b), and (ii) to capture in precise mathematical terms the limiting growth rate of the expected stopping time as the error probability vanishes.


\subsection{Motivating Examples} \label{subsec:motivation}
{\color{black} Systemic biases~\cite[Chapter 6]{kirkup2006introduction} in data are common in federated multi-armed bandit problems, where the performance of an arm can vary significantly across clients due to variations in user behavior and contextual factors~\cite{zhu2021federated}. This poses a significant challenge for selecting the best arm, as the local estimates of the arm's performance may not reflect its true value across all clients. To address this challenge, we propose a definition of the best arm that considers both the local and global performance of each arm. Our proposed definition aggregates the local estimates of each arm's performance, which effectively reduces the overall bias and improves the estimation accuracy. To illustrate the importance of considering the global performance, we provide examples from market survey and democratic elections, where the best arm should be selected based on the average performance across all clients.}

\textbf{Market Survey:} \hspace{0.3\baselineskip} Consider $M$ retailers (clients), each of whom sells products from a subset of $K$ popular brands (arms). 
To determine the best product among their suite of products, suppose that each retailer conducts a market survey and collects consumers' ratings for different products. 
It is possible that different retailers accrue different expected rating scores (representative of $\mu_{i,m}$ in our problem setup) for the same brand (this corresponds to an arm having different means at different clients). For instance, skilled advertising by some retailers may influence customers' liking for certain brands over others, thereby forcing a certain degree of {\em bias} in the customers' ratings for products from such brands. 
Alternatively, in the event when the customers are asked to rate a product as ``good'', ``satisfactory'', ``very good'', etc. in the survey, there is possibility that these ratings are miscalibrated due to subjective differences in the perception of ratings by humans~\cite{wang2019your}.
Thus, for instance, a ``very good'' rating may translate to a numerical score of $4$ for one retailer and to $5$ for another (say, on a scale of $0$ to $5$), depending on how the surveyors of various retailers perceive the customers' ratings. The above scenarios are only a handful examples of {\em systemic} biases in the collected data which make reliable estimation of the true value of a brand based on customer ratings a challenging task. In such scenarios, for each brand, it is logical to {\em average} the ratings across the retailers, and decide the best brand based on the average ratings, given that averaging is a definitive means of reducing the overall systemic bias. Our definition of the best arm (as $\arg\max_{i} \mu_i$, where $\mu_i$ is the {\em average} of $\mu_{i,m}$ values for $i \in S_m$) precisely accomplishes this.

{\color{black}\textbf{Paper reviewing process:} \hspace{0.3\baselineskip}
Another  instance of miscalibration can be observed in the peer review process for academic papers. Consider a scenario where a set of $K$ papers is distributed to $M$ reviewers via an automated system. Each reviewer $m$ is presented with a subset $S_m$ of these papers. However, it is essential to acknowledge that the reviewers vary in terms of their expertise and seniority. Junior reviewers may evaluate the same paper quite differently from their senior counterparts. This disparity in evaluations could arise because junior reviewers tend to focus on different aspects of the papers, such as scrutinizing detailed proofs, while senior reviewers may prioritize assessing the broader impact and significance of the research. Consequently, it is highly probable that the same paper will receive a wide range of scores due to these differing evaluation criteria, and it is this phenomenon that we allude to as ``miscalibration''.   Our proposal for mitigating this miscalibration is to take the {\em average} of the reviewers' scores; this is also commonly done in real-life.}

\textbf{Democratic Elections:} \hspace{0.3\baselineskip} 
In democratic elections involving multiple political parties represented across one or more states, one among a subset of parties (arms) is voted to power within each state (client). Favouritism in election---voting in favour of the party that has a demonstrated record of winning many past elections---is not uncommon; this is akin to voting in favor of party $\arg\max_{i\in S_m}\mu_{i,m}$, where $\mu_{i,m}$ is a collective measure of party $i$'s performance (revenue generated, infrastructural improvements, etc.) as perceived by the people of state $m$. 
However, favouritism in voting is antithetical to the spirit of democracy, and calls for a more careful evaluation of a party's performance by the voters---one that gives a fair chance to non-favourite or new parties to come to power. Indeed, the voters of state $m$ may want to exercise votes in favour of a non-favorite party that is seemingly performing well in {\em other states}, with the hope that it will deliver a similar (or better) performance in state $m$. Our proposal of evaluating party $i$'s performance via $\mu_i$ (by averaging the performances across states) and voting in favour of party $\arg\max_{i} \mu_i$ precludes favouritism and supports voting in favour of the party that shows the greatest potential for performance overall.

\subsection{Contributions}
\label{subsec:contributions}
We now bring out the main contributions of this paper {\color{black} and highlight the challenges in the analysis}.
\begin{itemize}[leftmargin=*]
    \item We derive a problem instance-specific asymptotic lower bound on the expected stopping time (i.e., the time required to find the best arms of the clients). As in the prior works on best arm identification \cite{garivier2016optimal,moulos2019optimal}, we show that given an error probability threshold $\delta\in (0,1)$, the lower bound scales as $\Omega(\log (1/\delta))$ (all logarithms  are natural logarithms). We   characterise the instance-dependent constant multiplying $\log(1/\delta)$. This constant, we show, is the solution to a max-inf optimisation problem in which the outer `max' is over all probability distributions on the arms and the inner `inf' is over the set of alternative problem instances, and is a measure of the ``hardness'' of the instance. 
	\item The max-inf optimisation in the instance-dependent constant is seemingly hard to solve analytically. The hardness stems from set of alternative problem instances in the inner inf not admitting a closed-form expression, unlike in the prior works where simple closed-form expressions for the set of alternative instances exist. Notwithstanding this, we recast the inf over the (uncountably infinite) set of alternative problem instances as a min over the (finite) set of arms, and demonstrate that the max-min optimisation resulting from the latter can be solved analytically and differs from the true max-inf by at most a factor of $2$. 
	\item For any algorithm whose upper bound on the expected time to find the best arms of the clients matches the lower bound up to a multiplicative constant (an {\em almost-optimal algorithm}), we show that the ratio of any two consecutive communication instants {\em must be bounded}, a result that is of independent interest. That is, in order to achieve order-wise optimality in the expected time to find the best arms, an algorithm may communicate at most exponentially sparsely, e.g., at communication time instants of the form $t=\lceil (1+\lambda)^r \rceil $, $r\in\mathbb{N}$, for some $\lambda>0$. In this sense, the class of all algorithms communicating exponentially sparsely (with different exponents) forms the {\em boundary} of the class of almost-optimal algorithms. Using this result, we show that given any error probability $\delta$, there exists a sequence of problem instances with increasing hardness levels on which the expected number of communication rounds until stoppage grows with $\delta$ as $\Omega(\log \log (1/\delta))$ for any algorithm with a bounded ratio between consecutive communication time instants. This is the first-of-its kind result in the literature.
	
	\item We design a {\em Track-and-Stop}-based algorithm, called \underline{\textit{Het}}erogeneous \underline{\textit{T}}rack-and-\underline{\textit{S}}top and abbreviated as $\textsc{Het}\mhyphen\textsc{TS}(\lambda)$, that communicates only at exponential time instants of the form $t=\lceil (1+\lambda)^r \rceil $, $r\in\mathbb{N}$, for an input parameter $\lambda>0$. We show that given any $\delta\in (0,1)$, the $\textsc{Het}\mhyphen\textsc{TS}(\lambda)$ algorithm (a) identifies the best arms correctly with probability at least $1-\delta$, (b) is asymptotically almost-optimal up to the constant $2\, (1+\lambda)$,
	and (c) takes $O(\log \log (1/\delta))$ many communication rounds on the average. Here, $\lambda$ serves as a tuning parameter to trade-off between the expected number of communication rounds  and the expected stopping time. 
\end{itemize}

\subsection{Related Works}
\label{subsec:related}
\textbf{Federated bandits:} Best arm identification in the fixed-confidence regime for independent and identically distributed (i.i.d.) observations has been studied in \cite{garivier2016optimal,Kaufmann2016}. The recent works \cite{moulos2019optimal, karthik2022best} extend the results of \cite{garivier2016optimal} to the setting of Markov observations from the arms. {\color{black} 
The problem of multi-armed bandits in federated learning has been studied in several recent works, including those with similar setups to our own \cite{shi2021federated,zhu2021federated,yang2022optimal,yan2022federated,shi2021federatedtriple,dubey2020differentially}. These works are generally classified as belonging to the class of ``federated bandit'' problems, first proposed by \cite{shi2021federated}  in which  each client has access to {\em all} the arms.}\ The notion of {\em global mean} defined in these works coincides with our definition of {\em mean reward} (i.e., average of the arm means across the clients), and the  goal is to design an algorithm that minimises the cumulative {\em regret} over a finite time horizon of $T$ time units. The paper \cite{mitra2021exploiting} studies best arm identification in a federated learning setting in which each client has access to a subset of arms that is disjoint from the arms subsets of the other clients, and the clients coordinate with each other to find the overall best arm (the arm with the largest mean); notice that in this setting, the best arm is necessarily the best arm of one of the clients. 
In our work, we allow for non-disjoint subsets of arms across the clients, and the best arm of one client may not necessarily be the best arm of another. The paper \cite{reddy2022almost} studies an optimal stopping variant of the problem in \cite{shi2021federated} in which the uplink from each client to the server entails a fixed cost of $C\geq 0$ units, each client has access to {\em all} the arms, and the goal is to determine the arm with the largest mean at each client and also the arm with the largest {\em global mean} with minimal total cost, defined as the sum of the number of arm selections and the total communication cost. 
When each client has access to all the arms, our problem distils down to finding the arm with the largest global mean. 

{\color{black}
\noindent
\textbf{Collaborative bandits:} Another line of related works goes collectively by the name of {\em collaborative bandits}~\cite{tao2019collaborative, karpov2020collaborative,hillel2013distributed}. Here, each agent within a set of agents is capable of identifying the best arm in a {\em single} bandit environment without communication; the analytical task then is to quantify how much communication aids in reducing the overall sample complexity. In contrast, in our work, communication is clearly necessary to estimate the mean of the global best arm. This is the essential difference between our setting and that of collaborative bandits. Hillel et al.~\cite{hillel2013distributed} initially carry out a study on pure exploration within the collaborative bandit framework. They demonstrate that a single communication round among agents is sufficient to identify the best arm efficiently. Tao {\em et al.}~\cite{tao2019collaborative} quantifies the power of collaboration under limited interaction (or, communication steps), as interaction is expensive in many settings. They measure the running time of a distributed algorithm as the speedup over the the best centralized algorithm where there is only one agent.   Karpov {\em et al.}~\cite{karpov2020collaborative} study the problem of top-$m$ arms identification in both fixed budget and fixed confidence cases under the setting of collaborative bandits. More recently, Karpov and Zhang~\cite{karpov2022collaborative2} studied the problem of {\em fixed-budget} best arm identification in  collaborative bandits on non-i.i.d.\ data. This bears similarities to the study of federated bandits. However, we tackle the problem of fixed-confidence best arm identification on a novel problem setting in which each client only has access to a possibly strict subset of arms (cf.\ Fig.~\ref{fig:setup}).


\textbf{Bandits with communication constraints}: Additionally, some other studies take into account the effect of {\em   communication constraints}, which is  related to our work as communication or privacy constraints are often times incorporated into the federated learning setting. 
Hanna {\em et al.}~\cite{hanna2022solving} consider a bandit model in which the learner receives the reward value via a communication channel.  Their findings indicate that a communication rate of $1$-bit per time step is sufficient to achieve near-optimal regret when rewards are bounded. 
Mitra {\em et al.}~\cite{mitra2023linear}  generalize this framework to the linear bandit setting.  They show that a bit rate (i.e., number of bits  per time step)  linear in the dimension of the unknown parameter vector suffices to achieve near-optimal regret.  Pase {\em et al.}~\cite{pase2022rate} consider the  Bayesian regret  and   demonstrate that to achieve sublinear regret, the  bit rate   needs to exceed $H(A^*)$,  the entropy of the marginal distribution of the arm pulls under the optimal strategy. In addition, the model presented in Wang {\em et al.}~\cite{wang2019distributed} considers the {\em total number of bits} instead of the bit rate. They demonstrate that to achieve near-optimal regret, the total number of bits has to depend  logarithmically on the horizon $T$. Mayekar {\em et al.}~\cite{mayekar2023} recently analyzed communication-constrained bandits under additive Gaussian noise and showed that the regret depends on the capacity of the channel.   Although these studies~\cite{hanna2022solving, mitra2023linear, pase2022rate,wang2019distributed, mayekar2023}   contribute towards understanding bandits with communication constraints, their focus is primarily on the number of bits of transmission in the communication channel. In contrast, our research focuses on the {\em frequency  of transmission} in the communication channels from the clients to the server.
}
\section{Notations and Preliminaries}\label{sec:problem-setup}

For $n\in \mathbb{N}\coloneqq \{1, 2, \ldots\}$, we let $[n]\coloneqq \{1,\dots,n\}$.
We consider a federated multi-armed bandit with $K$ arms, a central server, and $M$ clients, in which each client has access to a subset of arms. For $m\in [M]$, let $S_m$ denote the subset of arms accessible by client $m$. 
Without loss of generality, we assume that $\lvert S_m \rvert \ge 2$ for all $m$.
Pulling arm $i\in S_m$ at time $t\in \mathbb{N}$ generates the observation $X_{i,m}(t)$ that is Gaussian distributed with mean $\mu_{i,m}\in \mathbb{R}$ and unit variance.  A {\em problem instance} $v=(\{\mu_{i,1}\}_{i\in S_1}, \{\mu_{i,2}\}_{i\in S_2},\dots, \{\mu_{i,M}\}_{i\in S_M})$ is defined by the collection of the means of the arms in each client's set of accessible arms. For any $i\in [K]$, we define the {\em reward} $X_i(t)$ of arm $i$ as the average of the observations obtained at time $t$ from all clients $m$ for which $i\in S_m$, i.e., $X_i(t) \coloneqq \frac{1}{M_i} \sum_{m=1}^M \mathbf{1}_{\{i \in S_m\}} X_{i,m}(t) $, where $M_i \coloneqq \sum_{m=1}^M \mathbf{1}_{\{i \in S_m\}}$ is the number of clients that have access to arm $i$. We let $\mu_i= \frac{1}{M_i} \sum_{m=1}^M \mathbf{1}_{\{i \in S_m\}} \mu_{i,m}$ denote the mean reward of arm $i$. Arm $i$ is said to be the {\em best arm} of client $m$ if it has the largest mean reward among all the arms in $S_m$. We assume that each client has a {\em single} best arm,  and we let $a_m^* := \arg\max_{i\in S_m} \mu_i$ denote the best arm of client $m$. We let $a^*\coloneqq (a_m^*)_{m\in [M]}\in S_1\times S_2\times\ldots \times S_M$ denote the tuple of best arms. More explicitly, we write $a^*(v)$ to denote the tuple of best arms under the problem instance $v$, and let $\mathcal{P}$ be the set of the all problem instances with a single best arm at each client.

We assume that the clients and the central server are time-synchronised and that the clients communicate with the server at certain pre-defined time instants. 
Given $(M,K,\{S_m\}_{m=1}^M)$, a problem instance $v$, and a confidence level $\delta\in (0,1)$, we wish to design an algorithm for finding the best arm of each client with (a) the fewest number of time steps and communication rounds, and (b) error probability less than $\delta$. 
By an algorithm, we mean a tuple $\Pi=(\Pi_{\rm comm},\Pi_{\rm cli},\Pi_{\rm svr})$ of (a) a strategy for {\em communication} between the clients and the server, (b) a strategy for selection of arms at each {\em client}, and (c) a combined stopping and recommendation rule at the {\em server}. The communication strategy $\Pi_{\rm comm}$
%
consists of the following components: (a) $\{b_r\}_{r \in \mathbb{N}}$: the time instants of communication, with $b_r \in \mathbb{N}$ and $b_r \le b_{r+1}$ for all $ r \in \mathbb{N}$, (b) $\Sigma$: the set of values transmitted from the server to each client, (c) $\Phi$: the set of values transmitted from each client to the server; this is assumed to be identical for all the clients, (d) $\hbar_r:{\Phi}^{Mr} \to \Sigma$:  a function deployed at the server, which aggregates the information transmitted from all the clients in the communication rounds $1,\ldots,r$, and generates an output value to be transmitted to each client, and (e) $\rho_r^m: (\mathbb{R}\times S_m) ^{b_r} \to \Phi$: a function deployed at client $m\in [M]$, which aggregates the observations seen by client $m$ in the time instants $1,\ldots, b_r$ from the arms in $S_m$, and generates an output value to be transmitted to the server.

The arms selection strategy $\Pi_{\rm cli}$
consists of component arm selection rules $\pi_t^m: (\mathbb{R}\times S_m) ^{t} \times \Sigma \rightarrow S_m$, $m\in [M]$. Here, $\pi_t^m$ takes as input the observations seen from the arms in $S_m$ pulled by client $m$ up to time $t$ and the information received from the server to decide which arm in $S_m$ to pull at time $t+1$.
Lastly, the stopping and recommendation strategy $\Pi_{\rm svr}$
at the server consists of the following components: (a) the stopping rule $\Upsilon_r: \Phi^{Mr} \to \{0,1\}$ that decides whether the algorithm stops in the $r$-th  communication round  and (b) the recommendation rule $\Psi_r: \Phi^{Mr} \rightarrow S_1 \times S_2 \times \cdots \times S_M$ to output the empirical best arm of each client if the algorithm stops in the $r$th communication round. We let $\hat{a}_{\delta,m}$ denote the empirical best arm of client $m$ output by the algorithm under confidence level $\delta$, and define $\hat{a}_\delta\coloneqq (\hat{a}_{\delta,m})_{m\in [M]}$.

\noindent
We assume that all the functions defined above are Borel-measurable. 
Note that if an algorithm stops in the $r$th communication round, then its stopping time $\tau=b_r$. Given $\delta\in (0,1)$, we say that an algorithm $\Pi$ is {\em $\delta$-probably approximately correct} (or $\delta$-PAC) if 
$\mathbb{P}_v^{\Pi}\left(\tau_\delta < +\infty \right)=1$ and  
$\mathbb{P}_v^{\Pi}\left( \hat{a}_\delta \neq a^*(v) \right) \le \delta$ for any problem instance $v\in \mathcal{P}$; here, $\mathbb{P}_v^\Pi(\cdot)$ the probability measure induced by the algorithm $\Pi$ and the problem instance $v$. Writing $\tau_\delta(\Pi)$ and $\mathfrak{r}_\delta(\Pi)$ to denote respectively the stopping time and the associated number of communication rounds corresponding to the confidence level $\delta$ under the algorithm $\Pi$, our interest is in the following   optimisation problems:
\begin{equation}
\inf_{\Pi \text{ is }\delta\text{-PAC}}\ \mathbb{E}_v^\Pi[\tau_\delta(\Pi)], \quad \inf_{\Pi \text{ is }\delta\text{-PAC}}\ \mathbb{E}_v^\Pi[\mathfrak{r}_\delta(\Pi)].
\label{eq:optimisation-problems}
\end{equation}
In \eqref{eq:optimisation-problems}, $\mathbb{E}_v^\Pi$ denotes expectation with respect to the measure $\mathbb{P}_v^\Pi$. Prior works \cite{Kaufmann2016,moulos2019optimal} show that the first term in \eqref{eq:optimisation-problems} grows as $\Theta(\log(1/\delta))$ as $
\delta \to 0$. We anticipate that a similar growth rate holds for our problem setting. Our objective is to precisely characterise
\begin{equation}
\liminf_{\delta \to 0}\ \inf_{\Pi \text{ is }\delta\text{-PAC}}\ \frac{\mathbb{E}_v^\Pi[\tau_\delta(\Pi)]}{\log(1/\delta)}.
\label{eq:objective-1}
\end{equation}
In the following section, we present a lower bound for \eqref{eq:objective-1}. Furthermore, we demonstrate that on any sequence of problem instances $\{v^{(l)}\}_{l=1}^{\infty}$ with increasing levels of ``hardness'' (to be made precise soon), the second term in \eqref{eq:optimisation-problems} grows as $\Theta(\log \log (1/\delta))$ and obtain a precise characterisation of this growth rate, the first-of-its-kind result in the literature to the best of our knowledge.
\section{Lower Bound: Converse}
Below, we first derive a problem-instance specific asymptotic lower bound on the expected stopping time. Then, we present a simplification to the constant appearing in the lower bound and provide the explicit structure of its optimal solution. Next, we show that for any algorithm to be  almost-optimal (in the sense to be made precise later in this section), the ratio of any two consecutive communication time instants must be bounded, a result that may be of independent interest. Using this result, we obtain an $O(\log \log (1/\delta))$ lower bound on the expected number of communication rounds for a sub-class of $\delta$-PAC algorithms.
\subsection{Lower bound on the Expected Stopping Time}
Let ${\rm Alt}(v) \coloneqq \{v' \in \mathcal{P}: a^*(v)\ne a^*(v') \}$ denote the set of alternative problem instances corresponding to the problem instance~$v$. Let $\Lambda$ denote the simplex of probability distributions on $K$ variables, and let $\Lambda_m \coloneqq \{\alpha \in \Lambda: \alpha_i=0 \ \forall\, i\notin S_m\}$ denote the subset of $\Lambda$ corresponding to client $m\in[M]$. We write $\Gamma \coloneqq \Lambda_1 \times \cdots \times \Lambda_M$ to denote the Cartesian product of $\{\Lambda_m\}_{m=1}^{M}$.
The following proposition presents the first main result of this paper. 
\begin{proposition}
\label{prop:lower-bound}
For any $v\in \mathcal{P}$ and $\delta \in (0,1)$,
\begin{equation}
\inf_{\Pi\text{ is }\delta\text{-PAC}}\ \mathbb{E}_v^\Pi[\tau_{\delta}(\Pi)] \ge c^*(v) \log \left(\frac{1}{4\delta} \right),
\label{eq:lower-bound}
\end{equation}
where the constant $c^*(v)$ is given by
\begin{align}
\label{eq:c-star-v} 
\begin{split}\hspace{-.2in}
 c^*(v)^{-1} = 
 \max_{\omega \in \Gamma}\  \inf_{v'\in {\rm Alt}(v)}\! \sum_{m=1}^{M} \sum_{i\in S_m}\!  \omega_{i,m}\frac{\big(\mu_{i,m}(v)\!-\!\mu_{i,m}(v')\big)^2}{2}.
\end{split}
\end{align}
Dividing both sides by $\log(1/\delta)$ and letting $\delta \to 0$ in \eqref{eq:lower-bound},
\begin{equation}
\liminf_{\delta \to 0}\ \inf_{\Pi \text{ is }\delta\text{-PAC}}\ \frac{\mathbb{E}_v^\Pi[\tau_{\delta}(\Pi)]}{\log (1/\delta)} \geq c^*(v).
\label{eq:lower-bound-asymptotic}
\end{equation}
\end{proposition}
The term $c^*(v)$ defined in \eqref{eq:c-star-v} is a measure of the ``hardness'' of the instance $v$ and is the solution to a max-inf optimisation problem where the outer `max' is over all $M$-ary probability distributions $\omega \in \Gamma$ such that $\sum_{i \in S_m} \omega_{i,m} =1$ for all $m \in [M]$ (here, $\omega_{i,m}$ is the probability of pulling arm $i$ of client $m$), and the inner `inf' is over the set of alternative problem instances corresponding to the instance $v$. The proof of Proposition~\ref{prop:lower-bound} is similar to the proof of \cite[Theorem 1]{garivier2016optimal} and is omitted for brevity. The key ideas in the proof to note are (a) the transportation lemma of \cite[Lemma 1]{Kaufmann2016} relating the error probability to the expected number of arm pulls and the Kullback--Leibler divergence between two problem instances $v$ and $v'\in {\rm Alt}(v)$ with distinct best arm locations, and (b) Wald's identity for  i.i.d.\ observations. 

\subsection{A Simplification} \label{sec:simplified}
 A close examination of the proof of the lower bound in  \cite{garivier2016optimal} reveals that an important step in the proof therein is a further simplification of the max-inf optimisation in the instance-dependent constant; see \cite[Theorem 5]{garivier2016optimal}. However, an analogous simplification of \eqref{eq:c-star-v} is not possible as  ${\rm Alt}(v)$ does not admit a closed-form expression.
 Nevertheless, we propose the following simplification. For any $\omega \in \Gamma$ and instance $v$, let
\begin{equation}
g_v(\omega) \coloneqq \inf_{v'\in {\rm Alt}(v)} \sum_{m=1}^{M} \sum_{i \in S_m}  \omega_{i,m}\frac{(\mu_{i,m}(v)-\mu_{i,m}(v'))^2}{2}
\label{eq:inner-minimum-term}
\end{equation}
denote the inner minimum in \eqref{eq:c-star-v}. Our simplification of \eqref{eq:inner-minimum-term} is given by
\begin{equation}
\widetilde{g}_v(\omega) \coloneqq \min_{i \in [K]} \frac{\Delta_i^2(v)/2}{\frac{1}{M^2_i}\sum_{m=1}^M  \mathbf{1}_{\{i \in S_m\}}\frac{1}{\omega_{i,m}} }, 
\label{eq:simplified-inner-minimum-term}
\end{equation}
where for each $i\in [K]$, 
\begin{equation} 
\Delta_i(v) \coloneqq \min_{m\in [M]:\, i\in S_m} \Big\lvert \mu_i(v) - \max_{j \in S_m \setminus \{i\}}\mu_{j}(v) \Big\rvert. 
\label{eq:definedelta}
\end{equation}
In particular, if $\omega_{i,m}=0$ for some $m\in[K]$ and $i\in S_m$, then $\widetilde{g}_v(\omega)=0$.
Notice that the infimum in \eqref{eq:inner-minimum-term} is over the {\em uncountably infinite} set ${\rm Alt}(v)$, whereas the simplified minimum in \eqref{eq:simplified-inner-minimum-term} is over the {\em finite} set $[K]$. Our next result shows that these two terms differ at most by a factor of $2$.
\begin{lemma}
\label{lemma:relaxed-g}
For $v\in \mathcal{P}$ and $\omega \in \Gamma$, let $g_v(\omega)$ and $\widetilde{g}_v(\omega)$ be as defined in \eqref{eq:inner-minimum-term} and \eqref{eq:simplified-inner-minimum-term} respectively. Then,
$\frac{1}{2}\widetilde{g}_v(\omega)   \le g_v(\omega)  \le  \widetilde{g}_v(\omega)$.
\end{lemma}
As a consequence of Lemma \ref{lemma:relaxed-g}, it follows that $\widetilde{c}(v)\coloneqq \max_{\omega \in \Gamma} \widetilde{g}_v(\omega)$ and $c^*(v)=\max_{\omega \in \Gamma} g_v(\omega)$ differ only by a multiplicative factor of $2$. It is not clear if the optimiser of $c^*(v)$, if any, can be computed analytically. 
On the other hand, as we shall soon see, the optimiser of $\widetilde{c}(v)$ can be computed in closed-form and plays an important role in the design of an asymptotically almost-optimal algorithm. 

\begin{definition}[balanced condition]
\label{def:balanced-condition}
An $\omega \in \Gamma$ satisfies {\em balanced condition} if  $\frac{\omega_{i_1,m_1}}{\omega_{i_2,m_1}}  = \frac{\omega_{i_1,m_2}}{\omega_{i_2,m_2}}$ for all~$i_1, i_2 \in S_{m_1} \cap S_{m_2}$ and $m_1, m_2 \in [M]$.
\end{definition}
That is, $\omega$ satisfies {\em balanced condition} if the ratios of the arm selection probabilities are consistent (or {\em balanced}) across the clients. The next result shows that $\max_{\omega \in \Gamma} \widetilde{g}_v(\omega)$ admits a solution that satisfies {\em balanced condition}.

\begin{proposition}
\label{proposition-meet-condition-1}
Given $v\in \mathcal{P}$, there exists $\omega \in \Gamma$ that attains the maximum in the expression for $\widetilde{c}(v)$ and satisfies { balanced condition}.
\end{proposition}
Proposition \ref{proposition-meet-condition-1} follows in a straightforward manner from a more general result, namely Theorem \ref{theorem:unique-allocation}, which we state later in the paper. 
\begin{corollary}
\label{cor:exact-form-of-optimal-solution}
Let $\widetilde{\omega}(v)\in \Gamma$ be any $M$-ary probability distribution that attains the maximum in the expression for $\widetilde{c}(v)$ and satisfies balanced condition. Then, there exists a $K$-dimensional vector $G(v)=[G(v)_i]_{i\in [K]}$ such that 
\begin{equation}
\widetilde{\omega}(v)_{i,m} = \frac{G(v)_i}{\sum_{i' \in S_m} G(v)_{i'}}, \quad  i \in S_m, \ m\in [M].
\label{eq:G-characterises-omega-tilde}
\end{equation}
\end{corollary}
Corollary \ref{cor:exact-form-of-optimal-solution} elucidates the rather simple form of the optimiser of $\widetilde{c}(v)$, one that is characterised by a $K$-dimensional vector $G(v)$ which, in the sequel, shall be referred to as the {\em global vector} corresponding to the instance $v$. We shall soon see that it plays an important role in the design of an almost-optimal algorithm for finding the best arms of the clients. In fact, we show that in order to inform each client of its arm selection probabilities, the server needs to broadcast only the global vector instead of sending a separate probability vector to each client, thereby leading to significantly less downlink network traffic, {\color{black} especially when $M$
is large. For example, using a broadcast-type protocol instead of a unicast-type protocol such as user datagram protocol (UDP) for transmitting data from the server to clients over the internet is known to reduce the network traffic significantly~\cite[Chapter 20]{unix}.}

\subsection{Lower bound on the Expected Number of Communication Rounds}
In this section, we present a lower bound on the expected   number of communication rounds required by any ``good'' algorithm to find the best arms of the clients. By ``good'' algorithms, we mean the class of all {\em almost-optimal} $\delta$-PAC algorithms as defined below.

\begin{definition}[almost-optimal algorithm]
\label{defn:almost-optimal-algo} Given  $\delta\in(0,1)$, and $\alpha\ge1$, a $\delta$-PAC algorithm $\Pi$ is said to be {\em almost-optimal} up to a constant $\alpha$ if
\begin{equation}
 \mathbb{E}_v^\Pi[\tau_{\delta}(\Pi)] \le \alpha \, c^*(v) \log\left(\frac{1}{4\delta}\right) \quad \forall v \in \mathcal{P}.
\label{eq:almost-optimal-algo}
\end{equation}
In addition, $\Pi$ is said to be almost-optimal if it is almost-optimal up to a constant $\alpha$ for some $\alpha\ge1$.
\end{definition} 

Definition \ref{defn:almost-optimal-algo} implies that the expected stopping time of an almost-optimal algorithm matches the lower bound in \eqref{eq:lower-bound} up to the multiplicative constant $\alpha$. Notice that the sparser (more infrequent) the communication between the clients and the server, the larger the time required to find the best arms of the clients. Because \eqref{eq:almost-optimal-algo} implies that the expected stopping time of an almost-optimal algorithm cannot be infinitely large, it is natural to ask what is the sparsest level of communication achievable in the class of almost-optimal algorithms. The next result provides a concrete answer to this question.
{\color{black}
\begin{theorem}
\label{lemma:upper_bound_on_comm_rouds}
Fix $\delta\in(0,\frac{1}{4})$ and a $\delta$-PAC algorithm $\Pi$ with communication time instants $\{b_r\}_{r\in \mathbb{N}}$. If $\Pi$ is almost-optimal, then
$
    \sup_{r \in \mathbb{N} } \frac{b_{r+1}}{b_{r}} < +\infty.
$ 
\end{theorem}
}
Theorem \ref{lemma:upper_bound_on_comm_rouds}, one of the key results of this paper and of independent interest, asserts that the ratio of any two consecutive communication time instants of an almost-optimal algorithm {\em must be bounded}. An important implication of Theorem~\ref{lemma:upper_bound_on_comm_rouds} is that an almost-optimal algorithm can communicate at most exponentially sparsely, i.e., at exponential time instants of the form $t=\lceil (1+\lambda)^r \rceil $, $r\in \mathbb{N}$, for some $\lambda>0$. For instance, an algorithm that communicates at time instants that grow {\em super-exponentially} (i.e., $t=2^{\kappa(r)}$ for any super-linear function $\kappa(r)$), does not satisfy the requirement in Theorem \ref{lemma:upper_bound_on_comm_rouds}, and hence cannot be almost-optimal. In this sense, the class of all exponentially sparsely communicating algorithms (with different exponents) forms the {\em boundary} of the class of all almost-optimal algorithms.
%


The proof of Theorem~\ref{lemma:upper_bound_on_comm_rouds} suggests that when an almost-optimal algorithm $\Pi$ stops at time step $\tau_\delta(\Pi)$ and {\color{black} $\sup_{r\in \mathbb{N}} \frac{b_{r+1}}{b_{r}} \le \eta$}, at least $\Omega(\log_{\eta}(\tau_\delta(\Pi)))$ 
communication {\em rounds} must have occurred, i.e.,  $\mathfrak{r}_\delta(\Pi)=\Omega(\log_\eta(\tau_\delta(\Pi)))$ almost surely (a.s.). The next result relates $\log_\eta(\tau_\delta(\Pi))$ with $\log_\eta(\mathbb{E}[\tau_\delta(\Pi)])$.

\begin{lemma}
\label{lemma:log-expectation-guarantee}
Let $\{v^{(l)}\}_{l=1}^\infty \subset \mathcal{P}$ be any sequence of problem instances with $\lim_{l\rightarrow \infty} c^*(v^{(l)}) = +\infty$. Given $\delta\in \left(0,\frac{1}{4}\right)$, for any almost-optimal algorithm $\Pi$ and $\beta\in(0,1)$,
\begin{equation*}
\liminf\limits_{l \rightarrow \infty} \mathbb{P}_{v^{(l)}}^{\Pi}\left(\log\left(\tau_{\delta}(\Pi) \right) \!>\! \beta \log\left(\mathbb{E}_{v^{(l)}}^\Pi[\tau_{\delta}(\Pi)] \right) \right)  \!\ge\! \frac{1}{4} -\delta.
\label{eq:log-expectation-guarantee}
\end{equation*}
\end{lemma}
Lemma~\ref{lemma:log-expectation-guarantee} shows that $\log(\tau_{\delta}(\Pi)) = \Omega(\log(\mathbb{E}_{v^{(l)}}^\Pi[\tau_{\delta}(\Pi)])$ with a non-vanishing probability on a sequence of problem instances $v^{(l)}$ with increasing hardness levels. Proposition \ref{prop:lower-bound} implies that $\mathbb{E}_{v^{(l)}}^\Pi[\tau_{\delta}(\Pi)]=\Omega(\log(1/\delta))$, which in conjunction with Lemma~\ref{lemma:log-expectation-guarantee} and the relation $\mathfrak{r}_\delta(\Pi)=\Omega(\log_\eta(\tau_\delta(\Pi)))$ a.s., yields $\mathfrak{r}_\delta(\Pi)=\Omega(\log_\eta \log(1/\delta))$ a.s., and consequently $\mathbb{E}[\mathfrak{r}_\delta(\Pi)]=\Omega(\log_\eta \log(1/\delta))$. The next result makes this heuristic precise.
\begin{theorem}
\label{theorem:comm_complexity}
 Fix $\{v^{(l)}\}_{l=1}^\infty \subset \mathcal{P}$ with $\lim_{l\rightarrow \infty} c^*(v^{(l)}) = +\infty$. Fix $\delta\in(0,\frac{1}{4})$. For any almost-optimal algorithm $\Pi$ with communication time instants $\{b_r\}_{r\in \mathbb{N}}$ satisfying $\frac{b_{r+1}}{b_r}\le \eta$ for all $ r\in \mathbb{N}$,  
\begin{equation}
\liminf_{l\to\infty}\ \frac{\mathbb{E}_{v^{(l)}}^\Pi[\mathfrak{r}_\delta(\Pi)]}{\log_\eta \left(\, c^*(v^{(l)})\log\left(\frac{1}{4\delta}\right)\right)} \ge \frac{1}{4} -\delta.
\label{eq:lower-bound-communication-round-for-each-delta}
\end{equation}

\end{theorem}
Theorem \ref{theorem:comm_complexity} is the analogue of Proposition \ref{prop:lower-bound} for the number of communication rounds, and is the first-of-its-kind result to the best of our knowledge.

{\color{black}\begin{remark} \label{rmk:alpha}
A natural desideratum  in the lower bound on the expected number of communication rounds would be that it depends explicitly on $\alpha$ for $\delta$-PAC algorithms $\Pi$ that are almost optimal up to a constant $\alpha$ (the multiplicative gap from the lower bound $c^*(v)\ \log\big(\frac{1}{\delta}\big)$ as per Definition~\ref{defn:almost-optimal-algo}). However, we see that Theorem~\ref{theorem:comm_complexity} is expressed in terms of $\eta$, a bound on the ratio between successive communication rounds $\frac{b_{r+1}}{b_r}$. Intuitively, it should hold that $\eta$ is monotonically increasing in  $\alpha$. However, our proof strategies to establish the lower bounds in Theorems~\ref{lemma:upper_bound_on_comm_rouds} and~\ref{theorem:comm_complexity} are not amenable to   elucidate the  explicit dependence of $\mathbb{E}_{v}^\Pi[\mathfrak{r}_\delta(\Pi)]$ on  ~$\alpha$ for algorithms $\Pi$ that are asymptotically optimal up to constant~$\alpha$. We will see, however, that our algorithm $\textsc{Het}\mhyphen\textsc{TS}(\lambda)$ to be introduced in the next section makes this dependence explicit; see Theorem~\ref{Theorem:mainresult} where $\alpha$ is roughly $2\eta$.
\end{remark}}

\section{The Heterogeneous Track-and-Stop (\texorpdfstring{$\textsc{Het}\mhyphen \textsc{TS}(\lambda)$}{Het-TS(lambda)}) Algorithm}
\label{sec:algorithm-for-best-arm-identification}
In this section, we propose an algorithm for finding the best arms of the clients based on the well-known {\em Track-and-Stop} strategy \cite{garivier2016optimal,Kaufmann2016} that  communicates exponentially sparsely. Known as \underline{\em Het}erogeneous \underline{\em T}rack-and-\underline{\em S}top and abbreviated as $\textsc{Het}\mhyphen\textsc{TS}(\lambda)$ for an input parameter $\lambda>0$, the individual components of our algorithm are described in detail below. 

\paragraph{\bf Communication strategy:} We set $b_r = \lceil (1+\lambda)^r \rceil$, $r\in \mathbb{N}$. In the $r$th communication round, each client sends to the server the empirical means of the observations seen from its arms up to time $b_r$. Note that
\begin{equation}
\hat{\mu}_{i,m}(t) \coloneqq \frac{1}{N_{i,m}(t)}\sum_{s=1}^{t} \mathbf{1}_{\{A_{m}(s) = i\}}X_{i,m}(s) 
\label{eq:empirical-mean}
\end{equation}is the empirical mean of  $i\in S_m$ after $t$ time instants, where $\hat{\mu}_{i,m}(t)=0$ if $N_{i,m}(t)=0$. In~\eqref{eq:empirical-mean}, $A_m(t)$ is the arm pulled by client $m$ at time $t$, and $N_{i,m}(t):= \sum_{s=1}^{t} \mathbf{1}_{\{A_{m}(s) = i\}}$ is the number of times arm $i$ of client $m$ was pulled up to time $t$. On the downlink, for each $t\in \{b_r\}_{  r\in \mathbb{N} }$, the server first computes the global vector $G(\hat{v}(t))$ according to the procedure outlined in Section~\ref{sec:optimal-allocation}   and broadcasts this vector to each client. Here, $\hat{v}(t)$ is the empirical problem instance at time $t$, defined by the empirical arm means $\{\hat{\mu}_{i,m}(t): i\in S_m, m\in [M]\}$ received from the clients. In particular, $G(\hat{v}(t))=\mathbf{1}^K$ if $\hat{v}(t) \notin \mathcal{P}$, where $\mathbf{1}^K$ denotes the all-ones vector of length $K$.

\paragraph{\bf Sampling strategy at each client:} We use a variant of the so-called {\em D-tracking} rule of \cite{garivier2016optimal} for pulling the arms at each of the clients. Accordingly, at any time $t$, client $m\in [M]$ first computes
\begin{equation}
\hat{\omega}_{i,m}(t) \coloneqq \frac{G(\hat{v}(b_{r(t)}))_i}{\sum_{i'\in S_m} G(\hat{v}(b_{r(t)}))_{i'}}, \quad  i\in S_m,
\label{eq:sampling-rule-at-client-m}
\end{equation}
based on the global vector received from the server in the most recent communication round $r(t) \coloneqq \min \{r\in \mathbb{N}: b_r \ge t\}-1$ (with $b_0 \coloneqq 0$), and subsequently pulls arm 
\begin{align}{\footnotesize 
A_m(t) \!\in\! \begin{cases}
\displaystyle\argmin_{i\in S_m} N_{i,m}(t\!-\!1), &\hspace{-.73in} \displaystyle\min_{i\in S_m} N_{i,m}(t\!-\!1)\! <\! \sqrt{\frac{t\!-\! 1}{\lvert S_m\rvert}},\\
\displaystyle \arg\min_{i\in S_m} N_{i,m}(t-1)- t\,\hat{\omega}_{i,m}(t) , &\text{otherwise}.
\end{cases}} \label{eq:D-tracking-sampling-rule}
\end{align}
Ties, if any, are resolved uniformly at random. Notice that the rule in \eqref{eq:D-tracking-sampling-rule} ensures that in the long run, each arm is pulled at least $O(\sqrt{t})$ many times after $t$ time instants.

\paragraph{\bf Stopping and recommendation rules at the server:} We use a version of {\em Chernoff's stopping rule} at the server, as outlined below. Let
\begin{equation}
Z(t) \coloneqq \inf_{v' \in {\rm Alt}(\hat{v}(t))}\! \sum_{m=1}^M \, \sum_{i\in S_m} N_{i,m}(t)\,  \frac{(\mu_{i,m}^\prime - \hat{\mu}_{i,m}(t))^2}{2},
\end{equation}
where $\hat{v}(t)$ is the empirical problem instance at time $t$, defined by the empirical means $\{\hat{\mu}_{i,m}(t): i \in S_m, \, m\in [M]\}$ received from the clients, and $v^\prime$ is defined by the means $\{\mu_{i,m}^\prime: i \in S_m, \, m\in [M]\}$. Then, the (random) stopping time of the algorithm is defined as
\begin{equation}
 \tau_\delta(\Pi_{{\rm Het\mhyphen TS}}) = \min\{t \in \{b_r\}_{  r\in \mathbb{N} }: Z(t) \!>\! \beta(t, \delta), t\ge K\},
 \label{eq:chernoff-stopping-time}
\end{equation}
where $\beta(t, \delta)= K' \log(t^2+t) + f^{-1}(\delta)$, with $K'=\sum_{m=1}^M \lvert S_m \rvert$ and $f:(0,+\infty) \rightarrow (0,1)$ defined as
\begin{equation}
\label{eq:deff}
   f(x) \coloneqq \sum_{i=1}^{K'} \frac{x^{i-1}e^{-x}}{(i-1)!}, \quad x \in (0, +\infty).
\end{equation}
Our algorithm outputs
$
\hat{a}_{\delta,m} =  \mathop{\arg\max}_{i \in S_m}  \hat{\mu}_i (\tau_\delta)$
as the best arm of client $m\in [M]$,
where $\hat{\mu}_i (\tau_\delta)=  \frac{1}{M_i} \sum_{m=1}^M \mathbf{1}_{\{i \in S_m\}}\  \hat{\mu}_{i,m}(\tau_\delta)$.

\begin{remark}
Notice that $\Pi_{{\rm Het\mhyphen TS}}$ only stops at the communication time instants $\{b_r\}_{r\in\mathbb{N}}$. This is evident from \eqref{eq:chernoff-stopping-time}.
\end{remark}
\begin{remark}
Our choice of $f$ in \eqref{eq:deff} ensures that the map $x \mapsto f(x)$ is strictly monotone and continuous, and therefore admits an inverse, say $f^{-1}(\cdot)$. Furthermore, an important property of $f^{-1}(\cdot)$ that results from the careful construction of $f$ as in \eqref{eq:deff} is that $\lim_{\delta \to 0} \frac{\log(1/\delta)}{f^{-1}(\delta)}=1$. See Lemma~\ref{lemma:f_delta} in the Appendix for the proof.
\end{remark}

The pseudo-code of $\textsc{Het}\mhyphen \textsc{TS}(\lambda)$ is presented in Algorithm~\ref{alg:client} (for each client $m \in [M]$) and Algorithm~\ref{alg:server} (for the server).

\begin{algorithm}[t]
\caption{$\textsc{Het}\mhyphen \textsc{TS}(\lambda)$: At client $m\in[M]$}
\begin{algorithmic}[1]
\REQUIRE ~~\\
$\delta\in (0,1)$: confidence level.\\
$\lambda>0$: communication frequency parameter.\\
$\{b_r=\lceil (1+\lambda)^r \rceil: r\in \mathbb{N}\}$: communication instants. \\
 $S_m$: arms subset of the client.\\
 
\ENSURE $\hat{a}_{\delta,m}$: the best arm in $S_m$. 
 
 \STATE Initialise $G \leftarrow \mathbf{1}^K$
 \FOR{ $t\in \{1,2,\ldots\} $}
 \STATE Compute  $\{\hat{\omega}_{i,m}(t): i \in S_m\}$ via \eqref{eq:sampling-rule-at-client-m}.
 \IF{$\min_{i\in S_m} N_{i,m}(t-1) < \sqrt{(t-1) / \lvert S_m\rvert}$} 
        \STATE Pull arm $A_m(t) \in \argmin_{i\in S_m}N_{i,m}(t-1)$; resolve ties uniformly.
 \ELSE
    \STATE Pull arm $A_m(t) \in \argmin_{i\in S_m}N_{i,m}(t-1)- t\,  \hat{\omega}_{i,m}(t)$; resolve ties uniformly.
 \ENDIF
 \STATE Update the empirical means $\{\hat{\mu}_{i,m}(t): i\in S_m\}$.
\IF{$t \in \{b_r: r \in \mathbb{N} \}$} 
 \STATE Send the empirical means $\{\hat{\mu}_{i,m}(t): i\in S_m\}$ from client $m$ to the server. 
 \STATE $G$ $\leftarrow$ latest global vector received from the server.
 \ENDIF
  \IF{Server signals to stop further arm pulls}
        \STATE Receive best arm $\hat{a}_{\delta,m}$ from the server.
        \STATE Break.
  \ENDIF
\ENDFOR

\RETURN Best arm $\hat{a}_{\delta, m}$.
\end{algorithmic}
\label{alg:client}
\end{algorithm}

\begin{algorithm}[!ht]
\caption{$\textsc{Het}\mhyphen \textsc{TS}$: At central server}
\begin{algorithmic}[1]

\REQUIRE ~~\\
$\delta\in (0,1)$: confidence level.\\
$\lambda>0$: communication frequency parameter.\\
$\{b_r=\lceil (1+\lambda)^r \rceil: r\in \mathbb{N}\}$: communication instants. \\
$S_1,\ldots, S_M$: sets of clients' accessible arms.

 \STATE Initialize $G \leftarrow \mathbf{1}^K$
 \FOR{ $t \in \{b_r: r \in \mathbb{N}\}$}
 \FOR{$m\in [M]$}
 \STATE Receive $\{\hat{\mu}_{i,m}(t): i \in S_m\}$ from client~$m$.
 \ENDFOR
 
 \IF {$t\ge K$ and $Z(t) > \beta(t,\delta)$}
 \FOR{$m\in [M]$} 
    \STATE Compute the empirical best arm $\hat{a}_{\delta,m}$.
    \STATE Send $\hat{a}_{\delta,m}$ and signal of stop to client~$m$.
 \ENDFOR
\STATE Break.
 \ENDIF
 \STATE Compute the global vector $G$ via the empirical means $\{\hat{\mu}_{i,m}(t): i\in S_m,\, m\in [M]\}$.
 \STATE Broadcast vector $G$ to all the clients.
 \ENDFOR
 
\end{algorithmic}
\label{alg:server}
\end{algorithm}
\section{Results on the Performance of  \texorpdfstring{$\textsc{Het}\mhyphen \textsc{TS}(\lambda)$}{Het-TS(lambda)}}
In this section, we state the results on the performance of $\textsc{Het}\mhyphen \textsc{TS}(\lambda)$ which we denote alternatively by $\Pi_{\textsc{Het}\mhyphen \textsc{TS}}$ (the input parameter $\lambda$ being implicit). The first result below asserts that $\Pi_{\textsc{Het}\mhyphen \textsc{TS}}$ is $\delta$-PAC for any $\delta\in (0,1)$.
\begin{theorem}
\label{theorem:deltapac}
$\textsc{Het}\mhyphen \textsc{TS}(\lambda)$ is $\delta$-PAC for each $\delta \in (0,1)$.
\end{theorem}
The next result provides an asymptotic upper bound on the expected stopping time of $\textsc{Het}\mhyphen \textsc{TS}(\lambda)$ (or $\Pi_{\textsc{Het}\mhyphen \textsc{TS}}$).

\begin{theorem}
\label{Theorem:mainresult}
Fix $\lambda>0$, and let $b_r = \lceil (1+\lambda)^r \rceil$, $r\in \mathbb{N}$. Given any $v\in \mathcal{P}$ and $\delta \in (0,1)$, $\tau_\delta(\Pi_{\textsc{Het}\mhyphen \textsc{TS}})$ satisfies
\begin{equation}
\mathbb{P}_v^{\Pi_{\textsc{Het}\mhyphen\textsc{TS}}} \left(\limsup_{\delta \rightarrow 0} \frac{\tau_{\delta}(\Pi_{\textsc{Het}\mhyphen\textsc{TS}})}{\log\left(\frac{1}{\delta}\right)} \le 2\, (1+\lambda)\, c^*(v) \right) = 1.
\label{eq:almost-surely-finite-stopping-time}
\end{equation}
Furthermore, $\mathbb{E}_v^{\Pi_{\textsc{Het}\mhyphen\textsc{TS}}}[\tau_\delta(\Pi_{\textsc{Het}\mhyphen\textsc{TS}})]$ satisfies
\begin{equation}
\limsup_{\delta \rightarrow 0} \frac{\mathbb{E}_v^{\Pi_{\textsc{Het}\mhyphen\textsc{TS}}}[\tau_\delta(\Pi_{ \textsc{Het}\mhyphen\textsc{TS}  })]}{\log\left(\frac{1}{\delta}\right)} \le 2\, (1+\lambda)\, c^*(v).
\label{eq:upper-bound-expected-stopping-time}
\end{equation}
Thus, in the limit as $\delta \to 0$, $\Pi_{\textsc{Het}\mhyphen\textsc{TS}}$ is asymptotically almost-optimal up to the constant~$\alpha=2\,(1+\lambda)$.
\end{theorem}

Theorem \ref{Theorem:mainresult} lucidly demonstrates the trade-off between the frequency of communication, which is parameterized by $\lambda$, and the expected stopping time. Because $\textsc{Het}\mhyphen\textsc{TS}(\lambda)$ communicates at time instances $b_r = \lceil (1+\lambda)^r \rceil$, as $\lambda$ increases, communication occurs with lesser frequency. This, however, leads to an increase in the multiplicative gap to asymptotic optimality, $2\, (1+\lambda)$. The  factor $1+\lambda$   arises due to the necessity of communicating at time instances whose ratios $\frac{b_{r+1}}{b_r}$ are bounded; see   Lemma~\ref{lemma:upper_bound_on_comm_rouds}. The other factor $2$ (in $2\, (1+\lambda)$) arises from approximating $g_v(\omega)$ by $\widetilde{g}_v(\omega) $ in Lemma~\ref{lemma:relaxed-g}. This factor is required to ensure that the optimal solution to $\widetilde{c}(v)$ and the arm selection probabilities at each time instant can be evaluated  in a tractable fashion.

\begin{corollary} \label{cor:rounds}
Fix $\lambda>0$, and let $b_r = \lceil (1+\lambda)^r \rceil$, $r\in \mathbb{N}$. Given any $v\in \mathcal{P}$ and $\delta \in (0,1)$, $\mathfrak{r}_\delta(\Pi_{ \textsc{Het}\mhyphen\textsc{TS} })$ satisfies
\begin{equation}
\mathbb{P}_v^{\Pi_{ \textsc{Het}\mhyphen\textsc{TS} }}\left(\limsup_{\delta \rightarrow 0} \frac{\mathfrak{r}_\delta(\Pi_{\textsc{Het}\mhyphen\textsc{TS}})}{\log_{1+\lambda} \left(\log \left(\frac{1}{\delta}\right) c^*(v) \right)} \le 1 \right) = 1.
\label{eq:almost-surely-finite-stopping-comm-round}
\end{equation}
Furthermore, $\mathbb{E}_v^{\Pi_{\textsc{Het}\mhyphen\textsc{TS}  }}[\mathfrak{r}_\delta(\Pi_{\textsc{Het}\mhyphen\textsc{TS} })]$ satisfies
\begin{equation}
\limsup_{\delta \rightarrow 0}\  \frac{\mathbb{E}_v^{\Pi_{\textsc{Het}\mhyphen\textsc{TS}}}[\mathfrak{r}_\delta(\Pi_{ \textsc{Het}\mhyphen\textsc{TS}})]}{\log_{1+\lambda} \left(\log\left(\frac{1}{\delta} \right)c^*(v)\right)} \le 1.
\label{eq:upper-bound-expected-stopping-comm-round}
\end{equation}
\end{corollary}

In contrast to Theorem \ref{Theorem:mainresult}, Corollary \ref{cor:rounds} is a statement concerning  the  number of  communication {\em rounds}. It says that the expectation of this quantity scales as $O(\log\log({1}/{\delta}))$. This is perhaps unsurprising given that ${\textsc{Het}\mhyphen\textsc{TS}}(\lambda)$ communicates at time instants $b_r=\lceil(1+\lambda)^r\rceil$, $r\in\mathbb{N}$.  
\section{Solving the Optimal Allocation}
\label{sec:optimal-allocation}

Recall from Section \ref{sec:simplified} that given any problem instance $v\in \mathcal{P}$, the optimal solution to $\max_{\omega \in \Gamma}\ \widetilde{g}_v(\omega)$ may be characterised by a $K$-dimensional global vector $G(v)$ (see Corollary \ref{cor:exact-form-of-optimal-solution} for more details). In this section, we provide the details on how to efficiently compute the global vector $G(v)$ corresponding to any problem instance $v \in \mathcal{P}$. 

\noindent
Consider the {\em relation} $R \coloneqq \{(i_1,i_2):\ \exists\, m\in [M], i_1,i_2 \in S_m \} $ on the arms. Let $R_{\mathrm{e}}$ be the equivalence relation generated by $R$, i.e., the smallest equivalence relation containing~$R$. Clearly, the above equivalence relation $R_{\mathrm{e}}$ partitions $[K]$ into equivalence classes. Let $Q_1, \ldots, Q_L$ be the equivalence classes.
For any $j\in[L]$, let $\widetilde{g}_{v}^{(j)}(\omega) \coloneqq \min_{i\in Q_j} \frac{\Delta^2_{i}(v)}{\frac{1}{M^2_{i}}\sum_{m=1}^M \mathbf{1}_{\{i \in S_m\}} \frac{1}{\omega_{i,m}} }$.
We define $\widetilde{g}^{(j)}_v(\omega) = 0$ if there exists $m\in[M]$ and $i \in S_m \cap Q_j$ such that $\omega_{i,m}=0$. 
In Eqn.~\eqref{eq:exist_solution} in Appendix \ref{appndx:proof-of-theorem-upper-bound}, we argue that the following optimisation problems  admit a common solution:
\begin{align}
    \max_{\omega \in \Gamma} \widetilde{g}_v(\omega),\quad \Big\lbrace\max_{\omega \in \Gamma} \widetilde{g}^{(l)}_v(\omega),\quad  l=1, \ldots, L\Big\rbrace.
    \label{eq:optimisation-problem-multiple}
\end{align}

\begin{definition}[pseudo-balanced condition]
\label{def:pseudo-balanced-condition}
Fix $v\in \mathcal{P}$. An $\omega \in \Gamma$ satisfies {\em pseudo-balanced condition} if  $\frac{\Delta^2_{i_1}(v)}{\frac{1}{M^2_{i_1}}\sum_{m=1}^M  \frac{\mathbf{1}_{\{i_1 \in S_m\}}}{\omega_{i_1,m}} } = \frac{\Delta^2_{i_2}(v)}{\frac{1}{M^2_{i_2}}\sum_{m=1}^M  \frac{\mathbf{1}_{\{i_2 \in S_m\}}}{\omega_{i_2,m}} }$ for all  $ j \in [L]$ and $i_1,i_2 \in Q_{j}$.
\end{definition}
The next result states that the common solution to \eqref{eq:optimisation-problem-multiple}  satisfies  {\em balanced condition} 
and {\em pseudo-balanced condition}.
\begin{theorem}
\label{theorem:unique-allocation}
 For any $v\in \mathcal{P}$, the common solution to the optimization problems in \eqref{eq:optimisation-problem-multiple} is unique and satisfies balanced condition (Def. \ref{def:balanced-condition}) and pseudo-balanced condition.
\end{theorem}
Let $\widetilde{w}(v)$ be the unique common solution to \eqref{eq:optimisation-problem-multiple} corresponding to the instance $v$. Let $G(v)$ be the unique global vector characterising $\widetilde{w}(v)$ (see Corollary \ref{cor:exact-form-of-optimal-solution}) with $G(v)>0$ and
\begin{equation}
\label{eq:unique-GV}
\lVert G^{(j)}(v) \rVert_2 =1  \quad \forall\, j\in[L],
\end{equation}
where $G^{(j)}(v) \in \mathbb{R}^{\lvert Q_j \rvert}$ denotes the sub-vector of $G(v)$ formed from the rows corresponding to the arms $i \in Q_j$.
Let $H(v) \in \mathbb{R}^{K \times K}$ be the matrix defined by
\begin{equation}
    H(v)_{i_1,i_2} := \frac{1}{\Delta^2_{i_1}(v) M^2_{i_1}}\sum_{m=1}^M \mathbf{1}_{\{i_1,i_2\in S_m\}}, \  i_1,i_2 \in [K].
    \label{eq:H(v)}
\end{equation}
For $j\in [L]$, let $H^{(j)}(v) \in \mathbb{R}^{\lvert Q_j \rvert \times \lvert Q_j \rvert}$ be the sub-matrix of $H(v)$ formed from the rows and columns corresponding to the arms in $Q_j$. It is easy to verify that $\big( H^{(j)}(v) \big)^\top H^{(j)}(v) = H^{(j)}(v)\big( H^{(j)}(v) \big)^\top$. That is, $H^{(j)}(v)$ is a {\em normal matrix} \cite[Chapter 2, Section 2.5]{horn2012matrix} and therefore has $\lvert Q_j \rvert$ linearly independent eigenvectors. In Appendix \ref{sec:proofs-of-optimal-allocation}, we show that $G^{(j)}(v)$ is an eigenvector of the matrix $H^{(j)}(v)$ and that the eigenspace of $H^{(j)}(v)$ is one-dimensional. Building on these results,
the main result of this section, a recipe for computing the global vector corresponding to an instance, is given below.
\begin{proposition}
\label{proposition:compute-global-vector}
Fix $j\in [L]$ and a problem instance $v \in \mathcal{P}$. Among any   set of $\lvert Q_j \rvert$ linearly independent eigenvectors of $H^{(j)}(v)$, there exists only one vector $\mathbf{u}$ whose elements are all negative ($\mathbf{u}<\mathbf{0}$) or all positive ($\mathbf{u} > \mathbf{0}$). Furthermore, 
\begin{equation}
 G^{(j)}(v)=
\begin{cases}
      -\frac{\mathbf{u}}{\lVert \mathbf{u} \lVert_2},\ &\text{if } \mathbf{u}<\mathbf{0} , \\
      \frac{\mathbf{u}}{\lVert \mathbf{u} \lVert_2},\ &\text{if }\mathbf{u}>\mathbf{0}.
\end{cases}
\label{eq:all-positive-or-all-negative-entries}
\end{equation}
\end{proposition}
Proposition~\ref{proposition:compute-global-vector}  provides an efficient recipe to compute the global vector $G(v)$: it is the unique eigenvector of $H^{(j)}(v)$
with either all-positive or all-negative entries and normalised to have unit norm. We use this recipe in our implementation of $\textsc{Het}\mhyphen\textsc{TS}(\lambda)$ on synthetic and real-world datasets (e.g., the MovieLens dataset \cite{cantador2011second}).  

\section{Experimental Results} \label{sec:expts}

In this section, we corroborate our theoretical results by implementing $\textsc{Het}\mhyphen \textsc{TS}(\lambda)$ and performing a variety of experiments on a synthetic dataset and the MovieLens dataset. 

\subsection{Synthetic Dataset}

 In our synthetic dataset, the instance we used contains $M=5$ clients and $K=5$ arms. The expected mean reward $\mu_{i,m}$ is chosen uniformly at random from $[7-i, 7-i+1]$. As a consequence, $\mu_{i}$ is also uniformly random in $[7-i, 7-i+1]$.
\begin{figure}[!ht]
    \centering
\includegraphics[width = .75\columnwidth]{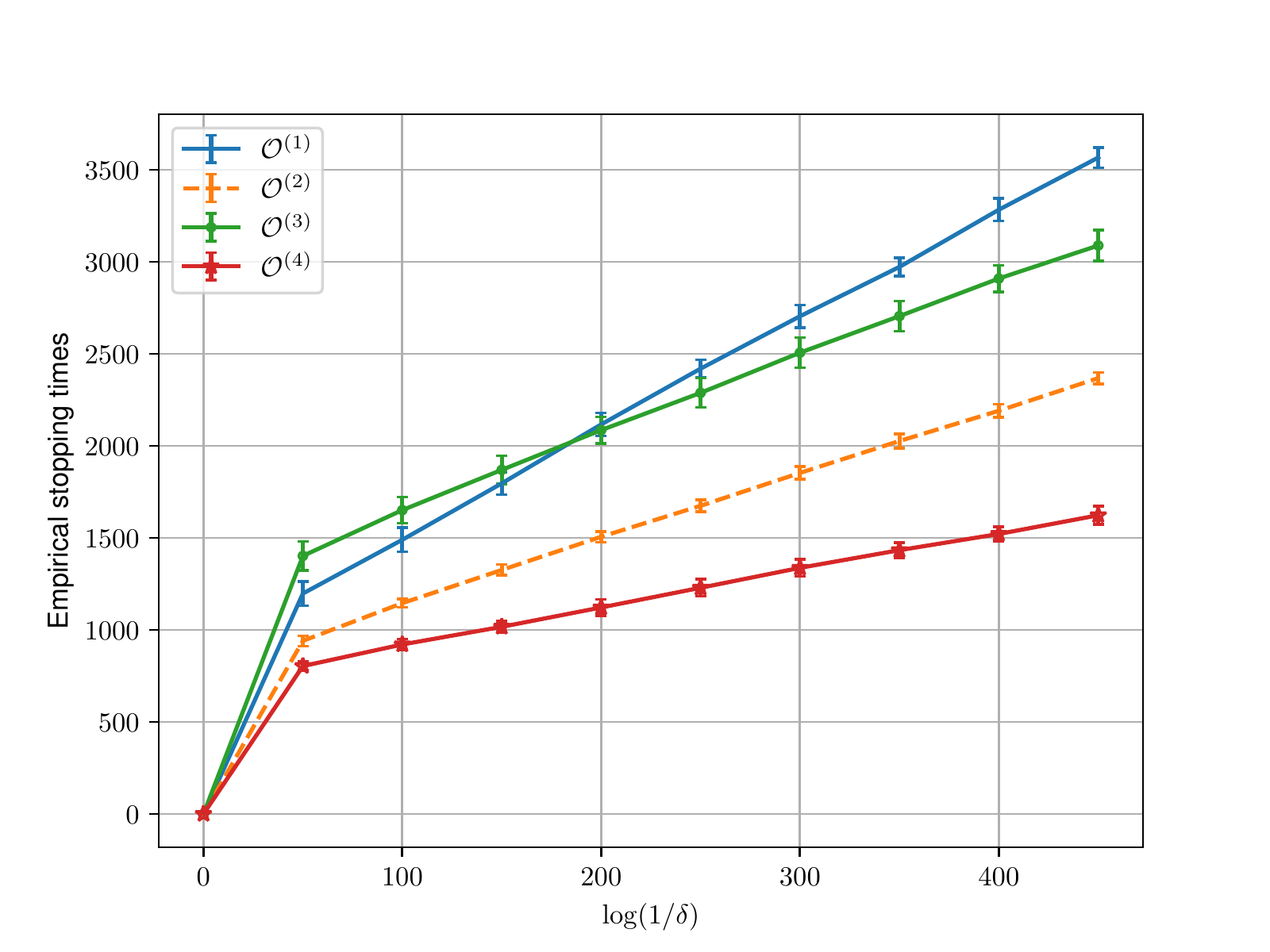}
    \caption{Expected stopping times for various overlap patterns as described in~\eqref{eqn:overlap}.}
    \label{fig:overlaps}
\end{figure}
\begin{figure}[!ht]
    \centering
\includegraphics[width = .75\columnwidth]{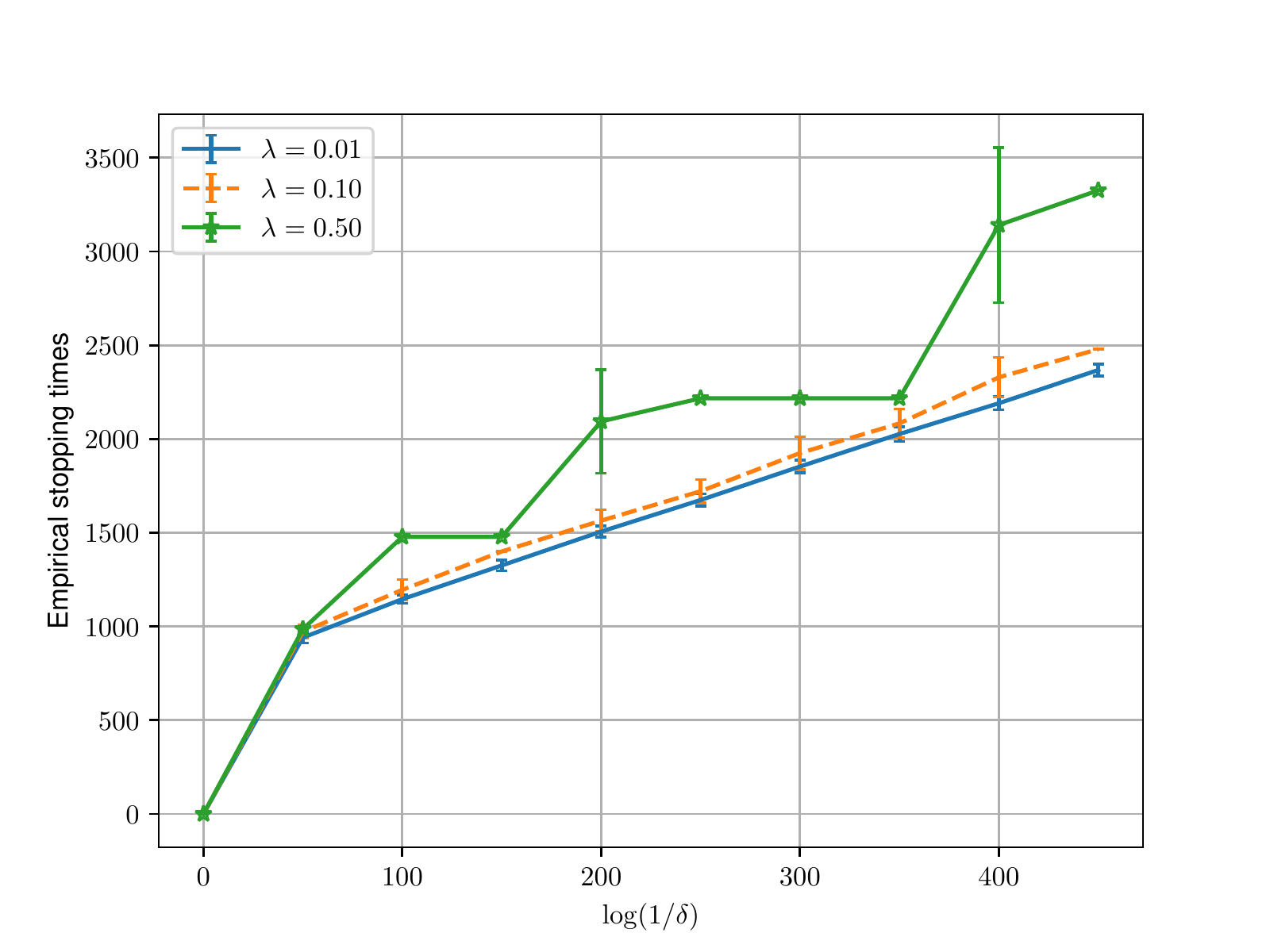}
    \caption{Expected stopping times for various $\lambda$'s and for overlap pattern $\mathcal{O}^{(2)}$.}
    \label{fig:lambdas}
\end{figure}
To empirically evaluate the effect of various sets $\{S_m\}_{m\in [M]}$ on the expected stopping time, we consider different {\em overlap patterns} (or {\em multisets}) of the form $\mathcal{O}^{(p)} = \{ S_{1}^{(p)} ,S_{2}^{(p)} ,\ldots, S_{5}^{(p)}  \}$, where 
\begin{align}
    \mathcal{O}^{(1)}  &= \big\{ \{1,2\}, \{2,3\}, \{3,4\},\{4,5\}, \{5,1\} \big\}, \nonumber \\
    \mathcal{O}^{(2)}  &= \big\{ \{1,2,3\}, \{2,3,4\}, \{3,4,5\} ,\{4, 5,1\} , \{5,1,2\} \big\}\nonumber \\
    \mathcal{O}^{(3)}  &= \big\{ \{1,2,3,4\}, \{2,3,4,5\},  \{3,4,5,1\},  \{4,5,1,2\}, \{5,1,2,3\} \big\},\quad\mbox{and} \nonumber\\
    \mathcal{O}^{(4)}  &= \big\{ \{1,2,3,4,5\}, \{1,2,3,4,5\},  \{1,2,3,4,5\}, \{1,2,3,4,5\}, \{1,2,3,4,5\} \big\} . \label{eqn:overlap}
\end{align}
Thus, the larger the index  of the overlap  pattern $p$, the larger the  overlap among the sets $\{S_m^{(p)}\}_{m\in [M]}$, and therefore the larger the number of clients that have access to a fixed arm $i \in [K]$. The mean values $\{\mu_{i,m}\}_{i\in[K],m\in[M]}$, together with an overlap pattern $\mathcal{O}^{(p)}$, uniquely defines a problem instance  $v=(\{\mu_{i,1}\}_{i\in S_1^{(p)}}, \{\mu_{i,2}\}_{i\in S_2^{(p)}},\dots, \{\mu_{i,M}\}_{i\in S_M^{(p)}})$.

\subsubsection{Effect of Amount of Overlap}
The empirical expected stopping times of $\textsc{Het}\mhyphen\textsc{TS}(\lambda)$ for $\lambda=0.01$  are displayed in Fig.~\ref{fig:overlaps}. It can be seen that as $\delta$ decreases, the empirical stopping time increases, as expected. More interestingly, note that for a fixed $\delta$, the stopping time is {\em not monotone} in the amount of overlap. This   is due to two  factors that work in opposite directions as one increases the  amount of overlap of $S_m$'s among various clients. On the one hand, each client has access to more arms, yielding more information about the bandit instance for the client. On the other hand, with more arms, the set of arms that can   potentially be the best arm for that particular client also increases. 
This observation is interesting and, at first glance, counter-intuitive.

\begin{figure}[!ht]
    \centering
\includegraphics[width =.75\columnwidth]{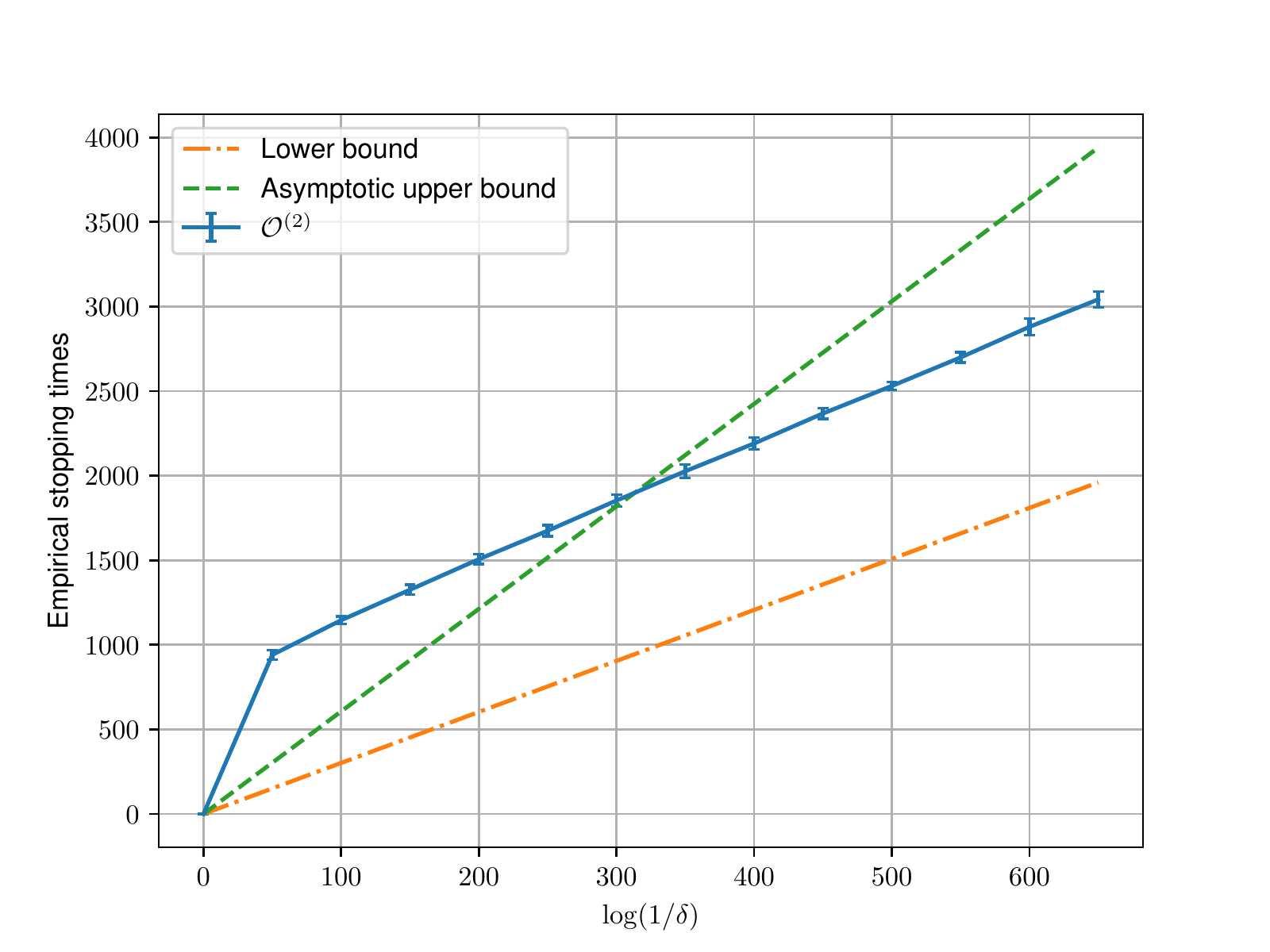}
    \caption{Comparisons of the upper and lower bounds.}
    \label{fig:compare}
\end{figure}

\subsubsection{Effect of Communication Frequency}
Recall that $\textsc{Het}\mhyphen\textsc{TS}(\lambda)$ communicates and stops at those time instants $t$ of the form $b_r=\lceil (1+ \lambda)^r \rceil$ for $r\in\mathbb{N}$. As $\lambda$ increases, the communication frequency decreases. In other words, $\textsc{Het}\mhyphen\textsc{TS}(\lambda)$ is is communicating at {\em sparser} time instants.  Thus as $\lambda$ grows, we should  expect that the stopping times  increase commensurately as the server receives less data per unit time. This is reflected in Fig.~\ref{fig:lambdas} where we use the instance with $\mathcal{O}^{(2)}$. 

We note another interesting phenomenon, most evident from the curve indicated by $\lambda = 0.5$. The growth pattern of the empirical stopping time has a piecewise linear shape. This is because  $\textsc{Het}\mhyphen\textsc{TS}(\lambda)$ does not stop at any arbitrary integer time; it only does so at the times that correspond to {\em communication rounds} $b_r =  \lceil (1+ \lambda)^r \rceil$ for $r\in\mathbb{N}$.  Hence, for $\delta$ and $\delta'$ sufficiently close, the empirical stopping times will be {\em exactly the same} with high probability.  This explains the piecewise linear stopping pattern as $\log(1/\delta)$ grows. 

\subsubsection{Comparison to Theoretical Bounds}

In the final experiment for synthetic data, we set $\lambda=0.01$ and overlap pattern $\mathcal{O}^{(2)}$ as our instance $v$. In Fig.~\ref{fig:compare}, we compare the empirical stopping time  to the lower bound in Proposition \ref{prop:lower-bound} and the upper bound in Theorem~\ref{Theorem:mainresult}. Recall that the asymptotic ratio of the expected stopping time ${\mathbb{E}_v^{\Pi_{\textsc{Het}\mhyphen\textsc{TS}}}[\tau_\delta(\Pi_{ \textsc{Het}\mhyphen\textsc{TS}  })]}$ to $\log(1/\delta)$ is $c^*(v)$ and $2(1+\lambda)c^*(v)$ in the lower and upper bounds respectively.   We observe that as $\delta$ becomes sufficiently small,  the slope of the empirical curve lies between the upper and lower bounds, as expected. 

Furthermore, we see that  ${\mathbb{E}_v^{\Pi_{\textsc{Het}\mhyphen\textsc{TS}}}[\tau_\delta(\Pi_{ \textsc{Het}\mhyphen\textsc{TS}  })]}/{\log\left(1/{\delta}\right)}$ is  close to the lower bound, which strongly suggests our learned allocation $\widetilde{\omega}(\hat{v}(t))$ is very close to optimal allocation $\arg\max_{\omega \in \Gamma} g_{\hat{v}(t)}(\omega)$. We observe from Fig.~\ref{fig:compare} that the empirical performance or, more precisely, the slope of the expected stopping time as a function of $\log(1/\delta)$ is close to $(1+\lambda)\, c^*(v)$. This suggests that the factor $1+\lambda$ (in $2\, (1+\lambda)$)  in Theorem~\ref{Theorem:mainresult} is unavoidable if we communicate at time instances that grow as $\Theta( (1+\lambda)^r)$. The presence of the  factor $2$ (in $2\, (1+\lambda)$) is  to enable the optimal allocation $\hat{\omega}_{i,m}(t)$ to be solved in a tractable fashion. For more details concerning this point, see the discussion following Theorem~\ref{Theorem:mainresult}.
\subsection{MovieLens Dataset}
In the MovieLens dataset \cite{cantador2011second}, there are about 2.2 million rating samples and 10,197 movies. Following the experimental settings in \cite{reddy2022almost}, we view each {\em country} and {\em genre} as a client and an arm, respectively. Besides, we normalize the rating score in the range of 0 to 100. We note that in the raw dataset that there are very few or even no samples for some combinations of country and genre. Thus, in our experiment we discard any country and genre  pair with fewer than ten samples. As a result, we end up with 10,044 movies and $M=48$ clients across $K=19$ arms. It is natural that different clients  have different arm sets in the  dataset; this dovetails neatly with our problem setting in which $S_m$'s need not be the same as one another and they need not be the full set $[K]$.

\noindent
As in \cite{yang2022optimal}, we compare our algorithm to a baseline method which we call {\sc Uniform}, having the same stopping rule as $\textsc{Het}\mhyphen\textsc{TS}(\lambda)$, but using a uniform sampling rule at each client (i.e., each client samples an arm uniformly at random at each time instant). We note that {\sc Uniform} is $\delta$-PAC for all $\delta \in(0,1)$.  Our numerical results, which are obtained by averaging over four independent experiments and by setting $\lambda=0.01$, are presented in  Table~\ref{tab:movielens}. We observed from our experiments that the statistical variations of the results are minimal (and virtually non-existent) as the algorithm necessarily stops at one of the time instants of the form $b_r = \lceil(1+\lambda)^r\rceil$ for $r\in\mathbb{N}$. Hence, ``error bars'' are not indicated. From Table~\ref{tab:movielens},  we observe that the ratio of empirical stopping time between {\sc Uniform} and $\textsc{Het}\mhyphen\textsc{TS}(\lambda)$ is {\em approximately eight},  showing that the sampling rule of  $\textsc{Het}\mhyphen\textsc{TS}(\lambda)$ is highly effective in rapidly identifying the best arms in this real-world dataset.
\begin{table}[!t]
\centering
\begin{tabular}{|l|c|c|c|c|c|c|c|}
\hline
$\log(1/\delta)$ & 10 & 50  & 100    & 200 & 500 & 1000    \\ \hline
$\textsc{Het}\mhyphen\textsc{TS} (\lambda)$, $\lambda=0.01$      &  {\bf 32,473} & {\bf 32,798} & {\bf 33,457 } & {\bf 34,129} & {\bf 35,870} & {\bf 38,458}  \\
\hline
{\sc Uniform}       & 252,184 & 254,706 & 259,826&  265,048 & 278,568 & 292,778 \\ 
\hline
\end{tabular}
\vspace{0.5\baselineskip}
\caption{A comparison of the empirical stopping times of $\textsc{Het}\mhyphen\textsc{TS}(\lambda)$ and {\sc Uniform} for $\lambda=0.01$.}
\label{tab:movielens}
\end{table}

\section{Concluding Remarks and Future Work}\label{app:conclusion}
We studied best arm identification in a federated multi-armed bandit with heterogeneous clients in which each client can access a {\em subset} of the arms; this was 
 mainly motivated from the unavailability of authorised vaccines in certain countries. We showed, among other results, that any {\em almost-optimal} algorithm must necessarily communicate such that the ratio of consecutive time instants is bounded, and that an algorithm may communicate at most exponentially sparsely while being almost-optimal. 
We proposed a track-and-stop-based algorithm that communicates exponentially sparsely and is almost-optimal up to an identifiable multiplicative constant in the regime of vanishing error probabilities. Future work includes carefully examining the effects of heterogeneity, possible corruptions, and the quantisation of various messages on the uplinks and downlinks. {\color{black}Additionally, an outstanding question concerns the  derivation of a lower bound on the number of  communication rounds as a function of $\alpha$ (the multiplicative gap to the lower bound per Definition~\ref{defn:almost-optimal-algo}) rather than $\eta$ (which parameterizes the frequency of communication); see Remark~\ref{rmk:alpha}.}

\bibliographystyle{IEEEtran}
\bibliography{references}

\appendix




\section{Proof of Lemma \ref{lemma:relaxed-g}}
\label{appndx:proof-of-lemma-relaxed-g}
\begin{proof}
Fix $v=\{\mu_{i,m}: i\in S_m, \, m\in [M]\}\in \mathcal{P}$ and $\omega\in \Gamma$. Let 
\begin{equation}
\mathcal{C}(v) \coloneqq \bigcup_{m\in [M]} \Big\{(a^*_m(v),i): i \in S_m \setminus \{a^*_m(v)\} \Big\}.
\label{eq:C(v)}
\end{equation} 
First, we note that by definition, $g_v(\omega)=0$ if $\omega_{i,m}=0$ for some $m\in[K]$ and $i\in S_m$. Therefore, it suffices to consider the case when $\omega_{i,m}>0$ for all $m\in[K]$ and $i\in S_m$. In what follows, we abbreviate $\mu_{i,m}(v')$ and  $\mu_{i}(v')$ to $\mu'_{i,m}$  and  $\mu'_{i}$ respectively.
We have
\begin{align}
     g_v(\omega) 
    & = \inf_{v'\in {\rm Alt}(v)}\  \sum_{m=1}^{M}\  \sum_{i \in S_m}  \omega_{i,m}\frac{(\mu_{i,m}-\mu'_{i,m})^2}{2} \nonumber\\
                &=  \min_{(i_1,i_2)\in \mathcal{C}(v)}\  \inf_{v':\mu'_{i_1}<\mu'_{i_2}}\  \sum_{m=1}^{M}\  \sum_{i \in S_m}  \omega_{i,m}\frac{(\mu_{i,m}-\mu'_{i,m})^2}{2} \nonumber\\
                &=  \min_{(i_1,i_2)\in \mathcal{C}(v)}\  \inf_{v':\mu'_{i_1} \leq \mu'_{i_2}} \ \sum_{m=1}^{M} \mathbf{1}_{\{i_1\in S_m\}}  \omega_{i_1,m}\frac{(\mu_{i_1,m}-\mu'_{i_1,m})^2}{2}+ 
                \mathbf{1}_{\{i_2\in S_m\}}  \omega_{i_2,m}\frac{(\mu_{i_2,m}-\mu'_{i_2,m})^2}{2} \nonumber\\
                &= \min_{(i_1,i_2)\in \mathcal{C}(v)} \ \frac{(\mu_{i_1}-\mu_{i_2})^2/2}{\frac{1}{M^2_{i_1}}\sum_{m=1}^M \mathbf{1}_{\{i_1 \in S_m\}} \frac{1}{\omega_{i_1,m}} + \frac{1}{M^2_{i_2}}\sum_{m=1}^M \mathbf{1}_{\{i_2 \in S_m\}} \frac{1}{\omega_{i_2,m}}},
                \label{eq:g_v(omega)-simplification}
\end{align}
where \eqref{eq:g_v(omega)-simplification} follows from the penultimate line by using the method of Lagrange multipliers and noting that the inner infimum in the penultimate line is attained at 
\begin{align}
    \mu'_{i_1,m} &= \mu_{i_1,m} - \frac{\mu_{i_1}-\mu_{i_2}}{M_{i_1}\omega_{i_1,m}(\sum_{i \in \{i_1,i_2\}}\sum_{m':i \in S_{m'}} \frac{1}{ \omega_{i,m'}  M^2_{i}})} \quad \forall\,  m : i_1 \in S_{m},\nonumber\\
    \mu'_{i_2,m} &= \mu_{i_2,m} + \frac{\mu_{i_1}-\mu_{i_2}}{M_{i_2}\omega_{i_2,m}(\sum_{i \in \{i_1,i_2\}}\sum_{m':i \in S_{m'}} \frac{1}{\omega_{i,m'} M^2_{i} })} \quad \forall \, m :i_2 \in S_{m}
    \label{eq:optimisers-of-inner-infimum}
\end{align}
From the definition of $\widetilde{g}_v(\omega)$ in (\ref{eq:simplified-inner-minimum-term}), it is easy to verify that 
\begin{equation}
    \widetilde{g}_v(\omega) = \min_{(i_1,i_2)\in \mathcal{C}(v)} \min \bigg\{ \frac{(\mu_{i_1}-\mu_{i_2})^2/2}{\frac{1}{M^2_{i_1}}\sum_{m=1}^M \mathbf{1}_{\{i_1 \in S_m\}} \frac{1}{\omega_{i_1,m}} }, 
    \frac{(\mu_{i_1}-\mu_{i_2})^2/2}{\frac{1}{M^2_{i_2}}\sum_{m=1}^M \mathbf{1}_{\{i_2 \in S_m\}} \frac{1}{\omega_{i_2,m}}} \bigg\},
\end{equation}
whence it follows that $
\frac{\widetilde{g}_v(\omega)}{2} \le g_v(\omega)  \le  \widetilde{g}_v(\omega)$.
\end{proof}

{\color{black}
\section{Proof Proposition \ref{prop:lower-bound}}
We begin with a useful lemma that is used several times to establish various lower bounds. 
\begin{lemma}
\label{lemma:useful-lemma}
Let $T < \infty$ be any fixed time instant, and let $\mathcal{F}_T = \sigma(\{X_{A_m(t),m}(t),A_m(t): t\in [T], m\in[M]\})$ be the history of all the arm pulls and rewards seen up to time $T$ at all the clients under an algorithm $\Pi$. Let $E$ be any event such that $
\mathbf{1}_{E}$ is $\mathcal{F}_T$-measurable. Then, for any pair of problem instances $v$ and $v'$, 
\begin{align}
		\sum_{t=1}^{T}\  \sum_{m=1}^M\  \sum_{i \in S_m} \mathbb{E}_{v}^\Pi\left[\mathbf{1}_{\{A_m(t)=i\}}\  D_{\mathrm{KL}}(v_{i,m} \| v'_{i,m} )\right] \ge  d_{\mathrm{KL}}\left(\mathbb{P}_{v}^\Pi(E), \mathbb{P}_{{v'}}^{\Pi}(E)\right),
		\label{eq:transportation-lemma}
\end{align}  
where $D_{\mathrm{KL}}(p \| q)$ denotes the Kullback--Leibler (KL) divergence between  distributions $p$ and $q$, and $d_{\mathrm{KL}}(x,y)$ denotes the KL divergence between two Bernoulli distributions with parameters $x$ and $y$. 
\end{lemma}
The proof of Lemma \ref{lemma:useful-lemma} follows along the exact same lines as the proof of~\cite[Lemma 19]{Kaufmann2016}, and is hence omitted.  
\begin{proof}[Proof of Proposition \ref{prop:lower-bound}]
Fix $v \in \mathcal{P}$, and a $\delta$-PAC policy $\Pi$, and $\delta \in (0,1)$. Let $E$ denote the event that the empirical best arm is $a^*(v)$, i.e.,
$$
E = \{ \hat{a}_\delta = a^*(v) \}
$$

By Lemma~\ref{lemma:useful-lemma}, for any $v' \in {\rm Alt}(v)$, we have,
\begin{align}
  \sum_{t=1}^{\infty}\  \sum_{m=1}^M\  \sum_{i \in S_m} \mathbb{E}_{v}^\Pi\left[\mathbf{1}_{\{A_m(t)=i\}}\  D_{\mathrm{KL}}(v_{i,m} \| v'_{i,m} )\right] \ge  d_{\mathrm{KL}}\left(\mathbb{P}_{v}^\Pi(E), \mathbb{P}_{{v'}}^{\Pi}(E)\right).
  \end{align}
By using the fact that the arm reward distributions are Gaussian with unit variance, we have 
\begin{align}
  \sum_{t=1}^{\infty}\  \sum_{m=1}^M\  \sum_{i \in S_m} \mathbb{E}_{v}^\Pi\left[\mathbf{1}_{\{A_m(t)=i\}}\  \frac{(\mu_{i,m}-\mu'_{i,m})^2}{2} \right] \ge  d_{\mathrm{KL}}\left(\mathbb{P}_{v}^\Pi(E), \mathbb{P}_{{v'}}^{\Pi}(E)\right).
  \end{align}
Now, by  observing that  $\mathbb{P}_{v}^\Pi(E) \ge 1-\delta$ and $\mathbb{P}_{v'}^\Pi(E) \le \delta$ since all policies considered are $\delta$-PAC, we obtain
  \begin{align}
  \sum_{t=1}^{\infty}\  \sum_{m=1}^M\  \sum_{i \in S_m} \mathbb{E}_{v}^\Pi\left[\mathbf{1}_{\{A_m(t)=i\}}\  \frac{(\mu_{i,m}-\mu'_{i,m})^2}{2} \right] \ge  \log\Big(\frac{1}{4\delta}\Big) \label{proof_lowerboud:step1},
\end{align}

Then, denoting $\bar{\omega}_{i,m} \coloneqq \frac{1}{\mathbb{E}_{v}^\Pi[ \tau_\delta(\Pi)]} \sum_{t=1}^{+\infty} \mathbb{E}_{v}^\Pi\left[\mathbf{1}_{\{A_m(t)=i\}} \right]$,~\eqref{proof_lowerboud:step1} implies  that
\begin{align}
\mathbb{E}_{v}^\Pi [ \tau_\delta(\Pi) ] \sum_{m=1}^M\  \sum_{i \in S_m} \bar{\omega}_{i,m} \frac{(\mu_{i,m}-\mu'_{i,m})^2}{2}  \ge  \log\Big(\frac{1}{4\delta}\Big)
\end{align}
which leads to 
\begin{align}\mathbb{E}_{v}^\Pi [ \tau_\delta(\Pi) ]   \ge  
 \frac{\log(\frac{1}{4\delta})}{ \sum_{m=1}^M\  \sum_{i \in S_m} \bar{\omega}_{i,m} \frac{(\mu_{i,m}-\mu'_{i,m})^2}{2}}. \label{proof_lowerboud:step2}
\end{align}
By using the fact that~\eqref{proof_lowerboud:step2} holds for any $v' \in {\rm Alt}(v)$, we have
\begin{equation}
\mathbb{E}_{v}^\Pi [\tau_\delta(\Pi)]  \ge  
  \frac{\log(\frac{1}{4\delta})}{\inf_{v' \in {\rm Alt}(v)} \sum_{m=1}^M\  \sum_{i \in S_m} \bar{\omega}_{i,m} \frac{(\mu_{i,m}-\mu'_{i,m})^2}{2}} . \label{proof_lowerboud:step3}
\end{equation}
Next, by using the fact that $\bar{\omega} \in \Gamma$,~\eqref{proof_lowerboud:step3} implies  that
\begin{equation}
\mathbb{E}_{v}^\Pi[  \tau_\delta(\Pi)]   \ge  
  \frac{\log(\frac{1}{4\delta})}{\sup_{\omega \in \Gamma} \inf_{v' \in {\rm Alt}(v)} \sum_{m=1}^M\  \sum_{i \in S_m} \omega_{i,m} \frac{(\mu_{i,m}-\mu'_{i,m})^2}{2}}. \label{proof_lowerboud:step4}
\end{equation} 
By~\eqref{eq:g_v(omega)-simplification}, $g_v(\omega) = \frac{1}{2}\inf_{v'\in {\rm Alt}(v)} \sum_{m=1}^{M} \sum_{i \in S_m}  \omega_{i,m} (\mu_{i,m}(v)-\mu_{i,m}(v'))^2$ is  continuous in $\omega$. Furthermore, by noting that $\Gamma$ is compact, \eqref{proof_lowerboud:step4} implies that 
\begin{equation}
\mathbb{E}_{v}^\Pi [ \tau_\delta(\Pi) ]  \ge  
  \frac{\log(\frac{1}{4\delta})}{\max_{\omega \in \Gamma} \inf_{v' \in {\rm Alt}(v) } \sum_{m=1}^M\  \sum_{i \in S_m} \omega_{i,m} \frac{(\mu_{i,m}-\mu'_{i,m})^2}{2}} \label{proof_lowerboud:step5}
\end{equation}
as desired. This completes the proof.
\end{proof}
}

\section{Proof of Theorem \ref{lemma:upper_bound_on_comm_rouds}}
Define $v_{\dagger}(\rho)$ to be a special problem instance in which the arm means are given by
\begin{equation}
    \mu(v_{\dagger}(\rho))_{i,m} = \frac{i}{\sqrt{\rho}}, \quad m\in [M], \ i \in S_m.
    \label{eq:arm-means}
\end{equation}
Then, it follows that $\mu(v_{\dagger}(\rho))_{i} = \frac{i}{\sqrt{\rho}}$ for all $i\in [K]$. The following result will be used in the proof of Theorem \ref{lemma:upper_bound_on_comm_rouds}.
\begin{lemma}
\label{lemma:construct_instances}
Given any $\rho>0$, the problem instance $v_{\dagger}(\rho)$,  defined in \eqref{eq:arm-means}, satisfies
\begin{equation}
\label{eq:constructing_instance_hardness}
    \frac{4\rho}{MK^2}   \le c^*\big(v_{\dagger}(\rho)\big) \le  4K\rho.
\end{equation}
\end{lemma}
\begin{proof}[Proof of Lemma \ref{lemma:construct_instances}]
 Recall the definition of $\mathcal{C}(\cdot)$ in (\ref{eq:C(v)}). Let
\begin{equation}
    \Delta_{\rm min}(\rho)\coloneqq \min_{(i,j) \in \mathcal{C}(v_{\dagger}(\rho))} \rvert \mu_i(v_{\dagger}(\rho)) - \mu_j(v_{\dagger}(\rho)) \vert,
    \label{eq:Delta-min-rho}
\end{equation}
and let $\omega^{\rm trivial} \in \Gamma$ be defined as
\begin{equation}
    \omega^{\rm trivial}_{i,m} = \frac{1}{\vert S_m \rvert}, \quad m\in [M], \ i \in S_m.
    \label{eq:omega-trivial}
\end{equation}
Notice that
\begin{align}
    c^*(v_{\dagger}(\rho))^{-1} &= \max_{\omega \in \Gamma}\  \inf_{v'\in {\rm Alt}(v)}\  \sum_{m=1}^{M}\  \sum_{i\in S_m}  \omega_{i,m}\frac{\big(\mu_{i,m}(v)-\mu_{i,m}(v')\big)^2}{2} \nonumber\\
    &\leq \inf_{v'\in {\rm Alt}(v)}\  \sum_{m=1}^{M}\  \sum_{i\in S_m}  1\cdot \frac{\big(\mu_{i,m}(v)-\mu_{i,m}(v')\big)^2}{2}\nonumber\\
    &=g_{v_{\dagger}(\rho)}(\mathbf{1}^{K\times M}),
    \label{eq:c-star-upper-bound}
\end{align}
where $\mathbf{1}^{K \times M}$ denotes the all-ones matrix of dimension $K \times M$. Also notice that
\begin{align}
    c^*(v_{\dagger}(\rho))^{-1} &= \max_{\omega \in \Gamma}\  \inf_{v'\in {\rm Alt}(v)}\  \sum_{m=1}^{M}\  \sum_{i\in S_m}  \omega_{i,m}\frac{\big(\mu_{i,m}(v)-\mu_{i,m}(v')\big)^2}{2} \nonumber\\
    & \geq \inf_{v'\in {\rm Alt}(v)}\  \sum_{m=1}^{M}\  \sum_{i\in S_m}  \omega_{i,m}^{\rm trivial}\, \frac{\big(\mu_{i,m}(v)-\mu_{i,m}(v')\big)^2}{2} \nonumber\\
    &= g_{v_{\dagger}(\rho)}(\omega^{\rm trivial}).
    \label{eq:c-star-lower-bound}
\end{align}
From \eqref{eq:c-star-upper-bound} and \eqref{eq:c-star-lower-bound}, we have
\begin{align}
 & g_{v_{\dagger}(\rho)}(\omega^{\rm trivial}) \le c^*\big(v_{\dagger}(\rho)\big)^{-1} \le g_{v_{\dagger}(\rho)}(\mathbf{1}^{K\times M}) \nonumber \\
 & \stackrel{(a)}{\implies}  \min_{(i,j) \in \mathcal{C}(v_{\dagger}(\rho))} \frac{\big(\mu_i(v_{\dagger}(\rho))-\mu_j(v_{\dagger}(\rho))\big)^2/2}{\frac{1}{M^2_i}\sum_{m=1}^M  \mathbf{1}_{\{i \in S_m\}}\frac{1}{\omega^{\rm trivial}_{i,m}} +\frac{1}{M^2_j}\sum_{m=1}^M  \mathbf{1}_{\{j \in S_m\}}\frac{1}{\omega^{\rm trivial}_{j,m}} } \nonumber  \\
 &\hspace{2cm} \le c^*\big(v_{\dagger}(\rho)\big)^{-1}  \le  \min_{(i,j) \in \mathcal{C}(v_{\dagger}(\rho))} \frac{\big(\mu_i(v_{\dagger}(\rho))-\mu_j(v_{\dagger}(\rho))\big)^2/2}{\frac{1}{M^2_i}\sum_{m=1}^M  \mathbf{1}_{\{i \in S_m\}}\frac{1}{1} +\frac{1}{M^2_j}\sum_{m=1}^M  \mathbf{1}_{\{j \in S_m\}}\frac{1}{1} } \nonumber \\
 & \stackrel{(b)}{\implies} \frac{\Delta^2_{\rm min}(\rho)}{4K}   \le c^*\big(v_{\dagger}(\rho)\big)^{-1} \le \frac{M\Delta^2_{\rm min}(\rho)}{4} \nonumber \\
  & \stackrel{(c)}{\implies} \frac{1}{4K\rho}   \le c^*\big(v_{\dagger}(\rho)\big)^{-1} \le \frac{MK^2}{4\rho} \nonumber \\
   & \implies \frac{4\rho}{MK^2}   \le c^*\big(v_{\dagger}(\rho)\big) \le  4K\rho,
   \label{eq:proving-lemma-upper-and-lower-bound-on-cstar-v-cst-rho}
\end{align}
where $(a)$ above follows from \eqref{eq:g_v(omega)-simplification} of Lemma \ref{lemma:relaxed-g}, in writing $(b)$, we make use of the observation that for all $m\in [M]$ and $i\in S_m$,
$$
\frac{1}{\omega^{\rm trivial}_{i,m}} = |S_m| \leq K,  
$$
and $(c)$ makes use of the fact that $\Delta^2_{\rm min}(\rho) \in (\frac{1}{\rho}, \frac{K^2}{\rho})$. This completes the desired proof. 
\end{proof}

\begin{proof}[Proof of Theorem \ref{lemma:upper_bound_on_comm_rouds}]
Fix a confidence level $\delta \in (0,\frac{1}{4})$ arbitrarily, and let $\Pi$ be $\delta$-PAC and almost-optimal up to $\alpha \geq 1$.
Suppose, on the contrary, that
\begin{equation}
\limsup_{r\rightarrow \infty} \frac{b_{r+1}}{b_{r}} = \infty.
\label{eq:limsup-is-infinity}
\end{equation}
Then, there exists an increasing sequence $\{z_l\}_{l=1}^\infty$ such that
$\lim_{l\rightarrow \infty} \frac{b_{z_l}}{b_{z_l+1}} = 0$ and $b_{z_l} <b_{{z_l}+1} $ for all $l\in\mathbb{N}$.
Let $T^*_{\delta}(v) \coloneqq \log\left(\frac{1}{4\delta}\right) \, c^*(v) $. 
\noindent
Let 
\begin{equation}
    v^{(l)} \coloneqq v_{\dagger}\left(\frac{\sqrt{b_{z_l+1}b_{z_l}}}{4\log(\frac{1}{4\delta})}\right), \quad \mbox{for all} \; l \in \mathbb{N}.
    \label{eq:sequence-of-problem-instances}
\end{equation}
By Lemma \ref{lemma:construct_instances}, we then have
\begin{equation}
\label{eq:cstart_relation}
    \frac{\sqrt{b_{z_l+1}b_{z_l}}}{MK^2\log(\frac{1}{4\delta})}  \le c^*(v^{(l)}) \le \frac{K \sqrt{b_{z_l+1}b_{z_l}}}{\log(\frac{1}{4\delta})}.
\end{equation}
Also, we have
\begin{equation}
\label{eq:Tstart_relation}
\frac{\sqrt{b_{z_l+1}b_{z_l}}}{MK^2}  \le T^*_{\delta}(v^{(l)}) \le K \sqrt{b_{z_l+1}b_{z_l}} .
\end{equation}

\noindent
Let 
\begin{equation}
E_l \coloneqq \{\mbox{empirical best arms }  \hat{a}_\delta=a^*(v^{(l)})  \mbox{ and stopping time }  \tau_\delta(\Pi) \le b_{z_l} \}, \quad l \in \mathbb{N},
\label{eq:event-E-l}
\end{equation}
be the event that (a) $\hat{a}_\delta=(\hat{a}_{\delta,m})_{m\in [M]}$, the vector of the empirical best arms of the clients at confidence level $\delta$, equals the vector $a^*(v^{(l)})$, and (b) the stopping time $\tau_\delta(\Pi) \leq b_{z_l}$. 
From Lemma \ref{lemma:useful-lemma}, for any $l\in \mathbb{N}$, we have
\begin{equation}
\sum_{t=1}^{b_{z_l}}\  \sum_{m=1}^M \  \sum_{i \in S_m} \mathbb{E}_{v^{(l)}}^\Pi \left[\mathbf{1}_{\{A_t(m)=i\}}\  D_{\mathrm{KL}}(v_{i,m}^{(l)} \|v'_{i,m})\right] \ge d_{\mathrm{KL}}\left(\mathbb{P}_{v^{(l)}}^{\Pi}(E_l), \mathbb{P}_{{v'}}^{\Pi}(E_l)\right) 
\label{eq:useful-lemma-applied-to-vl-and-v-prime}
\end{equation}
for all $v' \in {\rm Alt}(v^{(l)})$. Note that 
\begin{align}
\mathbb{P}_{v^{(l)}}^\Pi(E_l) & = 1 - \mathbb{P}_{v^{(l)}}^\Pi(E_l^c)\nonumber\\
&\stackrel{(a)}{\ge} 1 - \mathbb{P}_{v^{(l)}}^\Pi(\hat{a}_\delta\ne a^*(v^{(l)})) - \mathbb{P}_{v^{(l)}}^\Pi(\tau_\delta(\Pi) > {b}_{z_l}) \nonumber \\
& \stackrel{(b)}{=} 1 - \mathbb{P}_{v^{(l)}}^\Pi(\hat{a}_\delta\ne a^*(v^{(l)})) - \mathbb{P}_{v^{(l)}}^\Pi(\tau_\delta(\Pi) \ge b_{z_l+1})  \nonumber\\
& \stackrel{(c)}{\ge} 1-\delta - \frac{\mathbb{E}_{v^{(l)}}^\Pi[\tau_\delta(\Pi)]}{b_{z_l+1}} \nonumber\\
& \stackrel{(d)}{\ge} 1-\delta - \frac{\alpha\ T^*(v^{(l)})}{b_{z_l+1}} \nonumber \\
& \stackrel{(e)}{\ge}  1-\delta - \alpha\,K\sqrt{\frac{b_{z_l}}{b_{z_l+1}}},
\end{align}
where $(a)$ above follows from the union bound, $(b)$ follows by noting that $\mathbb{P}_{v^{(l)}}^\Pi(\tau_\delta(\Pi) \ge b_{z_l+1}) = \mathbb{P}_{v^{(l)}}^\Pi(\tau_\delta(\Pi) > b_{z_l})$ as $\tau_\delta(\Pi) \in \{b_r\}_{r\in \mathbb{N}}$ and $b_{z_l} < b_{{z_l}+1}$, $(c)$ follows from Markov's inequality and the fact that  $\mathbb{P}_{v^{(l)}}^\Pi (\hat{a}_\delta\ne a^*(v)) \leq \delta$ as $\Pi$ is $\delta$-PAC, $(d)$ follows from the fact that $\mathbb{E}_{v^{(l)}}^\Pi[\tau_\delta(\Pi)] \leq \alpha \, T^*(v^{(l)})$ as $\Pi$ is almost-optimal up to the constant $\alpha$, and   (e) follows from \eqref{eq:Tstart_relation}. Because the algorithm $\Pi$ is $\delta$-PAC, it can be shown that $\mathbb{P}_{v'}^\Pi(E_l) \leq \delta$ for all $v' \in {\rm Alt}(v^{(l)})$. 

Continuing with \eqref{eq:useful-lemma-applied-to-vl-and-v-prime} and using the fact that $d_{\mathrm{KL}}(x,y) \geq \log\left(\frac{1}{4\,\delta'}\right)$ whenever $x\geq 1-\delta'$ and $y \leq \delta'$ (see, for instance, \cite{Kaufmann2016}), setting $\delta'=\delta + \alpha K \sqrt{b_{z_l}/b_{z_l+1}}$, we have 
\begin{align}
&\inf_{v'\in {\rm Alt}(v^{(l)})}\ \sum_{t=1}^{b_{z_l}}\  \sum_{m=1}^M \  \sum_{i \in S_m} \mathbb{E}_{v^{(l)}}^\Pi \left[\mathbf{1}_{\{A_t(m)=i\}} \ D_{\mathrm{KL}}(v_{i,m}^{(l)} \| v'_{i,m} ) \right] \ge \log \left(\frac{1}{4\delta+4\alpha\,K\sqrt{\frac{b_{z_l}}{b_{z_l+1}}}}\right)
 \nonumber \\
&\implies 
b_{z_l} \inf_{v'\in {\rm Alt}(v^{(l)})}\
 \sum_{t=1}^{b_{z_l}}\  \sum_{m=1}^M \ \sum_{i \in S_m} \frac{\mathbb{E}_{v^{(l)}}^\Pi \left[\mathbf{1}_{\{A_t(m)=i\}} \right]}{b_{z_l}} \cdot
 \frac{(\mu_{i,m}^{(l)}-\mu'_{i,m})^2}{2} \ge \log \left(\frac{1}{4\delta+4\alpha\,K\sqrt{\frac{b_{z_l}}{b_{z_l+1}}}}\right)
 \nonumber \\
& \stackrel{(a)}{\implies}
\frac{b_{z_l}}{c^*(v^{(l)})} \ge \log \left(\frac{1}{4\delta+4\alpha\,K\sqrt{\frac{b_{z_l}}{b_{z_l+1}}}}\right)
\nonumber \\
& \stackrel{(b)}{\implies}
MK^2\log \left(\frac{1}{4\delta}\right)\ \sqrt{b_{z_l}/b_{z_l+1}}\ge \log \left(\frac{1}{4\delta+4\alpha\,K\sqrt{\frac{b_{z_l}}{b_{z_l+1}}}}\right). 
\label{eq:contradiction}
\end{align}
In the above set of inequalities, $(a)$ follows from the definition of $c^*(v^{(l)})$, and $(b)$ follows from \eqref{eq:cstart_relation}. 
Letting $l \rightarrow \infty$ and using \eqref{eq:limsup-is-infinity}, we observe that the left-hand side of \eqref{eq:contradiction} converges to $0$, whereas the right-hand side converges to $\log(\frac{1}{4\delta})$, thereby resulting in $0 \geq \log(\frac{1}{4\delta})$, a contradiction. This proves that $\limsup_{r\rightarrow \infty} \frac{b_{r+1}}{b_{r}} < \infty$.
\end{proof}

\section{Proof of Lemma \ref{lemma:log-expectation-guarantee}}
\begin{proof}
Fix $\delta \in (0,\frac{1}{4})$ and $\beta \in (0,1)$ arbitrarily.
Let $\Pi$ be almost-optimal up to a constant, say $\alpha\ge 1$. Let $\{v^{(l)}\}_{l=1}^{\infty}$ be any sequence of problem instances such that $\lim_{l\to \infty} c^*(v^{(l)})=+\infty$, where this sequence must exist because of Lemma~\ref{lemma:construct_instances}. Let $T_l \coloneqq \big\lceil \left(\mathbb{E}_{v^{(l)}}^{\Pi}[\tau_{\delta}(\Pi)]\right)^\beta \big\rceil$. Because $\lim_{l\to \infty} c^*(v^{(l)})=+\infty$, we have
\begin{equation}
    \lim_{l \rightarrow \infty}\ \frac{T_l}{\mathbb{E}_{v^{(l)}}^\Pi[\tau_{\delta}(\Pi)]} = 0.
    \label{eq:contradiciton-1}
\end{equation}
For $l\in \mathbb{N}$, let 
\begin{equation}
F_l \coloneqq \{\mbox{empirical best arms } \hat{a}_\delta=a^*(v^{(l)}) \mbox{ and stopping time } \tau_\delta(\Pi) \le T_l \}
\label{eq:event-F-l}
\end{equation}
be the event that (a) the vector of empirical best arms matches with the vector of best arms under $v^{(l)}$, and (b) the stopping time $\tau_\delta(\Pi) \leq T_l$. Also, let $p_l \coloneqq \mathbb{P}_{v^{(l)}}^\Pi\left(\tau_\delta > T_l \right)$. From Lemma \ref{lemma:useful-lemma}, we know that
\begin{equation}
 \sum_{t=1}^{T_l}\  \sum_{m=1}^M \ \sum_{i \in S_m} \mathbb{E}_{v^{(l)}}^{\Pi} \left[\mathbf{1}_{\{A_t(m)=i\}}\  D_{\mathrm{KL}}(v_{i,m}^{(l)} \|v'_{i,m}) \right] \ge d_{\mathrm{KL}}\left(\mathbb{P}_{v^{(l)}}^\Pi(F_l), \mathbb{P}_{{v'}}^\Pi(F_l)\right) 
\end{equation}
for all problem instances $v'$. In particular, for $v' \in {\rm Alt}(v^{(l)})$, we note that for any $l\in \mathbb{N}$,
\begin{align}
\mathbb{P}_{v^{(l)}}^\Pi(F_l)  & \ge 1 - \mathbb{P}_{v^{(l)}}(\hat{a}_\delta\ne a^*(v^{(l)})) - \mathbb{P}_{v^{(l)}}(\tau_\delta > T_l) \nonumber \\
& \ge 1 - \delta - p_l. 
\end{align}
Along similar lines, it can be shown that $\mathbb{P}_{v'}^\Pi(F_l) \leq \delta+p_l$ for any $v' \in {\rm Alt}(v^{(l)})$. Then, using the fact that $d(x,y)\geq \log \left(\frac{1}{4\delta'}\right)$ whenever $x \geq 1-\delta'$ and $y \leq \delta'$, setting $\delta'=\delta + p_l$, we have
\begin{align}
& \inf_{v'\in {\rm Alt}(v^{(l)})}\ \sum_{t=1}^{T_l}\  \sum_{m=1}^M\  \sum_{i \in S_m} \mathbb{E}_{v^{(l)}}^\Pi \left[\mathbf{1}_{\{A_t(m)=i\}}\  D_{\mathrm{KL}}(v_{i,m}^{(l)} \| v'_{i,m}) \right] \ge \log \left(\frac{1}{4\delta+ 4p_l}\right)
 \nonumber \\
& \implies T_l
  \inf_{v'\in {\rm Alt}(v^{(l)})} 
 \sum_{t=1}^{T_l}\ \sum_{m=1}^M\ \sum_{i \in S_m}\ \frac{\mathbb{E}_{v^{(l)}}^\Pi \left[\mathbf{1}_{\{A_t(m)=i\}}  \right]}{T_l} \cdot \frac{(\mu_{i,m}^{(l)}-\mu'_{i,m})^2}{2} \ge \log \left(\frac{1}{4\delta+ 4p_l}\right)
 \nonumber \\
& \implies \frac{T_l}{c^*(v^{(l)})} \ge \log \left(\frac{1}{4\delta+ 4p_l}\right)
\label{eq:contradiction0} 
\end{align}
for all $l\in \mathbb{N}$, 
where the last line above follows from the definition of $c^*(v^{(l)})$. Because $\Pi$ is almost-optimal up to constant $\alpha\ge1$, we have 
\begin{equation}
 c^*(v^{(l)}) \log \left(\frac{1}{4\delta}\right)  \le \mathbb{E}_{v^{(l)}}^\Pi(\tau_{\delta}(\Pi)) \le \alpha\, c^*(v^{(l)}) \, \log \left(\frac{1}{4\delta}\right) \quad \text{for all }l\in \mathbb{N}.
 \label{eq:contradiction1}
\end{equation}
Combining \eqref{eq:contradiction0} and \eqref{eq:contradiction1}, we get
\begin{equation}
    \frac{T_l}{\mathbb{E}_{v^{(l)}}^\Pi[\tau_{\delta}(\Pi)]} \ge \frac{\log \left(\frac{1}{4\delta+ 4p_l}\right)}{\alpha \,\log \left(\frac{1}{4\delta}\right)} \quad \text{for all }l\in \mathbb{N}. 
    \label{eq:contradiction2}
\end{equation}

Suppose now that there exists $\epsilon \in \left(0, \frac{1}{4}-\delta\right)$ such that
\begin{equation}
\liminf_{l \rightarrow \infty}\  \mathbb{P}_{v^{(l)}}^\Pi \left( \log\left(\tau_{\delta}(\Pi) \right) > \beta \log \left(\mathbb{E}_{v^{(l)}}^\Pi \left[\tau_{\delta}(\Pi)\right] \right) \right)  \le \frac{1}{4} - \delta - \epsilon.
\label{eq:liminf-smaller-than-1/4-minus-delta-minus-epsilon}
\end{equation}
This implies from the definitions of $T_l$ and $p_l$ that there exists an increasing sequence $\{l_n:n\geq 1\}$ such that $p_{l_n} \le \frac{1}{4}-\delta -\epsilon $ for all $n\geq 1$. Using this in~\eqref{eq:contradiction2}, we get that
\begin{equation}
\limsup_{l \rightarrow \infty}\  \frac{T_l}{\mathbb{E}_{v^{(l)}}^\Pi \left[\tau_{\delta}(\Pi)\right]} \ge \limsup_{n \rightarrow \infty}\  \frac{T_{l_n}}{\mathbb{E}_{v^{(l_n)}}^\Pi \left[\tau_{\delta}(\Pi)\right]} \ge 
 \frac{\log \left(\frac{1}{1-4\epsilon}\right)}{\alpha \log \left(\frac{1}{4\delta}\right)} > 0,
 \label{eq:contradiction3}
\end{equation}
which clearly contradicts \eqref{eq:contradiciton-1}. This proves that there is no $\epsilon \in \left(0, \frac{1}{4}-\delta\right)$ such that \eqref{eq:liminf-smaller-than-1/4-minus-delta-minus-epsilon} holds, thereby establishing the desired result.
\end{proof}

\section{Proof of Theorem \ref{theorem:comm_complexity}}
\begin{proof}
Fix a sequence of problem instances $\{v^{(l)}\}_{l=1}^{\infty}$ with $\lim_{l\to\infty} c^*(v^{(l)})=+\infty$, a confidence level $\delta \in (0,\frac{1}{4})$, and an algorithm $\Pi$ that is almost optimal up to a constant, say $\alpha\ge 1$. From Theorem \ref{lemma:upper_bound_on_comm_rouds}, we know that there exists $\eta >0$ such that
\begin{equation}
    \label{eq:repeat_comm_round}
    \mathfrak{r}_\delta(\Pi) \ge \log_\eta(\tau_\delta(\Pi)) \quad \text{almost surely}.
\end{equation}
Also, from Lemma \ref{lemma:log-expectation-guarantee}, we know that for any $\beta \in (0,1)$ and any sequence of problem instances $\{v^{(l)}\}_{l=1}^\infty$ with $\lim_{l \rightarrow \infty} c^*(v^{(l)})=+\infty$,  
\begin{equation}
    \label{eq:repeat_log_guarantee}
    \liminf_{l\to\infty}\  \mathbb{P}_{v^{(l)}}^{\Pi}\left(\log\left(\tau_{\delta}(\Pi) \right) > \beta \log\left(\mathbb{E}_{v^{(l)}}^\Pi[\tau_{\delta}(\Pi)] \right) \right)  \ge \frac{1}{4} -\delta.
\end{equation}
Using \eqref{eq:repeat_comm_round} in \eqref{eq:repeat_log_guarantee}, we have
\begin{align}
    &\liminf_{l\to\infty}\ \mathbb{P}_{v^{(l)}}^\Pi\left(\mathfrak{r}_\delta(\Pi) >\beta\, \log_\eta \left(\mathbb{E}_{v^{(l)}}^\Pi [\tau_{\delta}(\Pi)] \right) \right)  \ge \frac{1}{4} -\delta \quad \forall \ \beta \in (0,1)
    \nonumber \\
    & \stackrel{(a)}{\implies} \liminf_{l\to\infty}\ \mathbb{P}_{v^{(l)}}^\Pi\left(\mathfrak{r}_\delta(\Pi) >\beta\, \log_\eta \Big(  \log\Big(\frac{1}{4\delta}\Big) c^*(v^{(l)}) \Big) \right)  \ge \frac{1}{4} -\delta \quad \forall \ \beta \in (0,1)
    \nonumber \\
    & \stackrel{(b)}{\implies} \liminf_{l\to\infty}\ \frac{\mathbb{E}_{v^{(l)}}^\Pi \left[\mathfrak{r}_\delta(\Pi) \right]}{\log_\eta \left( \,\log\left(\frac{1}{4\delta}\right) c^*(v^{(l)}) \right)} \geq \beta\,\left( \frac{1}{4} -\delta \right) \quad \forall \ \beta \in (0,1)
    \nonumber \\
    & \stackrel{(c)}{\implies} \liminf_{l\to\infty}\ \frac{\mathbb{E}_{v^{(l)}}^\Pi \left[\mathfrak{r}_\delta(\Pi) \right]}{\log_\eta \left( \,\log\left(\frac{1}{4\delta}\right) c^*(v^{(l)}) \right)} \geq \frac{1}{4} -\delta.
\end{align}
In the above set of inequalities, $(a)$ follows from Proposition \ref{prop:lower-bound} and the hypothesis that $\Pi$ is $\delta$-PAC, $(b)$ follows from Markov's inequality, and $(c)$ follows from $(b)$ by letting $\beta \to 1$. 
The desired result is thus established.
\end{proof}

\section{Proof of Theorem \ref{theorem:deltapac}}
Below, we record some important results that will be useful for proving Theorem \ref{theorem:deltapac}.
\begin{lemma}\cite[Lemma 33.8]{lattimore_szepesvari_2020}
\label{lemma:latimaore_338}
Let $Y_1,Y_2,\ldots$ be independent Gaussian random variables with mean $\mu$ and unit variance. Let $\hat{\mu}_n \coloneqq \frac{1}{n} \sum_{i=1}^n Y_i$. Then,
\[
\mathbb{P} \left( \exists\,  n \in \mathbb{N}: \frac{n}{2}(\hat{\mu}_n-\mu)^2 \ge \log(1/\delta) + \log(n(n+1)) \right) \le \delta.
\]
\end{lemma}
\begin{lemma}
\label{lemma:mydeltaunion}
Fix $n \in \mathbb{N}$. Let $Y_1,Y_2,\ldots,Y_n$ be independent random variables with $\mathbb{P}(Y_i \le y) \le y$ for all $y\in[0,1]$ and $i\in[n]$. Then, for any $\epsilon > 0$,
\begin{equation}
\label{eq:lemmaf}
\mathbb{P} \bigg( \sum_{i=1}^n \log(1/Y_i) \ge \epsilon \bigg) \le f_n(\epsilon)
\end{equation}
where $f_n:(0,+\infty) \rightarrow (0,1) $ is defined by 
\[
f_n(x) = \sum_{i=1}^{n} \frac{x^{i-1}e^{-x}}{(i-1)!}, \quad x \in (0, +\infty).
\]
\end{lemma}

\begin{proof}[Proof of Lemma \ref{lemma:mydeltaunion}]
First, for $i\in[n]$ we define the random variable $Z_i \coloneqq F_{i}(Y_i)$, where $F_{i}$ is the cumulative distribution function (CDF) of $Y_i$. Clearly, $Z_i$ is a uniform random variable. Notice that $\mathbb{P}(Y_i \le y) \le y = \mathbb{P}(Z_i \le y)$ for all $y\in (0,1)$, from which it follows that
\begin{equation*}
\mathbb{P} \bigg( \sum_{i=1}^n \log(1/Y_i) \ge \epsilon \bigg) \le \mathbb{P} \bigg( \sum_{i=1}^n \log(1/Z_i) \ge \epsilon \bigg).
\end{equation*}
Therefore, it suffices to prove Lemma~\ref{eq:lemmaf} for the case when $Y_1, \ldots, Y_n$ are independent and uniformly distributed on $[0,1]$. Suppose that this is indeed the case. Then, we note that $\mathbb{P} \big(\sum_{i=1}^n \log(1/Y_i) \ge \epsilon \big)=\mathbb{P} \big( \prod_{i=1}^n Y_i  \le \exp(-\epsilon) \big) $. Let $h_s(x) \coloneqq \mathbb{P} \big( \prod_{i=1}^s Y_i  \le x \big)$ for $s\in [n]$ and $x\in (0, 1)$. We then have
\begin{align}
h_1(x) &= x, \nonumber \\
\forall \ s>1, \quad h_s(x) &= \int_{0}^1 h_{s-1}\big(\min\{x/y, 1\}\big)\  \mathrm{d}y = x+ \int_{x}^1 h_{s-1}(x/y) \ \mathrm{d}y. \nonumber 
\end{align}
Using mathematical induction, we demonstrate below that 
\begin{equation}
    \label{eq:mitarget}
    h_s(x) = \sum_{i=1}^{s} \frac{(\log \frac{1}{x})^{i-1}x}{(i-1)!}
\end{equation}
for all $s \in [n]$ and $x\in (0,1)$.

\vspace{0.3cm}

\noindent{\bf Base case}: It is easy to verify that \eqref{eq:mitarget} holds for $s=1$. For $s=2$, we have
\begin{align}
 h_2(x) =x+ \int_{x}^1 h_1\left(\frac{x}{y}\right)\  \mathrm{d}y = x+ \int_{x}^1 \frac{x}{y}\  \mathrm{d}y = x + \log(1/x)\,x,
\end{align}
thus verifying that \eqref{eq:mitarget} holds for $s=2$.

\vspace{0.3cm}
\noindent
{\bf Induction step:} Suppose now that \eqref{eq:mitarget} holds for $s=k$ for some $k > 2$. Then,
\begin{align}
h_{k+1}(x) &  = x+ \int_{x}^1 h_{k}(x/y)\  \mathrm{d}y \nonumber \\
& \stackrel{(a)}{=}  x+ \int_{x}^1 \sum_{i=1}^{k} \frac{(\log \frac{y}{x})^{i-1} (\frac{x}{y}) }{(i-1)!} \ \mathrm{d}y \nonumber \\
& = x+  \sum_{i=1}^{k} \int_{x}^1 \frac{(\log \frac{y}{x})^{i-1} (\frac{x}{y}) }{(i-1)!} \ \mathrm{d}y \nonumber \\
& =x+  \sum_{i=1}^{k} \frac{x}{(i-1)!}\  \int_{x}^1  \frac{(\log \frac{y}{x})^{i-1}}{y}\ \mathrm{d}y \nonumber \\
& \stackrel{(b)}{=} x+  \sum_{i=1}^{k} \frac{x}{(i-1)!} \int_{1}^{1/x}  \frac{(\log y')^{i-1}}{y'} \ \mathrm{d}y' \nonumber \\
& \stackrel{(c)}{=}x+  \sum_{i=1}^{k} \frac{x}{(i-1)!} \frac{(\log \frac{1}{x})^i}{i} \nonumber \\
& = \sum_{i=1}^{k+1} \frac{(\log \frac{1}{x})^{i-1}x}{(i-1)!}, \label{eq:inductstep}
\end{align}
where $(a)$ follows from the induction hypothesis, in writing $(b)$ above, we set $y' = y/x$, and $(c)$ follows by noting that
\[
\int \frac{(\log y)^j}{y} \, \mathrm{d}y = \frac{1}{j+1} (\log y)^{j+1}.
\]
This demonstrates that \eqref{eq:inductstep} holds for $s=k+1$.

Finally, we note that
\begin{align}
    \mathbb{P} \bigg( \sum_{i=1}^n \log(1/Y_i) \ge \epsilon \bigg) & = h_n\big(\exp(-\epsilon)\big) \nonumber \\ 
    & = \sum_{i=1}^{n} \nonumber  \frac{\epsilon^{i-1}e^{-\epsilon}}{(i-1)!} \nonumber \\
    & = f_n(\epsilon),
\end{align}
thus establishing the desired result.
\end{proof}

With the above ingredients in place, we are now ready to prove Theorem \ref{theorem:deltapac}.

\begin{proof}[Proof of Theorem \ref{theorem:deltapac}]
Fix a confidence level $\delta \in (0,1)$ and a problem instance $v \in \mathcal{P}$ arbitrarily. We claim that $\tau_\delta(\Pi_{\mathrm{Het\mhyphen TS}}) < + \infty$ almost surely; a proof of this is deferred until the proof of Lemma \ref{lemma:stopping_time_finite}. Assuming that the preceding fact is true, for $m\in[M]$ and $i\in S_m$, let 
\[ \xi_{i,m} \coloneqq  \sup_{t \ge K} \frac{N_{i,m}(t)}{2}\big(\hat{\mu}_{i,m}(t)-\mu_{i,m}(v)\big)^2 - \log\big(N_{i,m}(t)(N_{i,m}(t)+1)\big).
\]
From Lemma \ref{lemma:latimaore_338}, we know that for any confidence level $\delta' \in (0,1)$,
\begin{align}
\mathbb{P}_v^{\Pi_{\mathrm{Het\mhyphen TS}}} (\xi_{i,m} \ge \log(1/\delta')) \le \delta'.
\end{align}
Let $\xi'_{i,m} \coloneqq \exp(-\xi_{i,m})$. Recall that $K'=\sum_{m=1}^M \lvert S_m \rvert$ From Lemma \ref{lemma:mydeltaunion}, we know that for any $\epsilon>0$,
\begin{align}
& \mathbb{P}_v^{\Pi_{\mathrm{Het\mhyphen TS}}} \left(\sum_{m\in [M]}\  \sum_{i\in S_m}\  \log(1/\xi'_{i,m}) \ge \epsilon \right) \le f_{K'}(\epsilon) \nonumber \\
& \stackrel{(a)}{\implies} \mathbb{P}_v^{\Pi_{\mathrm{Het\mhyphen TS}}} \left(\sum_{m\in [M]}\  \sum_{i\in S_m}\  \xi_{i,m} \ge \epsilon \right) \le f_{K'}(\epsilon) \nonumber \\
& \stackrel{(b)}{\implies} \mathbb{P}_v^{\Pi_{\mathrm{Het\mhyphen TS}}} \left(\sum_{m\in [M]}\  \sum_{i\in S_m}\  \xi_{i,m} \ge \epsilon \right) \le f(\epsilon) \nonumber \\
& \stackrel{(c)}{\implies} \mathbb{P}_v^{\Pi_{\mathrm{Het\mhyphen TS}}} \left(\sum_{m\in [M]}\  \sum_{i\in S_m}\  \xi_{i,m} \ge f^{-1}(\delta)  \right) \le \delta,  \label{eq:combinationbound} 
\end{align}
where $(a)$ above follows from the definition of $\xi'_{i,m}$,  $(b)$ follows from the definition of $f$ in \eqref{eq:deff}, and in writing $(c)$, we (i) make use of the fact that $f$ is continuous and strictly decreasing and therefore admits an inverse, and (ii) set $\epsilon = f^{-1}(\delta)$.
Eq. \eqref{eq:combinationbound} then implies
 
\begin{align}
& \mathbb{P}_v^{\Pi_{\mathrm{Het\mhyphen TS}}} \bigg(\forall\, t\ge K \  \sum_{m\in [M]} \sum_{i\in S_m}
\frac{N_{i,m}(t)}{2}\big(\hat{\mu}_{i,m}(t)-\mu_{i,m}(v)\big)^2  \le  K'\log\big(t(t+1)\big) + f^{-1}(\delta) \bigg) \ge 1-\delta \nonumber \\ 
& \implies \mathbb{P}_v^{\Pi_{\mathrm{Het\mhyphen TS}}} \bigg(\forall\, t\ge K \ \sum_{m\in [M]} \sum_{i\in S_m}
\frac{N_{i,m}(t)}{2}\big(\hat{\mu}_{i,m}(t)-\mu_{i,m}(v)\big)^2  \le  \beta(t, \delta) \bigg) \ge 1-\delta.
\label{eq:importanceforpac}
\end{align}
Note that at the stopping time $\tau_\delta(\Pi_{\mathrm{Het\mhyphen TS}})$, we must have
\[
\inf_{v' \in {\rm Alt}(\hat{v}(\tau_\delta))} \sum_{m \in [M]} \sum_{i\in S_m} N_{i,m}(\tau_\delta) \frac{(\mu_{i,m}(v') - \hat{\mu}_{i,m}(\tau_\delta))^2}{2} > \beta(\tau_\delta, \delta).
\]
Thus, we may write \eqref{eq:importanceforpac} equivalently as $\mathbb{P}_v^{\Pi_{\mathrm{Het\mhyphen TS}}} \left( v \notin {\rm Alt} \left(\hat{v}(\tau_\delta(\Pi_{\mathrm{Het\mhyphen TS}}))\right) \right) \ge 1 - \delta$, which is identical to $\mathbb{P}_v^{\Pi_{\mathrm{Het\mhyphen TS}}}\left( a^*(v) = a^*(\hat{v}(\tau_\delta(\Pi_{\mathrm{Het\mhyphen TS}}))) \right) \ge 1 - \delta$. This completes the proof.
\end{proof}

\section{Proof of Theorem \ref{Theorem:mainresult}}
\label{appndx:proof-of-theorem-upper-bound}
We first state two results that will be used later in the proof of Theorem \ref{Theorem:mainresult}.
\begin{lemma}\cite{ sundaram1996first,yang2022optimal}
\label{lemma:sundaram}
Let $f: S \times  \Theta \to \mathbb R $ be a continuous function, and $\mathcal{D}:\Theta \rightrightarrows S$ be a compact-valued continuous correspondence. Let $f^*: \Theta \to \mathbb R $ and $D^*: \Theta \rightrightarrows S $ be defined by
$$
f^{*}(\theta)=\max \{f(x, \theta) : x \in \mathcal{D}(\theta)\}
$$
and
$$
\mathcal{D}^{*}(\theta)=\argmax \{f(x, \theta) : x \in \mathcal{D}(\theta)\}=\{x \in \mathcal{D}(\theta):f(x, \theta)=f^{*}(\theta)\}.
$$
Then $f^{*}$ is a continuous function on $\Theta$, and $\mathcal{D}^{*}$ is a compact-valued, upper hemicontinuous correspondence on $\Theta$.

\end{lemma}

\begin{lemma}\cite[Lemma 17.6]{guide2006infinite}
\label{lemma:Guide}
A singleton-valued correspondence is upper hemicontinuous if
and only if it is lower hemicontinuous, in which case it is continuous as a function.
\end{lemma}

\begin{lemma}
\label{lemma:f_delta}
Let $f$ be as defined in \eqref{eq:deff}. Then, $f^{-1}(\delta)=(1+o(1)) \log(1/\delta)$ as $\delta \rightarrow 0$, i.e., 
\begin{equation}
    \lim_{\delta \rightarrow 0} \frac{\log(1/\delta)}{f^{-1}(\delta)} =1.
    \label{eq:f_delta}
\end{equation}
\end{lemma}
\begin{proof}
Let $x = f^{-1}(\delta)$. Then,
\begin{align}
\lim_{\delta \rightarrow 0} \frac{\log(1/\delta)}{f^{-1}(\delta)}
   & \stackrel{(a)}{=} \lim_{x \rightarrow +\infty} \frac{\log(\frac{1}{f(x)})}{x}  \nonumber \\
    & \stackrel{(b)}{=} \lim_{x \rightarrow +\infty} \frac{-f'(x)/f(x)}{1}  \nonumber \\
     & \stackrel{(c)}{=} \lim_{x \rightarrow +\infty} \frac{ \frac{x^{K'-1}e^{-x}}{(K'-1)!}}{\sum_{i=1}^{K'} \frac{x^{i-1}e^{-x}}{(i-1)!}} \nonumber \\
      & = \lim_{x \rightarrow +\infty} \frac{ \frac{x^{K'-1}}{(K'-1)!}}{\sum_{i=1}^{K'} \frac{x^{i-1}}{(i-1)!}} \nonumber \\
      & = 1,
      \label{eq:proof-f_delta}
\end{align}
where $(a)$ above follows the fact that $x \rightarrow \infty$ as $\delta \rightarrow 0$, $(b)$ follows from the L'Hospital's rule, and $(c)$ makes use of the fact that $f'(x)=\frac{-x^{K'-1}e^{-x}}{(K'-1)!}$.
This completes the proof.
\end{proof}

Before proceeding further, we introduce some additional notations.
For any $j \in [L]$ and $m \in [M]$, let 
\begin{equation}
    \Lambda_m^{(j)} \coloneqq \begin{cases}
          {\Lambda}_m, & \text{if }S_m \subseteq Q_j, \\
          \{\mathbf{0}^K\}, & \text{otherwise},
    \end{cases}
\end{equation}
where $\mathbf{0}^K$ denotes the all-zeros vector of dimension $K$.
For each $j\in [L]$, noting that $ \prod_{i=1}^M \Lambda_i^{(j)}:=\Lambda_1^{(j)}\times\ldots\Lambda_M^{(j)}$ is compact and that the mapping $\omega \mapsto \widetilde{g}^{(j)}_v(\omega)$ is continuous, there exists a solution to $\max_{\omega \in \prod_{i=1}^M \Lambda_i^{(j)}} \widetilde{g}^{(j)}_v(\omega)$. Let
\begin{equation}
    \widetilde{\omega}^{(j)}(v) \in \argmax_{\omega \in\prod_{i=1}^M \Lambda_i^{(j)} } \widetilde{g}^{(j)}_v(\omega),
\end{equation}
Further, let $\widetilde{\omega}(v) \coloneqq \sum_{j=1}^L \widetilde{\omega}^{(j)}(v)$. Then, it is easy to verify that $\widetilde{\omega} (v) \in \Gamma $ is a common solution to 
\begin{equation}
\label{eq:exist_solution}
    \max_{\omega \in \Gamma} \widetilde{g}_v(\omega), \max_{\omega \in \Gamma} \widetilde{g}^{(1)}_v(\omega), \ldots, \max_{\omega \in \Gamma} \widetilde{g}^{(L)}_v(\omega).
\end{equation}
Note that such common solution above is unique (we defer the proof of this fact to Theorem \ref{theorem:unique-allocation}), which then implies that the solution to $\argmax_{\omega \in\prod_{i=1}^M \Lambda_i^{(j)} } \widetilde{g}^{(j)}_v(\omega)$ is unique. Hence, $\widetilde{\omega}^{(j)}(v)$ and $\widetilde{\omega}(v)$ are well-defined.
\begin{lemma}
\label{lemma:emprical_allocation_equal_optimal}
Given any problem instance $v\in \mathcal{P}$, under $\Pi_{\mathrm{Het\mhyphen TS}}$,
\begin{equation}
\label{eq:almost_sure_omega}
\lim_{t \rightarrow \infty} \lVert \widetilde{\omega}(\hat{v}(t)) - \widetilde{\omega}(v) \rVert_\infty =0 \quad \text{almost surely}.
\end{equation}
Consequently, for any $m
\in [M]$ and $i\in S_m$, 
\begin{equation}
\label{eq:almost_sure_N}
    \lim_{t \rightarrow \infty} \left\lvert \frac{N_{i,m}(t)}{t} - \widetilde{\omega}(v)_{i,m} \right\rvert =0 \quad \text{almost surely}.
\end{equation}
\end{lemma}
\begin{proof}
Fix $j\in[L]$ and $v\in \mathcal{P}$ arbitrarily.
By the strong law of
large numbers, it follows that for any $i\in[K]$ and $m\in S_m$,
\begin{align}
& \lim_{t \rightarrow \infty}  \hat{\mu}_{i,m}(t) = \mu_{i,m}(v) \quad \text{almost surely} \nonumber \\
& \implies \lim_{t \rightarrow \infty} \hat{\mu}_{i}(t) = \mu_{i}(v) \quad \text{almost surely}   \label{eq:almost_sure_mu_i} \\
& \implies \lim_{t \rightarrow \infty}  \Delta_i(\hat{v}(t)) = \Delta_i(v) \quad \text{almost surely}. \label{eq:dieta_contin}
\end{align}
For any $v'\in \mathcal{P}$, note that $\widetilde{g}^{(j)}_{v'}(\omega)$ is a function of $(\Delta(v'),\omega)$ for $\Delta(v') \in (\mathbb{R^+})^K$ and $\omega \in\prod_{i=1}^M \Lambda_i^{(j)}$. From  Lemma~\ref{lemma:sundaram} and Lemma~\ref{lemma:Guide} that for any $\epsilon_1 >0$, there exists $\epsilon_2 >0$ such that for all $v' \in \mathcal{P}$ with $\lVert \Delta(v)-\Delta(v') \rVert_\infty \le \epsilon_2$,
\begin{equation}
\label{eq:maxcontinue}
    \lVert \widetilde{\omega}^{(j)}(v) - \widetilde{\omega}^{(j)}(v') \rVert_\infty \le \epsilon_1.
\end{equation}
Combining \eqref{eq:dieta_contin} and \eqref{eq:maxcontinue}, it follows that
\[
\lim_{t \rightarrow \infty} \lVert \widetilde{\omega}^{(j)}(v) - \widetilde{\omega}^{(j)}(\hat{v}(t)) \rVert_\infty =0 \quad \text{almost surely},
\]
which in turn implies that 
\begin{equation}
\label{eq:repeat_almost_sure_omega}
\lim_{t \rightarrow \infty} \lVert \widetilde{\omega}(v) - \widetilde{\omega}(\hat{v}(t)) \rVert_\infty =0
\end{equation} almost surely.
Recall the definition of $\hat{\omega}_{i,m}(t)$ in~\eqref{eq:sampling-rule-at-client-m}, which means $\hat{\omega}_{i,m}(t) =\widetilde{\omega}_{i,m}\big(\hat{v}(b_{r(t)})\big) $.
Then, by~\eqref{eq:repeat_almost_sure_omega} for any $m
\in [M]$ and $i\in S_m$ we have
\begin{equation}
\label{eq:repeat_almost_sure_omega2}
\lim_{t \rightarrow \infty} \lVert \widetilde{\omega}(v) - \hat{\omega}(t) \rVert_\infty =0
\end{equation}
 almost surely.

Consequently, by \cite[Lemma 17]{garivier2016optimal}, for any $m
\in [M]$ and $i\in S_m$, 
\begin{equation}
\label{eq:almost_sure_N_in_proof}
    \lim_{t \rightarrow \infty} \left\lvert \frac{N_{i,m}(t)}{t} - \widetilde{\omega}(v)_{i,m} \right\rvert =0 \quad \text{almost surely}.
\end{equation}
\end{proof}
\begin{lemma}
\label{lemma:zt_div_t}
Given any problem instance $v\in \mathcal{P}$, under $\Pi_{\mathrm{Het\mhyphen TS}}$,
\begin{equation}
\lim_{t \rightarrow \infty} \frac{Z(t)}{t} = g_v(\widetilde{\omega}(v))
\quad \text{almost surely}.
\end{equation}
\end{lemma}
\begin{proof}
Fix a problem instance $v\in \mathcal{P}$ arbitrarily. Define $\hat{N}(t) \in \Gamma$ as 
\begin{equation}
    \hat{N}_{i,m}(t) \coloneqq \frac{N_{i,m}(t)}{ t}, \quad i \in S_m, \ m \in [M]
    \label{eq:hat-N-im-t}.
\end{equation}
Then,
\begin{align}
\frac{Z(t)}{t} & = \inf_{v' \in {\rm Alt}(\hat{v}(t))}\  \sum_{m=1}^M \ \sum_{i\in S_m} \frac{N_{i,m}(t)}{t}\  \frac{(\mu_{i,m}(v') - \hat{\mu}_{i,m}(t))^2}{2} \nonumber \\
 & = \inf_{v' \in {\rm Alt}(\hat{v}(t))}\  \sum_{m=1}^M \ \sum_{i\in S_m} {\hat{N}_{i,m}(t)}\  \frac{(\mu_{i,m}(v') - \hat{\mu}_{i,m}(t))^2}{2} \nonumber \\
  & = g_{\hat{v}(t)}(\hat{N}(t)) \nonumber \\
  & \stackrel{(a)}{=} \min_{(i,j) \in \mathcal{C}(\hat{v}(t))} \frac{\big(\hat{\mu}_i(t)-\hat{\mu}_j(t)\big)^2/2}{\frac{1}{M^2_i}\sum_{m=1}^M  \mathbf{1}_{\{i \in S_m\}}\frac{1}{\hat{N}_{i,m}(t)} +\frac{1}{M^2_j}\sum_{m=1}^M  \mathbf{1}_{\{j \in S_m\}}\frac{1}{\hat{N}_{j,m}(t)} }, \label{eq:zt_div_t}
\end{align}
where $(a)$ follows from \eqref{eq:g_v(omega)-simplification} of Lemma \ref{lemma:relaxed-g}.
Because $v \in \mathcal{P}$ and $\lim_{t \rightarrow \infty} \hat{\mu}_{i}(t) = \mu_{i}\ \text{almost surely}$ for all $i\in [K]$ from (\ref{eq:almost_sure_mu_i}), we get that
\begin{equation}
\label{eq:almost_sure_ct}
    \lim_{t \rightarrow \infty}  \mathcal{C}(\hat{v}(t)) = \mathcal{C}(v) \quad \text{almost surely},
\end{equation}
where $\mathcal{C}(\cdot)$ is as defined in \eqref{eq:C(v)}. Combining \eqref{eq:almost_sure_omega}, \eqref{eq:almost_sure_N}, \eqref{eq:almost_sure_mu_i}, \eqref{eq:almost_sure_ct}, and Lemma \ref{lemma:emprical_allocation_equal_optimal}, we get that almost surely,
\begin{align}
    \lim_{t \to \infty}\ \frac{Z(t)}{t} & = \lim_{t\rightarrow \infty}
\min_{(i,j) \in \mathcal{C}(\hat{v}(t))} \frac{\big(\hat{\mu}_i(t)-\hat{\mu}_j(t)\big)^2/2}{\frac{1}{M^2_i}\sum_{m=1}^M  \mathbf{1}_{\{i \in S_m\}}\frac{1}{\hat{N}_{i,m}(t)} +\frac{1}{M^2_j}\sum_{m=1}^M  \mathbf{1}_{\{j \in S_m\}}\frac{1}{\hat{N}_{j,m}(t)} }  \nonumber \\
& = 
\min_{(i,j) \in \mathcal{C}(v)} \frac{\big({\mu}_i(v)-{\mu}_j(v)\big)^2/2}{\frac{1}{M^2_i}\sum_{m=1}^M  \mathbf{1}_{\{i \in S_m\}}\frac{1}{\widetilde{\omega}_{i,m}(v)} +\frac{1}{M^2_j}\sum_{m=1}^M  \mathbf{1}_{\{j \in S_m\}}\frac{1}{\widetilde{\omega}_{j,m}(v)} }  \nonumber \\
&  = g_{v}(\widetilde{\omega}(v)).
\end{align}
This completes the desired proof.
\end{proof}

\begin{lemma}
\label{lemma:stopping_time_finite}
Given any confidence level $\delta\in(0,1)$,
\begin{equation}
\tau_\delta(\Pi_{\mathrm{Het\mhyphen TS}}) < +\infty \quad \text{almost surely}.
\end{equation}
\end{lemma}
\begin{proof}
As a consequence of Lemma \ref{lemma:zt_div_t}, we have
\begin{align}
   \lim_{t \rightarrow \infty} \frac{\beta(t, \delta)}{Z(t)}
   & =  \lim_{t \rightarrow \infty} \frac{K'\log(t^2+t)+ f^{-1}(\delta)}{t \frac{Z(t)}{t}} \nonumber \\
   & =  \lim_{t \rightarrow \infty} \frac{K'\log(t^2+t)+ f^{-1}(\delta)}{t g_v(\widetilde{\omega}(v))} \nonumber \\
   & =  0 \quad \text{almost surely}. 
\end{align}
Therefore, there almost surely exists $0<T<+\infty$ such that $Z(t) > \beta(t, \delta)$ for all $t \ge T$, thus proving that $\tau_\delta(\Pi_{\textsc{Het}\mhyphen\textsc{TS}})$ is finite almost surely.
\end{proof}
\begin{lemma}
\label{lemma:Tlast}
Given any problem instance $v \in \mathcal{P}$ and  $\epsilon \in \left(0, g_v(\widetilde{\omega}(v)) \right)$, there exists  $\delta_{\rm upper}(v,\epsilon) >0$ such that for any $\delta \in (0, \delta_{\rm upper}(v,\epsilon))$,
\begin{equation}
\label{eq:laststoptime}
    t\,g_v(\widetilde{\omega}(v)) > \beta(t, \delta) + t\,\epsilon
\end{equation}
for all $t \ge T_{\rm last}(v,\delta, \epsilon)$,
where
\begin{equation}
\label{eq:denotetlast}
T_{\rm last}(v,\delta, \epsilon) \coloneqq \frac{f^{-1}(\delta)}{g_v(\widetilde{\omega}(v)) - \epsilon} + K'\log\left(\left(\frac{2f^{-1}(\delta)}{g_v(\widetilde{\omega}(v)) - \epsilon}\right)^2 + \frac{2f^{-1}(\delta)}{g_v(\widetilde{\omega}(v)) - \epsilon}  \right) \frac{1}{g_v(\widetilde{\omega}(v)) - \epsilon} +1.
\end{equation}
\end{lemma}
\begin{proof}
Fix $v \in \mathcal{P}$ and  $\epsilon \in \left(0, g_v(\widetilde{\omega}(v)) \right)$ arbitarily.
Recall that $\beta(t, \delta) = K'\log(t^2+t) + f^{-1}(\delta)$. 

To prove Lemma \ref{lemma:Tlast}, it suffices to verify that
\begin{enumerate}
    \item The derivative of the left-hand of \eqref{eq:laststoptime} with respect to $t$ is greater than that of the right-hand side of \eqref{eq:laststoptime} for all $t \geq T_{\rm last}(v, \delta, \epsilon)$, and
    \item Eq. \eqref{eq:laststoptime} holds for $t=T_{\rm last}(v, \delta, \epsilon)$.
\end{enumerate}
In order to verify that the condition in item $1$ above holds, we note from Lemma \ref{lemma:f_delta} that $\lim_{\delta \to 0} f^{-1}(\delta)=+\infty$, as a consequence of which we get that there exists $\delta_{\rm upper}(v, \epsilon)>0$ such that for all $\delta \in (0, \delta_{\rm upper}(v, \epsilon))$, 
\begin{equation*}
\label{eq:T_last_succ_1}
    T_{\rm last}(v, \delta, \epsilon) < \frac{2f^{-1}(\delta)}{g_v(\widetilde{\omega}(v)) - \epsilon} 
\end{equation*}
and
$$  
\frac{f^{-1}(\delta)}{g_v(\widetilde{\omega}(v)) - \epsilon} \geq  \frac{3K'}{g_v(\widetilde{\omega}(v)) - \epsilon}.
$$
Notice that the derivative of the left-hand side of \eqref{eq:laststoptime} with respect to $t$ is equal to $g_v(\widetilde{\omega}(v))$, whereas that of the right-hand side of \eqref{eq:laststoptime} is equal to $\frac{K'(2+\frac{1}{t})}{t + 1}+\epsilon$. Hence, to verify the condition in item $1$, we need to demonstrate that
\begin{equation}
    g_v(\widetilde{\omega}(v))-\epsilon > \frac{K'(2+\frac{1}{t})}{t + 1} \quad \text{for all } t \geq T_{\rm last}(v,\delta, \epsilon).
    \label{eq:condition-1}
\end{equation}
We note that for all $t \geq T_{\rm last}(v,\delta, \epsilon)$,
\begin{align}
    t+1 & > t\nonumber\\
    & \geq T_{\rm last}(v,\delta, \epsilon)\nonumber\\
    & \geq \frac{f^{-1}(\delta)}{g_v(\widetilde{\omega}(v)) - \epsilon}\nonumber\\
    & \geq  \frac{3K'}{g_v(\widetilde{\omega}(v)) - \epsilon} \nonumber\\
    & \geq \frac{(2+\frac{1}{t})K'}{g_v(\widetilde{\omega}(v)) - \epsilon},
    \label{eq:proof-of-condition-in-item-1}
\end{align}
where in writing the last line above, we use the fact that $3 \geq 2+\frac{1}{t}$ whenever $t \geq 1$. We then obtain \eqref{eq:condition-1} upon rearranging \eqref{eq:proof-of-condition-in-item-1} and using the fact that $\epsilon>0$. This verifies the condition in item $1$. To verify the condition in item $2$ above, we note that for all $T_{\rm last}(v, \delta, \epsilon) \leq t \leq \frac{2f^{-1}(\delta)}{g_v(\widetilde{\omega}(v)) - \epsilon}$, we have
\begin{align}
    t & \geq T_{\rm last}(v, \delta, \epsilon)\nonumber\\
    & > \frac{f^{-1}(\delta)}{g_v(\widetilde{\omega}(v)) - \epsilon} + K'\log\left(\left(\frac{2f^{-1}(\delta)}{g_v(\widetilde{\omega}(v)) - \epsilon}\right)^2 + \frac{2f^{-1}(\delta)}{g_v(\widetilde{\omega}(v)) - \epsilon}  \right) \frac{1}{g_v(\widetilde{\omega}(v)) - \epsilon} \nonumber\\
    & \geq \frac{f^{-1}(\delta)}{g_v(\widetilde{\omega}(v)) - \epsilon} + K'\log\left(t^2 +t  \right) \frac{1}{g_v(\widetilde{\omega}(v)) - \epsilon}.
    \label{eq:proof-of-condition-in-item-2}
\end{align}
Equivalently, upon rearranging the terms in \eqref{eq:proof-of-condition-in-item-2}, we get
\begin{equation}
    t\,g_v(\widetilde{\omega}(v)) > \beta(t, \delta) + t\,\epsilon 
    \label{eq:proof-of-condition-in-item-2-1}
\end{equation}
for all $T_{\rm last}(v, \delta, \epsilon) \leq t \leq \frac{2f^{-1}(\delta)}{g_v(\widetilde{\omega}(v)) - \epsilon}$. In particular, noting that \eqref{eq:proof-of-condition-in-item-2-1} holds for $t=T_{\rm last}(v, \delta, \epsilon)$ verifies the condition in item $2$, and thereby completes the proof.

\end{proof}

With the above ingredients in place, we are now ready to prove Theorem \ref{Theorem:mainresult}.

\begin{proof}[Proof of Theorem \ref{Theorem:mainresult}]
Fix a problem instance $v\in \mathcal{P}$ arbitrarily. Given any $\epsilon>0$, let $T_{\rm cvg}(v,\epsilon)$ denote the smallest positive integer such that
\begin{equation}
     \left\lvert \frac{Z(t)}{t} -g_v(\widetilde{\omega}(v))  \right\rvert \le \epsilon \quad \forall \ t \ge T_{\rm cvg}(v,\epsilon).
     \label{eq:T-cvg-definition}
\end{equation}
From Lemma \ref{lemma:zt_div_t}, we know that $T_{\rm cvg}(v,\epsilon) < +\infty$ almost surely. Therefore, for any $\epsilon \in \left(0, g_v(\widetilde{\omega}(v)) \right)$ and $\delta \in (0, \delta_{\rm upper}(v,\epsilon))$, it follows from Lemma \ref{lemma:Tlast} that
\begin{equation}
    Z(t) > \beta(t, \delta) \quad \forall\, t \ge \max\big\{T_{\rm cvg}(v,\epsilon),T_{\rm last}(v,\delta, \epsilon), K\big\} \quad  \text{almost surely}, 
    \label{eq:real_stop_time}
\end{equation}
where $\delta_{\rm upper}(v,\epsilon)$ and $T_{\rm last}(v,\delta, \epsilon)$ are as defined in Lemma \ref{lemma:Tlast}.
Recall that $b_r = \lceil (1+\lambda)^r \rceil$ in the {\rm Het-TS} algorithm. From \eqref{eq:real_stop_time}, it follows that
\begin{equation*}
    \tau_{\delta}(\Pi_{\mathrm{Het\mhyphen TS}}) \le (1+\lambda) \max\big\{T_{\rm cvg}(v,\epsilon),T_{\rm last}(v,\delta, \epsilon), K \big\} +1 \quad \text{almost surely}
\end{equation*}
for any $\epsilon \in \left(0, g_v(\widetilde{\omega}(v)) \right) $ and $\delta \in (0, \delta_{\rm upper}(v,\epsilon))$, which implies that 
\begin{equation}
\label{eq:usefulstaementforstoppingtime}
    \tau_{\delta}(\Pi_{\mathrm{Het\mhyphen TS}}) \le (1+\lambda) \,T_{\rm cvg}(v,\epsilon) + (1+\lambda)\, T_{\rm last}(v,\delta, \epsilon) + (1+\lambda)K +1 \quad \text{almost surely}.
\end{equation}
Then, for any $\epsilon \in \left(0, g_v(\widetilde{\omega}(v)) \right) $, the following set of relations hold almost surely:
\begin{align}
    & \limsup_{\delta \rightarrow 0} \frac{\tau_{\delta}(\Pi_{\mathrm{Het\mhyphen TS}})}{\log\left(\frac{1}{\delta}\right)} \nonumber \\
    &\le \limsup_{\delta \rightarrow 0} \frac{(1+\lambda)\, T_{\rm cvg}(v,\epsilon) + (1+\lambda) \, T_{\rm last}(v,\delta, \epsilon) + (1+\lambda)K +1 }{\log\left(\frac{1}{\delta}\right)} \nonumber \\
    &\stackrel{(a)}{=} \limsup_{\delta \rightarrow 0} \frac{ (1+\lambda)\, T_{\rm last}(v,\delta, \epsilon) }{\log\left(\frac{1}{\delta}\right)} \nonumber \\
    &\stackrel{(b)}{=} \limsup_{\delta \rightarrow 0} \frac{ (1+\lambda)\, \left(\frac{f^{-1}(\delta)}{g_v(\widetilde{\omega}(v)) - \epsilon} + K'\log\left(\left(\frac{2f^{-1}(\delta)}{g_v(\widetilde{\omega}(v)) - \epsilon}\right)^2 + \frac{2f^{-1}(\delta)}{g_v(\widetilde{\omega}(v)) - \epsilon}  \right) \frac{1}{g_v(\widetilde{\omega}(v)) - \epsilon} +1 \right) }{\log\left(\frac{1}{\delta}\right)} \nonumber \\
    & \stackrel{(c)}{=} \frac{1+\lambda}{g_v(\widetilde{\omega}(v)) - \epsilon} \nonumber \\
    & \stackrel{(d)}{\le} \frac{1+\lambda}{\frac{1}{2}c^*(v)^{-1} - \epsilon},
    \label{eq:resultwithepsilon}
\end{align}
where $(a)$ follows from the fact that $T_{\rm cvg}(v,\epsilon)$ is not a function of $\delta$ and that $T_{\rm cvg}(v,\epsilon) < +\infty$ almost surely, $(b)$ follows from the definition of $T_{\rm last}(v,\delta, \epsilon)$, $(c)$ follows from Lemma \ref{lemma:f_delta}, and $(d)$ makes use of Lemma \ref{lemma:relaxed-g}.
Letting $\epsilon \rightarrow 0$ in \eqref{eq:resultwithepsilon}, we get 
\[
\limsup_{\delta \rightarrow 0} \frac{\tau_{\delta}(\Pi_{\mathrm{Het\mhyphen TS}})}{\log\left(\frac{1}{\delta}\right)} \le 2\, (1+\lambda)\, c^*(v) \quad \text{almost surely}.
\]
Taking expectation on both sides of \eqref{eq:usefulstaementforstoppingtime}, we get 
\[
\mathbb{E}_v^{\Pi_{\mathrm{Het\mhyphen TS}}} [\tau_{\delta}(\Pi_{\mathrm{Het\mhyphen TS}})] \le (1+\lambda) \, \mathbb{E}_v^{\Pi_{\mathrm{Het\mhyphen TS}}} [T_{\rm cvg}(v,\epsilon)] + (1+\lambda)\, T_{\rm last}(v,\delta, \epsilon) +1
\]
for all $\epsilon \in \left(0, g_v(\widetilde{\omega}(v)) \right) $ and $\delta \in (0, \delta_{\rm upper}(v,\epsilon))$, from which it follows that
\begin{align}
    & \limsup_{\delta \rightarrow 0} \frac{\mathbb{E}_v^{\Pi_{\mathrm{Het\mhyphen TS}}} [\tau_{\delta}(\Pi_{\mathrm{Het\mhyphen TS}})]}{\log\left(\frac{1}{\delta}\right)} \nonumber \\
    &\le \limsup_{\delta \rightarrow 0} \frac{(1+\lambda)\, \mathbb{E}_v^{\Pi_{\mathrm{Het\mhyphen TS}}} [T_{\rm cvg}(v,\epsilon)] + (1+\lambda) \, T_{\rm last}(v,\delta, \epsilon) + (1+\lambda)K +1 }{\log\left(\frac{1}{\delta}\right)} \nonumber \\
    &\stackrel{(a)}{=} \limsup_{\delta \rightarrow 0} \frac{ (1+\lambda) \, T_{\rm last}(v,\delta, \epsilon) }{\log\left(\frac{1}{\delta}\right)} \nonumber \\
    &\stackrel{(b)}{=} \limsup_{\delta \rightarrow 0} \frac{ (1+\lambda)\left(\frac{f^{-1}(\delta)}{g_v(\widetilde{\omega}(v)) - \epsilon} + K'\log\left(\left(\frac{2f^{-1}(\delta)}{g_v(\widetilde{\omega}(v)) - \epsilon}\right)^2 + \frac{2f^{-1}(\delta)}{g_v(\widetilde{\omega}(v)) - \epsilon}  \right) \frac{1}{g_v(\widetilde{\omega}(v)) - \epsilon} +1 \right) }{\log\left(\frac{1}{\delta}\right)} \nonumber \\
    & \stackrel{(c)}{=} \frac{1+\lambda}{g_v(\widetilde{\omega}(v)) - \epsilon} \nonumber \\
    & \stackrel{(d)}{\le} \frac{1+\lambda}{\frac{1}{2}c^*(v)^{-1} - \epsilon},
    \label{eq:anotherresultwithepsilon}
\end{align}
where $(a)$ follows from the fact that $\mathbb{E}_v^{\Pi_{\mathrm{Het\mhyphen TS}}} [T_{\rm cvg}(v,\epsilon)]$ does not depend on $\delta$ and that $\mathbb{E}_v^{\Pi_{\mathrm{Het\mhyphen TS}}}[T_{\rm cvg}(v,\epsilon)] < +\infty$ because of the following Lemma~\ref{lemma:Tcvg}, $(b)$ follows from the definition of $T_{\rm last}(v,\delta, \epsilon)$, $(c)$ follows from Lemma \ref{lemma:f_delta}, and $(d)$ makes use of Lemma \ref{lemma:relaxed-g}. Letting $\epsilon \rightarrow 0$ in~\eqref{eq:anotherresultwithepsilon}, we get 
\[
\limsup_{\delta \rightarrow 0} \frac{\mathbb{E}_v^{\Pi_{\mathrm{Het\mhyphen TS}}} (\tau_{\delta}(\Pi_{\mathrm{Het\mhyphen TS}}))}{\log\left(\frac{1}{\delta}\right)} \le 2\, (1+\lambda) \, c^*(v).
\]
This completes the desired proof.
\end{proof}

\begin{lemma}
\label{lemma:Tcvg}
Given any $\epsilon >0$ and $v \in \mathcal{P}$,
 $$\mathbb{E}_v^{\Pi_{\mathrm{Het\mhyphen TS}}}[T_{\rm cvg}(v,\epsilon)] < +\infty.$$
\end{lemma}
\begin{proof}[Proof of Lemma \ref{lemma:Tcvg}]
Fix $v \in \mathcal{P}$ and $\epsilon >0$. For any $\epsilon' >0$, let $T'_{N}(v,\epsilon')$ denote the smallest positive integer such that for all $t \ge T'_{N}(v,\epsilon')$,
\begin{equation}
     \left\lvert \frac{N_{i,m}(t)}{t}-\widetilde{\omega}_{i,m}(v) \right\rvert \le \epsilon' \quad \forall m\in[M], i\in S_m.
\end{equation}
Let  $T'_{\mu}(v,\epsilon')$ denote the smallest positive integer such that for all $t \ge T'_{\mu}(v,\epsilon')$,
\begin{equation}
   \lvert \hat{\mu}_{i,m}(t)-\mu_{i,m}(v) \rvert  \le \epsilon'  \quad \forall m\in[M], i\in S_m.
\end{equation}
From \eqref{eq:zt_div_t}, we know that there exists $\epsilon'_1\in (0,1)$ such that
\begin{equation}
    T_{\rm cvg}(v,\epsilon) \le \max\{T'_{N}(v,\epsilon'_1), T'_{\mu}(v,\epsilon'_1)\}.  \label{eq:cfg_sub1}
\end{equation} 
Let $T'_{ \hat{\omega}}(v,\epsilon')$ denote the smallest positive integer such that for all $t \ge T'_{ \hat{\omega}}(v,\epsilon')$,
\begin{equation}
     \lvert \hat{\omega}_{i,m}(t)-\widetilde{\omega}_{i,m}(v) \rvert \le \epsilon' \quad \forall m\in[M], i\in S_m.
\end{equation}
Let  $T'_{\widetilde{\omega}}(v,\epsilon')$ denote the smallest positive integer such that for all $t \ge T'_{\widetilde{\omega}}(v,\epsilon')$,
\begin{equation}
   \lvert \widetilde{\omega}_{i,m}(\hat{v}(t)) - \widetilde{\omega}_{i,m}({v})\rvert  \le \epsilon'  \quad \forall m\in[M], i\in S_m.
\end{equation}
Recall the definition of $\hat{\omega}_{i,m}(t)$ in \eqref{eq:sampling-rule-at-client-m}. We then have 
\begin{equation}
    T'_{ \hat{\omega}}(v,\epsilon') \le (1+\lambda) T'_{\widetilde{\omega}}(v,\epsilon')+1 \quad \forall \epsilon'>0. \label{eq:times1pulslambda}
\end{equation}
In addition, by the D-tracking rule in \eqref{eq:D-tracking-sampling-rule}, we have for all $t\ge \max\{\frac{T'_{\hat{\omega}}(v,\epsilon')}{\epsilon'},\frac{1}{(\epsilon')^2}\}$ and all $\epsilon'\in(0,1)$
\begin{align}
N_{i,m}(t) & \stackrel{(a)}{\le}  \max\{N_{i,m}(T'_{\hat{\omega}}(v,\epsilon')), t(\widetilde{\omega}_{i,m}(v)+\epsilon')+1, \sqrt{t}+1 \} \nonumber \\ 
           & \le   \max\{t\epsilon', t(\widetilde{\omega}_{i,m}(v)+\epsilon')+1, t\epsilon' +1 \} \nonumber \\
           & \le  t(\widetilde{\omega}_{i,m}(v)+\epsilon')+1 \label{eq:tracking_concentrate_1}.
\end{align} 
In $(a)$, the first term inside $\max$ follows from the fact that $\epsilon'<1$, while the second and third terms follow from~\eqref{eq:D-tracking-sampling-rule}.
Consequently, for all $t\ge \max\{\frac{T'_{\hat{\omega}}(v,\epsilon')}{\epsilon'},\frac{1}{\epsilon'^2}, \frac{K}{\epsilon'}\}$ and all $\epsilon'\in(0,1)$, we have
\begin{align}
N_{i,m}(t) & = t - \sum_{j\ne i:j\in S_m} N_{j,m}(t) \nonumber \\
            & \ge t -t \sum_{j\ne i:j\in S_m} (\widetilde{\omega}_{j,m}(v)+\epsilon')- K \nonumber \\
            & = t(\widetilde{\omega}_{i,m}(v) - K \epsilon') - K \nonumber \\
            & \stackrel{(a)}{\ge} t(\widetilde{\omega}_{i,m}(v) - (K+1) \epsilon') \label{eq:tracking_concentrate_2},
\end{align}
where $(a)$ follows $t>\frac{K}{\epsilon'}$.
Then, from \eqref{eq:tracking_concentrate_1} and \eqref{eq:tracking_concentrate_2}, we have
\begin{align}
T'_N(v,\epsilon'_1) & \le \max\left\{\frac{(K+1)T'_{\hat{\omega}}(v,\frac{\epsilon'_1}{K+1})}{\epsilon'_1},\frac{(K+1)^2}{(\epsilon'_1)^2}, \frac{K(K+1)}{\epsilon'_1}\right\}  \nonumber \\
                 & \le \frac{(K+1)T'_{\hat{\omega}}(v, \frac{\epsilon'_1}{K+1})}{\epsilon'_1}+\frac{(K+1)^2}{(\epsilon'_1)^2}+ \frac{K(K+1)}{\epsilon'_1} \nonumber \\
                 &  \stackrel{(a)}{\le} \frac{(K+1)((1+\lambda)T'_{\widetilde{\omega}}(v, \frac{\epsilon'_1}{K+1})+1)}{\epsilon'_1}+\frac{(K+1)^2}{(\epsilon'_1)^2}+ \frac{K(K+1)}{\epsilon'_1},
                \label{eq:cfg_sub2}
\end{align}
where $(a)$ follows from \eqref{eq:times1pulslambda}.
From \eqref{eq:maxcontinue}, we know that there exists $\epsilon'_2>0$ such that
\begin{equation}
    T'_{\widetilde{\omega}}\left(v, \frac{\epsilon'_1}{K+1}\right) \le T'_{\mu}(v, \epsilon'_2).
    \label{eq:cfg_sub3}
\end{equation}
Combining \eqref{eq:cfg_sub1},  \eqref{eq:cfg_sub2}, and  \eqref{eq:cfg_sub3}, we have
\begin{equation}
    T_{\rm cvg}(v,\epsilon) \le T'_{\mu}(v,\epsilon'_1) + \frac{(K+1)((1+\lambda)T'_{\mu}(v, \epsilon'_2)+1)}{\epsilon'_1}+\frac{(K+1)^2}{(\epsilon'_1)^2}+ \frac{K(K+1)}{\epsilon'_1}.
\end{equation}
Note that by the force exploration in D-tracking rule, there exists a constant $T_{\rm start}>0$ (depending only on $K$) such that for all $t\ge T_{\rm start}$,
$$
N_{i,m}(t) \ge \sqrt{\frac{t-1}{K}}-1 \quad \forall m\in[M], i \in S_m,
$$
which in turn implies that for $\epsilon' \in \{\epsilon'_1, \epsilon'_2\}$,
\begin{align}
\mathbb{E}_v^{\Pi_{\mathrm{Het\mhyphen TS}}}[T'_{\mu}(v,\epsilon')] & \le T_{\rm start} + \sum_{t= T_{\rm start}}^{+\infty} \mathbb{P}_v^{\Pi_{\mathrm{Het\mhyphen TS}}}(T'_{\mu}(v,\epsilon') > t) \nonumber \\
& \le T_{\rm start} + \sum_{t= T_{\rm start}}^{+\infty} 2MK\exp\left(-\left(\sqrt{\frac{t-1}{K}}-1\right)(\epsilon')^2/2\right) \nonumber \\
& < +\infty.
\end{align}
Hence,
$$\mathbb{E}_v^{\Pi_{\mathrm{Het\mhyphen TS}}}[T_{\rm cvg}(v,\epsilon)] < +\infty.$$

This concludes the proof.
\end{proof}

\section{Proof of Theorem \ref{theorem:unique-allocation}} \label{sec:prf_allocation}
Fix a problem instance $v \in \mathcal{P}$ arbitrarily. Recall that there exists a common solution to \eqref{eq:optimisation-problem-multiple} (see the discussion after~\eqref{eq:exist_solution}). The following results show that this solution satisfies {\em balanced condition} (Def.~\ref{def:balanced-condition}) and {\em pseudo-balanced condition} (Def.~\ref{def:pseudo-balanced-condition}) and that it is unique. 
\begin{lemma}
\label{lemma:uniform-must-meet-good-2}
The common solution to \eqref{eq:optimisation-problem-multiple} satisfies pseudo-balanced condition.
\end{lemma}
\begin{proof}
Let $\widetilde{\omega}(v)$ be a common solution to \eqref{eq:optimisation-problem-multiple}. Let 
\begin{equation}
    Q^{\rm min}_j \coloneqq \argmin_{i \in Q_j}  \frac{\Delta^2_{i}(v)}{\frac{1}{M^2_{i}}\sum_{m=1}^M \mathbf{1}_{\{i \in S_m\}} \frac{1}{\widetilde{\omega}_{i,m}(v)}}, \quad j \in [L].
    \label{eq:Q_j_min}
\end{equation}
Suppose $\widetilde{\omega}(v)$ does not meet {\em pseudo-balanced condition}. Then, there  exists $l \in [L]$ such that $Q_l \ne Q^{\rm min}_l$. We now recursively construct an $\omega \in \Gamma$ such that $ \widetilde{g}^{(l)}_v(\omega) > \widetilde{g}^{(l)}_v(\widetilde{\omega}(v)) $, thereby leading to a contradiction.

\vspace{0.3cm}

\noindent{\bf Step 1: Initialization.} Set $\omega^{(0)} \coloneqq \widetilde{\omega}(v)$ and $Q^{(0)} \coloneqq Q^{\rm min}_l$.

\noindent
{\bf Step 2: Iterations.} For each $s\in \{0,1,2,\ldots\, |Q_l^{\rm min}|-1\}$, note that there exists $i_1 \in Q^{(s)}$, $i_2 \in Q_l \setminus Q^{(s)}$, and $m' \in [M]$ such that $i_1,i_2 \in S_{m'}$. Let $\epsilon > 0$ be sufficiently small so that 
\[
    \frac{\Delta^2_{i_2}(v)}{\frac{1}{M^2_{i_2}}\sum_{m=1}^M \mathbf{1}_{\{i_2 \in S_m\}} \frac{1}{\omega^{(s)}_{i_2,m}-\epsilon}} > \min_{i \in Q_{l}}  \frac{\Delta^2_{i}(v)}{\frac{1}{M^2_{i}}\sum_{m=1}^M \mathbf{1}_{\{i \in S_m\}} \frac{1}{\widetilde{\omega}_{i,m}(v)} } =\widetilde{g}^{(l)}_v(\widetilde{\omega}(v)) .
\]
Then, we construct $\omega^{(s+1)}$ as
\[
\forall\, m\in [M], i\in S_m, \ \ 
\omega^{(s+1)}_{i,m} \coloneqq
\begin{cases}
      \omega^{(s)}_{i,m}-\epsilon,\ &\text{if } i=i_2, m=m' \\
      \omega^{(s)}_{i,m}+\epsilon,\ &\text{if } i=i_1, m=m', \\
      \omega^{(s)}_{i,m},\ &\text{otherwise}, 
\end{cases}
\]
and set $Q^{(s+1)} \coloneqq Q^{(s)} \setminus \{i_1\}$. We then have
\[
\frac{\Delta^2_{i}(v)}{\frac{1}{M^2_{i}}\sum_{m=1}^M \mathbf{1}_{\{i \in S_m\}} \frac{1}{\omega^{(s+1)}_{i,m}}} > \widetilde{g}^{(l)}_v(\widetilde{\omega}(v)) \quad \forall \, i \in Q_l \setminus Q^{(s+1)}, 
\]
and 
\[
\frac{\Delta^2_{i}(v)}{\frac{1}{M^2_{i}}\sum_{m=1}^M \mathbf{1}_{\{i \in S_m\}} \frac{1}{\omega^{(s+1)}_{i,m}}} = \widetilde{g}^{(l)}_v(\widetilde{\omega}(v)) \quad \forall\, i \in Q^{(s+1)}.
\]
By following the above procedure for $s\in \{0, 1, 2, \ldots, |Q_l^{\rm min}|\}$, we arrive at $\omega^{(\lvert Q^{\rm min}_l \rvert)}$ such that $ \widetilde{g}^{(l)}_v(\omega^{(\lvert Q^{\rm min}_l \rvert)}) > \widetilde{g}^{(l)}_v(\widetilde{\omega}(v)) $, which is the clearly a contradiction.
\end{proof}

\begin{lemma}
\label{lemma:uniform-must-meet-good-1}
The common solution to \eqref{eq:optimisation-problem-multiple} satisfies {\em balanced condition}.
\end{lemma}
\begin{proof}
Let $\widetilde{\omega}(v)$ be a common solution to \eqref{eq:optimisation-problem-multiple}. From Lemma \ref{lemma:uniform-must-meet-good-2}, we know that $\widetilde{\omega}(v)$ satisfies {\em pseudo-balanced condition}. Suppose now that $\widetilde{\omega}(v)$ does not satisfy {\em balanced condition}. Then, there exists $l \in [L]$, $m_1,m_2 \in [M]$, and $i_1, i_2 \in S_{m_1} \cap S_{m_2} \subseteq Q_{l}$ such that 
\begin{equation}
     \label{eq:repeat-cond2}
     \frac{\widetilde{\omega}_{i_1,m_1}(v)}{\widetilde{\omega}_{i_2,m_1}(v)}  > \frac{\widetilde{\omega}_{i_1,m_2}(v)}{\widetilde{\omega}_{i_2,m_2}(v)}.
\end{equation}
Because $\widetilde{\omega}(v)$ satisfies {\em pseudo-balanced condition}, we must have
\[
\frac{\Delta^2_{i_2}(v)}{\frac{1}{M^2_{i_2}}\sum_{m=1}^M \mathbf{1}_{\{i_2 \in S_m\}} \frac{1}{\widetilde{\omega}_{i_2,m}(v)}} = \frac{\Delta^2_{i_1}(v)}{\frac{1}{M^2_{i_1}}\sum_{m=1}^M \mathbf{1}_{\{i_1 \in S_m\}} \frac{1}{\widetilde{\omega}_{i_1,m}(v)}} = \widetilde{g}^{(l)}_v(\widetilde{\omega}(v)).
\]

\noindent
Note that \eqref{eq:repeat-cond2} implies that $\frac{\widetilde{\omega}_{i_2,m_1}(v)}{\widetilde{\omega}_{i_2,m_2}(v)} < \frac{\widetilde{\omega}_{i_1,m_1}(v)}{\widetilde{\omega}_{i_1,m_2}(v)}  $. 
Let $\rho$ be any value such that
\begin{equation}
    \left(\frac{\widetilde{\omega}_{i_2,m_1}(v)}{\widetilde{\omega}_{i_2,m_2}(v)}\right)^2 < \rho < \left(\frac{\widetilde{\omega}_{i_1,m_1}(v)}{\widetilde{\omega}_{i_1,m_2}(v)}\right)^2.
    \label{eq:defining-rho}
\end{equation}

Using the fact that the derivative of $x \mapsto \frac{1}{x}$ is $ -\frac{ 1}{x^2}$, we have
\begin{equation}
\label{eq:o_epsilon_1}
 \frac{1}{\widetilde{\omega}_{i_1,m_1}(v) - \rho\epsilon} -    \frac{1}{\widetilde{\omega}_{i_1,m_1}(v)}  = \frac{\rho\epsilon}{\widetilde{\omega}^2_{i_1,m_1}(v)} + o(\epsilon)\quad\mbox{as }\epsilon\to0,
\end{equation}
and
\begin{equation}
\label{eq:o_epsilon_2}
 \frac{1}{\widetilde{\omega}_{i_1,m_2}(v)} -    \frac{1}{\widetilde{\omega}_{i_1,m_2}(v) + \epsilon}  = \frac{\epsilon}{\widetilde{\omega}^2_{i_1,m_2}(v)} + o(\epsilon)\quad\mbox{as }\epsilon\to0.
\end{equation}
Here $o(\epsilon)$ is a function in $\epsilon$ that satisfies $\lim_{\epsilon\to0}\frac{o(\epsilon)}{\epsilon}=0$.
By combining these equations,
\begin{align}
    &\frac{1}{\widetilde{\omega}_{i_1,m_1}(v) - \rho\epsilon} -    \frac{1}{\widetilde{\omega}_{i_1,m_1}(v)}-\bigg( \frac{1}{\widetilde{\omega}_{i_1,m_2}(v)} -    \frac{1}{\widetilde{\omega}_{i_1,m_2}(v) + \epsilon}  \bigg) \nonumber\\
    & = \frac{\rho\epsilon}{\widetilde{\omega}^2_{i_1,m_1}(v)} + o(\epsilon) - \bigg(\frac{\epsilon}{\widetilde{\omega}^2_{i_1,m_2}(v)} + o(\epsilon)  \bigg)\\
    &=\epsilon \bigg[  \frac{\rho }{\widetilde{\omega}^2_{i_1,m_1}(v)} \big(1+o_\epsilon(1) \big) - \frac{1}{\widetilde{\omega}^2_{i_1,m_2}(v)} \big(1+o_\epsilon(1) \big) \bigg] \label{eqn:rearrange_rho}
\end{align}
where $o_\epsilon(1)$ is a term that vanishes as $\epsilon\downarrow 0$.
From \eqref{eq:defining-rho}, we have,
\begin{align}
    \rho < \frac{\widetilde{\omega}_{i_1, m_1}^2(v)}{\widetilde{\omega}_{i_1, m_2}^2(v)} & \Longleftrightarrow \frac{\rho}{\widetilde{\omega}_{i_1, m_1}^2(v)} < \frac{1}{\widetilde{\omega}_{i_1, m_2}^2(v)}\label{eqn:rho_ineq}
\end{align}
By~\eqref{eqn:rearrange_rho} and~\eqref{eqn:rho_ineq}, there exists $\epsilon_1>0$ such that for all $\epsilon \in (0,\epsilon_1]$, 
\begin{align}
    \frac{1}{\widetilde{\omega}_{i_1,m_1}(v) - \rho\epsilon} -    \frac{1}{\widetilde{\omega}_{i_1,m_1}(v)}-\bigg( \frac{1}{\widetilde{\omega}_{i_1,m_2}(v)} -    \frac{1}{\widetilde{\omega}_{i_1,m_2}(v) + \epsilon}  \bigg)<0.
\end{align}
In other words, for all $\epsilon\in (0,\epsilon_1]$,
\begin{equation}
\label{eq:arm_1_mod}
\frac{1}{\widetilde{\omega}_{i_1,m_1}(v)}  + \frac{1}{\widetilde{\omega}_{i_1,m_2}(v)} > \frac{1}{\widetilde{\omega}_{i_1,m_1}(v) -  \rho \epsilon}  + \frac{1}{\widetilde{\omega}_{i_1,m_2}(v) + \epsilon}.
\end{equation}
Similarly, there exists $\epsilon_2 >0 $ for any $\epsilon \in (0, \epsilon_2]$ 
such that 
\begin{equation}
    \label{eq:arm_2_mod}
    \frac{1}{\widetilde{\omega}_{i_2,m_1} (v)}  + \frac{1}{\widetilde{\omega}_{i_2,m_2}(v)}  >
\frac{1}{\widetilde{\omega}_{i_2,m_1}(v) +\rho \epsilon}  + \frac{1}{\widetilde{\omega}_{i_2,m_2}(v) - \epsilon} .
\end{equation}
Set $\epsilon=\min\{\epsilon_1, \epsilon_2\}$. Let $\omega' \in \Gamma$ be defined as 
\begin{equation}
\forall\, m\in [M], i\in S_m, \ \ 
\omega'_{i,m} \coloneqq
\begin{cases}
      \widetilde{\omega}_{i,m}(v)-\rho\epsilon,\ &\text{if } i=i_1, m=m_1 \\
      \widetilde{\omega}_{i,m}(v)+\epsilon,\ &\text{if } i=i_1, m=m_2 \\
      \widetilde{\omega}_{i,m}(v) + \rho\epsilon,\ &\text{if } i=i_2, m=m_1 \\
      \widetilde{\omega}_{i,m}(v) - \epsilon,\ &\text{if } i=i_2, m=m_2 \\
      \widetilde{\omega}_{i,m}(v),\ &\text{otherwise}. 
\end{cases}
\end{equation}
Then, from \eqref{eq:arm_1_mod} and \eqref{eq:arm_2_mod}, we have
\begin{equation}
\label{eq:i1i2larger} 
\frac{\Delta^2_{i}(v)}{\frac{1}{M^2_{i}}\sum_{m=1}^M \mathbf{1}_{\{i \in S_m\}} \frac{1}{{\omega}^{\prime}_{i,m}}} >
\frac{\Delta^2_{i}(v)}{\frac{1}{M^2_{i}}\sum_{m=1}^M \mathbf{1}_{\{i \in S_m\}} \frac{1}{\widetilde{\omega}_{i,m}(v)}} \quad \forall\, i \in \{i_1,i_2\},
\end{equation}
and 
\begin{equation}
\label{eq:excepti1i2remain}
\frac{\Delta^2_{i}(v)}{\frac{1}{M^2_{i}}\sum_{m=1}^M \mathbf{1}_{\{i \in S_m\}} \frac{1}{{\omega}^{\prime}_{i,m}}} =
\frac{\Delta^2_{i}(v)}{\frac{1}{M^2_{i}}\sum_{m=1}^M \mathbf{1}_{\{i \in S_m\}} \frac{1}{\widetilde{\omega}_{i,m}(v)}} \quad \forall \, i \in Q_l \setminus \{i_1,i_2\}.
\end{equation}
We then consider the following two cases.\\
\noindent \underline{\bf Case 1}: $Q_{l} = \{i_1,i_2\}$. In this case, it follows from \eqref{eq:i1i2larger} that $\widetilde{g}^{(l)}_v(\omega') >\widetilde{g}^{(l)}_v(\widetilde{\omega}(v))$, which contradicts with the fact that $\widetilde{\omega}(v)$ is an optimum solution to
$\max_{\omega \in \Gamma} \widetilde{g}_v(\omega)$.

\noindent \underline{\bf Case 2}: $ \{i_1,i_2\} \subsetneq Q_{l}$. In this case, it follows from \eqref{eq:excepti1i2remain} that $\widetilde{g}^{(j)}_v(\omega') = \widetilde{g}^{(j)}_v(\widetilde{\omega}(v))$ for all $ j \in [L]$, which implies that $\omega'$ is a common solution to \eqref{eq:optimisation-problem-multiple} just as $\widetilde{\omega}(v)$ is. However, note that the right-hand sides of  \eqref{eq:excepti1i2remain} and \eqref{eq:i1i2larger} are equal because  $\widetilde{\omega}(v)$ satisfies {\em pseudo-balanced condition}. As a result, it follows that
\begin{equation}
\frac{\Delta^2_{i_1}(v)}{\frac{1}{M^2_{i_1}}\sum_{m=1}^M \mathbf{1}_{\{i_1 \in S_m\}} \frac{1}{\omega'_{i_1,m}}} >\frac{\Delta^2_{i}(v)}{\frac{1}{M^2_{i}}\sum_{m=1}^M \mathbf{1}_{\{i \in S_m\}} \frac{1}{{\omega}^{\prime}_{i,m}}} \quad \forall\, i \in Q_l \setminus\{i_1, i_2\}.
\end{equation}
This shows that $\omega'$ does not meet {\em pseudo-balanced condition}, thereby contradicting Lemma \ref{lemma:uniform-must-meet-good-2}.
\end{proof}

\begin{lemma}
\label{lemma:uniform-must-unique}
The common solution to \eqref{eq:optimisation-problem-multiple} is unique.
\end{lemma}
\begin{proof}
Suppose that $\widetilde{\omega}(v)$ and $\widetilde{\omega}'(v)$ are two common solutions to \eqref{eq:optimisation-problem-multiple}. Suppose further that $\widetilde{\omega}(v) \neq \widetilde{\omega}'(v)$. In the following, we arrive at a contradiction. Let $\widetilde{\omega}^{\rm avg}(v) \coloneqq (\widetilde{\omega}(v)+\widetilde{\omega}'(v))/2$. From Lemma \ref{lemma:uniform-must-meet-good-2}, we know that $\widetilde{\omega}(v)$ and $\widetilde{\omega}'(v)$ meet {\em pseudo-balanced condition}. This implies that
\begin{equation}
    \label{eq:cross-euqal}
    \frac{\Delta^2_{i}(v)}{\frac{1}{M^2_{i}}\sum_{m=1}^M \mathbf{1}_{\{i \in S_m\}} \frac{1}{\widetilde{\omega}_{i,m}(v)}} = \frac{\Delta^2_{i}(v)}{\frac{1}{M^2_{i}}\sum_{m=1}^M \mathbf{1}_{\{i \in S_m\}} \frac{1}{\widetilde{\omega}'_{i,m}(v)}} \quad \forall \, i \in [K],
\end{equation}
which in turn implies that 
\begin{equation}
    \label{eq:cross-euqal-1}
    {\sum_{m=1}^M \mathbf{1}_{\{i \in S_m\}} \frac{1}{\widetilde{\omega}_{i,m}(v)}} = {\sum_{m=1}^M \mathbf{1}_{\{i \in S_m\}} \frac{1}{\widetilde{\omega}'_{i,m}(v)}}.
\end{equation}
Using the relation $\frac{2}{(a+b)/2} < \frac{1}{a}+\frac{1}{b}$ whenever $a,b>0$ and $a \ne b$, we get that 
\begin{equation}
\frac{2}{\widetilde{\omega}^{\rm avg}_{i,m}(v)} < \frac{1}{\widetilde{\omega}_{i,m}(v)} + \frac{1}{\widetilde{\omega}'_{i,m}(v)} \quad \forall\, m\in [M], \, i\in S_m \;  \text{ such that }\widetilde{\omega}_{i,m}(v) \ne \widetilde{\omega}'_{i,m}(v).
\label{eq:avg-less-than-sum}
\end{equation}
Let $Q_{\rm diff} \coloneqq \{\iota \in [K]: \exists\, m, \widetilde{\omega}_{\iota,m}(v) \ne \widetilde{\omega}'_{\iota,m}(v)\}$.  As a consequence of \eqref{eq:avg-less-than-sum}, for all $i\in Q_{\rm diff} $, we have
\begin{align}
& {\sum_{m=1}^M \mathbf{1}_{\{i \in S_m\}} \frac{2}{\widetilde{\omega}^{\rm avg}_{i,m}(v)}} <
{\sum_{m=1}^M \mathbf{1}_{\{i \in S_m\}} \frac{1}{\widetilde{\omega}_{i,m}(v)}} + {\sum_{m=1}^M \mathbf{1}_{\{i \in S_m\}} \frac{1}{\widetilde{\omega}'_{i,m}(v)}} \nonumber \\
& \stackrel{(a)}{\implies} {\sum_{m=1}^M \mathbf{1}_{\{i \in S_m\}} \frac{1}{\widetilde{\omega}^{\rm avg}_{i,m}(v)}} < {\sum_{m=1}^M \mathbf{1}_{\{i \in S_m\}} \frac{1}{\widetilde{\omega}_{i,m}(v)}} \nonumber \\
& \implies  \frac{\Delta^2_{i}(v)}{\frac{1}{M^2_{i}}\sum_{m=1}^M \mathbf{1}_{\{i \in S_m\}} \frac{1}{\widetilde{\omega}^{\rm avg}_{i,m}(v)}} > \frac{\Delta^2_{i}(v)}{\frac{1}{M^2_{i}}\sum_{m=1}^M \mathbf{1}_{\{i \in S_m\}} \frac{1}{\widetilde{\omega}_{i,m}(v)}}, \label{eq:violate-good-2}
\end{align}
where $(a)$ follows from \eqref{eq:cross-euqal}.
In addition, it is clear to observe that for all $i \in [K] \setminus Q_{\rm diff}$,
\begin{equation}
\label{eq:trivial}
    \frac{\Delta^2_{i}(v)}{\frac{1}{M^2_{i}}\sum_{m=1}^M \mathbf{1}_{\{i \in S_m\}} \frac{1}{\widetilde{\omega}^{\rm avg}_{i,m}(v)}} = \frac{\Delta^2_{i}(v)}{\frac{1}{M^2_{i}}\sum_{m=1}^M \mathbf{1}_{\{i \in S_m\}} \frac{1}{\widetilde{\omega}_{i,m}(v)}}.
\end{equation}
By \eqref{eq:violate-good-2} and  \eqref{eq:trivial} we know that  $\widetilde{g}^{(j)}_v(\widetilde{\omega}^{\rm avg}(v)) \ge \widetilde{g}^{(j)}_v(\widetilde{\omega}(v))$ for each $j\in [L]$, which implies that $\widetilde{\omega}^{\rm avg}(v)$ is a common solution to \eqref{eq:optimisation-problem-multiple}. Now, there must exist $l \in [L]$ such that $Q_{\rm diff} \cap Q_l \ne \emptyset$, and then we consider two cases.

\noindent \underline{\bf Case 1}: $Q_{\rm diff} \cap Q_l = Q_l$.
Eq.~\eqref{eq:violate-good-2} implies that $\widetilde{g}^{(l)}_v(\widetilde{\omega}^{\rm avg}(v)) > \widetilde{g}^{(l)}_v(\widetilde{\omega}(v))$, which contradicts the fact that $\widetilde{\omega}(v)$ and $\widetilde{\omega}^{\rm avg}(v)$ are the common solution to \eqref{eq:optimisation-problem-multiple}.

\noindent \underline{\bf Case 2}: $Q_{\rm diff} \cap Q_l \subsetneq Q_l$.
 Note that the right-hand side of \eqref{eq:violate-good-2} and  \eqref{eq:trivial} are equal because $\widetilde{\omega}(v)$ meets pseudo-balanced condition. Hence, there exists $i_1,i_2 \in Q_l$ such that
 \[
  \frac{\Delta^2_{i_1}(v)}{\frac{1}{M^2_{i_1}}\sum_{m=1}^M \mathbf{1}_{\{i_1 \in S_m\}} \frac{1}{\widetilde{\omega}^{\rm avg}_{i_1,m}(v)}} >
   \frac{\Delta^2_{i_2}(v)}{\frac{1}{M^2_{i_2}}\sum_{m=1}^M \mathbf{1}_{\{i \in S_m\}} \frac{1}{\widetilde{\omega}^{\rm avg}_{i_2,m}(v)}} 
 \]
 which means that $\widetilde{\omega}^{\rm avg}(v)$ violates {\em pseudo-balanced condition}, a contradiction to Lemma \ref{lemma:uniform-must-meet-good-2}.
\end{proof}

Finally, Theorem \ref{theorem:unique-allocation} follows from Lemma \ref{lemma:uniform-must-meet-good-2}, Lemma \ref{lemma:uniform-must-meet-good-1}, and Lemma \ref{lemma:uniform-must-unique}.

\section{Proof of Proposition \ref{proposition:compute-global-vector}}
\label{sec:proofs-of-optimal-allocation}
Before proving Proposition \ref{proposition:compute-global-vector}, we first present two important results. The first result below asserts that $G^{(j)}(v)$ is an eigenvector of $H^{(j)}(v)$ with $\frac{1}{g_v^{(j)}(\widetilde{\omega}(v))}$ as its eigenvalue for every $j \in [L]$.

\begin{lemma}
\label{lemma:globalvector-meet-eigenvector}
    Given $v \in \mathcal{P}$,  $G^{(j)}(v)$ is a eigenvector of $H^{(j)}(v)$ with eigenvalue $\frac{1}{g_v^{(j)}(\widetilde{\omega}(v))}$ for all $j \in [L]$, i.e.,
\begin{equation}
    H^{(j)}(v) \, G^{(j)}(v) =\frac{1}{g_v^{(j)}(\widetilde{\omega}(v))}\, G^{(j)}(v) \quad \forall \, j\in [L].
    \label{eq:G-is-an-eigenvector-of-H}
\end{equation}
\end{lemma}
\begin{proof}
Fix $j \in [L]$ and a problem instance $v$ arbitrarily. Recall that $\widetilde{\omega}(v)$ is the unique common solution \eqref{eq:optimisation-problem-multiple} and $G(v)$ is the global vector characterising $\widetilde{\omega}(v)$ uniquely (via \eqref{eq:unique-GV}).  From Lemma \ref{lemma:uniform-must-meet-good-2}, we know that $\widetilde{\omega}(v)$ satisfies {\em pseudo-balanced condition}, which
implies that for any $i \in Q_j$,
\begin{align}
     & \frac{\Delta^2_{i}(v)}{\frac{1}{M^2_{i}}\sum_{m=1}^M \mathbf{1}_{\{i \in S_m\}} \frac{1}{\widetilde{\omega}(v)_{i,m}} } =\widetilde{g}_{v}^{(j)}(\widetilde{\omega}(v)) \nonumber \\ 
& \implies \frac{\sum_{m=1}^M \mathbf{1}_{\{i \in S_m\}} \frac{1}{\widetilde{\omega}(v)_{i,m}} }{M^2_{i} \Delta^2_{i}(v)} =\frac{1}{\widetilde{g}_{v}^{(j)}(\widetilde{\omega}(v))} \nonumber \\
& \stackrel{(a)}{\implies} \frac{\sum_{m=1}^M \mathbf{1}_{\{i \in S_m\}} \frac{\sum_{\iota \in S_m} 
G(v)_\iota}{G(v)_i}}{M^2_{i} \Delta^2_{i}(v)} =\frac{1}{\widetilde{g}_{v}^{(j)}(\widetilde{\omega}(v))} \nonumber  \\
& \implies \frac{\sum_{m=1}^M \mathbf{1}_{\{i \in S_m\}} {\sum_{\iota \in S_m} 
G(v)_\iota}}{M^2_{i} \Delta^2_{i}(v)} =\frac{{G(v)_i}}{\widetilde{g}_{v}^{(j)}(\widetilde{\omega}(v))} \nonumber  \\
& \implies \sum_{\iota \in Q_j}  G(v)_\iota \left( \frac{\sum_{m=1}^M \mathbf{1}_{\{i,\iota \in S_m\}} }{M^2_{i} \Delta^2_{i}(v)} \right) =\frac{{G(v)_i}}{\widetilde{g}_{v}^{(j)}(\widetilde{\omega}(v))}. \label{eq:matrxi-multi-form}
\end{align}
In the above set of implications, $(a)$ follows from \eqref{eq:unique-GV}. Noting that \eqref{eq:matrxi-multi-form} is akin to \eqref{eq:G-is-an-eigenvector-of-H} completes the desired proof.
\end{proof}
The second result below asserts that the eigenspace of $H^{(j)}(v)$ is one-dimensional.
\begin{lemma}
\label{lemma:spaceone}
     Given $v \in \mathcal{P}$, the dimension of the eigenspace of $H^{(j)}(v)$ associated with the eigenvalue  $\frac{1}{g_v^{(j)}(\widetilde{\omega}(v))}$ is equal to one for all $j \in [L]$. 
\end{lemma}
\begin{proof}
Fix $j \in [L]$ and a problem instance $v \in \mathcal{P}$ arbitrarily. It is easy to verify that $G^{(j)}(v)$ has strictly positive entries (else, $\widetilde{g}_v({\widetilde{\omega}(v)})=0$). Suppose that $\mathbf{u}\in \mathbb{R}^{\lvert Q_j \rvert}$ is another eigenvector of $H^{(j)}(v)$ corresponding to the eigenvalue $\frac{1}{g_v^{(j)}(\widetilde{\omega}(v))}$ and $\{\mathbf{u},\, G^{(j)}(v)\}$ is linearly independent. Let
\[
\mathbf{u}' \coloneqq G^{(j)}(v) + \epsilon \mathbf{u},
\] 
where $\epsilon > 0$ is any number such that each entry of $\mathbf{u}'$ is strictly positive. 
Let $\omega' \in \Gamma $ be defined as
\[
\forall\, m\in [M], i\in S_m, \ \ 
\omega'_{i,m} =
\begin{cases}
     
      \frac{\mathbf{u}'_{{\rm Idx}(i)}}{\sum_{\iota \in S_m } \mathbf{u}'_{{\rm Idx}(\iota)}}\  &\text{if}\ i\in Q_j \\
      \widetilde{\omega}(v)_{i,m}\ &\text{otherwise}, 
\end{cases}
\]
where for any $i\in Q_j$, ${\rm Idx}(i) \in [\lvert Q_j \rvert ]$ represents the index of arm $i$ within the arms set $Q_j$. Then, it follows from the definition of $\omega'$ that $\widetilde{g}_{v}^{(l)}(\widetilde{\omega}(v)) = \widetilde{g}_{v}^{(l)}(\omega')$ for all $l \ne j$ and $\omega' \ne \widetilde{\omega}(v)$. Note that $\mathbf{u}'$ is also an eigenvector of $H^{(j)}(v)$ corresponding to the eigenvalue $\frac{1}{g_v^{(j)}(\widetilde{\omega}(v))}$. This means that for all $i\in Q_j$,
\begin{align}
    & \sum_{\iota \in Q_j}  \mathbf{u}'_{{\rm Idx}(\iota)} \left( \frac{\sum_{m=1}^M \mathbf{1}_{\{i,\iota \in S_m\}} }{M^2_{i} \Delta^2_{i}(v)} \right) =\frac{ \mathbf{u}'_{{\rm Idx}(i)}}{\widetilde{g}_{v}^{(j)}(\widetilde{\omega}(v))} \nonumber  \\
    & \implies \frac{\sum_{m=1}^M \mathbf{1}_{\{i \in S_m\}} {\sum_{\iota \in S_m} 
\mathbf{u}'_{{\rm Idx}(\iota)}}}{M^2_{i} \Delta^2_{i}(v)} =\frac{{\mathbf{u}'_{{\rm Idx}(i)}}}{\widetilde{g}_{v}^{(j)}(\widetilde{\omega}(v))} \nonumber  \\
& \implies \frac{\sum_{m=1}^M \mathbf{1}_{\{i \in S_m\}} \frac{\sum_{\iota \in S_m} 
\mathbf{u}'_{{\rm Idx}(\iota)}}{\mathbf{u}'_{{\rm Idx}(i)}}}{M^2_{i} \Delta^2_{i}(v)} =\frac{1}{\widetilde{g}_{v}^{(j)}(\widetilde{\omega}(v))} \nonumber  \\
 &  \implies \frac{\Delta^2_{i}(v)}{\frac{1}{M^2_{i}}\sum_{m=1}^M \mathbf{1}_{\{i \in S_m\}} \frac{1}{\omega'_{i,m}} } =\widetilde{g}_{v}^{(j)}(\widetilde{\omega}(v)).
 \label{eq:uprime_to_g} 
\end{align}
From \eqref{eq:uprime_to_g}, it is clear that $\widetilde{g}_{v}^{(l)}(\widetilde{\omega}(v)) = \widetilde{g}_{v}^{(l)}(\omega')$ for all $l \in [L]$, which contradicts Lemma \ref{lemma:uniform-must-unique}. Thus, there is no eigenvector $\mathbf{u}\in \mathbb{R}^{\lvert Q_j \rvert}$ of $H^{(j)}(v)$ corresponding to the eigenvalue $\frac{1}{g_v^{(j)}(\widetilde{\omega}(v))}$ such that $\mathbf{u}$ and  $G^{(j)}(v)>0$ are linearly independent.
This completes the desired proof.
\end{proof}

\subsection{Proof of Proposition \ref{proposition:compute-global-vector}}
Let $\mathbf{v}$ be any eigenvector of $H^{(j)}(v)$ whose eigenvalue is not equal to $\frac{1}{g_v^{(j)}(\widetilde{\omega}(v))}$. Because $H^{(j)}(v)$ is a normal matrix, its eigenvectors corresponding to distinct eigenvalues are orthogonal \cite[Chapter 2, Section 2.5]{horn2012matrix}. This implies that $\langle \mathbf{v}, G^{(j)}(v)  \rangle=0$, where $\langle \cdot,\cdot \rangle$ denotes the vector inner product operator.
Note that $G^{(j)}(v)$ has strictly positive entries. Therefore, the entries of $\mathbf{v}$ cannot be all positive or all negative.

From  Lemma \ref{lemma:spaceone}, we know that any eigenvector $\mathbf{v'}$ associated with the eigenvalue  $\frac{1}{g_v^{(j)}(\widetilde{\omega}(v))}$ should satisfy 
\begin{equation}
\label{eq:reason1.5}
\mathbf{v'} = \alpha G^{(j)}(v),\quad  \text{for some } \alpha \in \mathbb{R}\setminus \{0\},
\end{equation}
which implies that the entries of $\mathbf{v}^{\prime}$ are either all positive or all negative.
Also, Lemma \ref{lemma:spaceone} implies that among any complete set of eigenvectors of $H^{(j)}(v)$, there is only one eigenvector $\mathbf{u}$ with eigenvalue $\frac{1}{g_v^{(j)}(\widetilde{\omega}(v))}$. From the exposition above, it then follows that the entries of $\mathbf{u}$ must be all positive or all negative. Noting that $G^{(j)}(v)$ has unit norm (see \eqref{eq:unique-GV}), we arrive at the form in \eqref{eq:all-positive-or-all-negative-entries}.
This completes the proof.


\end{document}